\documentclass{article}

\usepackage{microtype}
\usepackage{graphicx}
\usepackage{subfigure}
\usepackage{booktabs} %
\usepackage{kotex}
\usepackage{placeins}
\usepackage{enumitem}
\usepackage{hyperref}
\usepackage{comment}  %

\usepackage[accepted]{arxiv2025}
\title{
Mind the Gap:}

\usepackage{amsmath}
\usepackage{amssymb}
\usepackage{mathtools}
\usepackage{amsthm}

\usepackage[capitalize,noabbrev]{cleveref}

\theoremstyle{plain}

\theoremstyle{definition}

\theoremstyle{remark}

\usepackage[textsize=tiny]{todonotes}

\begin{document}

\twocolumn[
\title{Mind the Gap: Aligning the Brain with Language Models \\Requires a Nonlinear and Multimodal Approach}

\vspace{-10mm}
\author{Danny Dongyeop Han$^1$, Yunju Cho$^1$, Jiook Cha$^{1,*}$, Jay-Yoon Lee$^{1,*}$\\
\\
$^1$Seoul National University\\
\\
$^*$Corresponding authors:\\
Jiook Cha (connectome@snu.ac.kr)\\
Jay-Yoon Lee (lee.jayyoon@snu.ac.kr)
}

\maketitle

\vskip 0.3in
]

\begin{abstract}

Self-supervised language and audio models effectively predict brain responses to speech. However, traditional prediction models rely on linear mappings from unimodal features, despite the complex integration of auditory signals with linguistic and semantic information across widespread brain networks during speech comprehension. Here, we introduce a nonlinear, multimodal prediction model that combines audio and linguistic features from pre-trained models (e.g., LLAMA, Whisper). 
Our approach achieves a 17.2\% and 17.9\% improvement in prediction performance (unnormalized and normalized correlation) over traditional unimodal linear models, as well as a 7.7\% and 14.4\% improvement, respectively, over prior state-of-the-art models. %
These improvements represent a major step towards future robust in-silico testing and improved decoding performance.
They also reveal how auditory and semantic information are fused in motor, somatosensory, and higher-level semantic regions, aligning with existing neurolinguistic theories. Overall, our work highlights the often neglected potential of nonlinear and multimodal approaches to brain modeling, paving the way for future studies to embrace these strategies in naturalistic neurolinguistics research.

\end{abstract}

\section{Introduction}
The brain seamlessly deciphers spoken language, integrating auditory signals with linguistic and semantic meaning through a dynamic interplay of neural networks. This process relies on the brain’s capacity to combine information from multiple modalities (e.g., auditory, linguistic, and motor systems)  \citep{speech_multimodal,gpt_ghazanfar2006neocortex}. Furthermore, this process is inherently nonlinear, involving hierarchical and spatiotemporal transformations across distributed brain regions \citep{nonlienar_speech_perception}. Understanding these complex mechanisms is crucial not only for advancing cognitive neuroscience but also for developing brain inspired artificial intelligence systems. %

Language encoding models, which predict brain activity from speech stimuli, are a powerful tool for unraveling the neural processes of speech comprehension \citep{gpt_naselaris2011encoding,huthsemanticdecoding, huth_audio_model_NIPS, encoding_model_8_lebel2021voxelwise, encoding_model_4_jain2018incorporating, encoding_model_6_goldstein2022shared}. Unlike traditional contrast-based paradigms that rely on carefully controlled experiments, encoding models capitalize on naturalistic stimuli (e.g., spoken language) to capture real-world brain processing. This allows for a more comprehensive understanding of the brain’s activity in response to complex and ecologically valid tasks, offering greater generalizability compared to simplified, contrast-based tasks \citep{why_encoding_insilicoexperimentation}. Moreover, these models are increasingly used for in-silico experiments that help test brain function without collecting new data, \citep{why_encoding_insilicoexperimentation,why_encoding_huth_bashivan2019neural,why_encoding_huth_BODL_predictions_wehbe2016} and to build decoding models for language comprehension \cite{huthsemanticdecoding}. %

Early encoding models primarily mapped simple acoustic features (e.g., spectrograms) to brain activity \citep{encoding_model_3_hierarchical}. The introduction of word embeddings \citep{word2vec} enabled the incorporation of semantic information, revealing how meaning is represented across the brain \citep{encoding_model_2_huth2016}. Recent advances in large language models (LLMs) and sophisticated audio models have further enriched these features, leading to substantial gains in prediction accuracy \citep{huthscaling, huth_audio_model_NIPS}. However, most studies still rely on linear mappings of unimodal features, which fail to capture two fundamental principles of neural language processing.

First, the brain operates through nonlinear computations \citep{fMRIisNonlinear,dendTCN,nonlienar_speech_perception}, enabling complex spatiotemporal relationships across neural systems. Second, speech comprehension is a multimodal process, needing integration of diverse information sources (voice, gesture, linguistic) across distributed neural networks \citep{speech_multimodal}. These principles, along with the Motor Theory of Speech Perception \citep{motortheoryofspeechperction_org_paper_1} and the Convergence Divergence Zone model \citep{nature_ref_2_convergence_zone_ORG}, highlight the importance of encoding models capturing nonlinear dynamics and multimodal integration to reveal the brain’s functional organization. %

Although prior studies have explored multimodal models combining linguistic with visual features \citep{oota2022visio_language_encoding,wang2022incorporating_language_vision,mindseye_visiondecoding}, the integration of\textbf{ \textbf{auditory representations}}—especially from advanced speech models like Whisper—remains underexplored. This gap is significant, as auditory information is central to natural speech comprehension. Recent work by \citet{oota2023speech_LM_lack_imp_brain_relevant_semantics} elegantly shows that speech models, unlike text-based language models, capture brain activity patterns in auditory regions that cannot be explained by low-level stimulus features, underscoring the complementary nature of auditory and linguistic representations. Investigating how semantic and auditory features interact in the brain is therefore critical for advancing brain-aligned models of speech processing.

In this study, we address these gaps by introducing a nonlinear, multimodal encoding model that combines audio and semantic features extracted from advanced models like Whisper and LLAMA. Our contributions are as follows:

\begin{itemize}[itemsep=0pt, topsep=0pt, parsep=0pt]
    \item \textbf{Our nonlinear multimodal approach yields marked prediction performance improvements, showing a 17.2\% higher unnormalized correlation and 17.9\% higher normalized correlation compared to standard unimodal linear models \citep{huthscaling}, while surpassing previous state-of-the-art models—which rely on weighted averaging of linear unimodal predictions—by 7.7\% and 14.4\%, respectively (Appendix \ref{Appendix Stacked Regression}).} This performance boost demonstrates that incorporating nonlinearity and multimodality is crucial for capturing the brain's language processing mechanisms, promising more robust in-silico experiments and improved brain decoding capabilities.

    \item \textbf{We propose a novel spatiotemporal clustering analysis that tracks neural responses to semantic and auditory information over time, extending beyond traditional spatial-only approaches.} By analyzing relative error differences between semantic and audio encoding models, we demonstrate that nonlinear models achieve superior functional clustering compared to both linear encoders and standard connectivity analyses. This method reveals previously hidden patterns in brain organization, showcasing how nonlinear encoding models better capture the spatiotemporal dynamics of language processing.

    \item \textbf{We provide novel evidence for distributed multimodal processing in speech comprehension through variance partitioning and systematic comparisons of prediction accuracy when adding audio or semantic features}. Most brain regions utilize overlapping information from semantic and audio features, with neither modality strictly dominating. While both make unique contributions, their relative influence varies hierarchically from early sensory to higher-order areas. This extends existing neurolinguistic theories \citep{motortheoryofspeechperction_org_paper_1,nature_ref_2_convergence_zone_ORG,embodied} by revealing how different brain regions engage with multiple aspects of speech input.

\end{itemize}

\section{Method}
\subsection{MRI Data}
We used a publicly available fMRI dataset \citep{huthdatasetpaper, huthsemanticdecoding} of three subjects listening to approximately 20 hours of English podcast. The training data comprised ~95 stories across 20 scanning sessions (approximately 33,000 time points). For testing, we used three held-out stories: one averaged across ten repetitions and two averaged across five repetitions each, with no session containing repeated test stimuli. Each voxel was normalized to zero mean and unit variance across time, ensuring consistent training and testing data with \cite{huthscaling}.

\subsection{Feature Extraction} \label{Feature Extraction}
We extracted the features from the stimuli by taking the hidden layer representations of various LLMs and audio models exposed to the same stimuli as the subject. For semantic feature extraction, we utilized LLAMA-1 \citep{llama1} models with 7B, 13B, 33B, and 65B parameters, LLAMA-2 7B \citep{llama2}, and LLAMA-3 8B \citep{llama3}. For audio feature extraction, we employed Whisper \citep{whisper1} models, including Tiny, Base, Small, Large, and Large v2 and v3. All models were obtained from Hugging Face \citep{huggingface} and computed using half-precision (float16) for efficiency.

For LLAMA models, the stimuli were presented using a dynamically sized context window strategy \citep{huthscaling} to balance computational efficiency and contextual coherence (details are in Appendix \ref{appendix : llama featuree dynamical size window extraceiotn}). For Whisper models, the audio stimuli (waveform) were processed using a sliding-window approach with a fixed window size of 16 seconds and a stride of 0.1 seconds. Features were extracted exclusively from the encoder portion of the model, as it processes only the raw audio waveform input. This ensured that the extracted features accurately captured audio-specific representations relevant to the stimuli. Refer to \cite{huthscaling} for further details model contexts handling.

The following process adheres to  \cite{huthscaling} for fair comparison. We temporally aligned the hidden states from the $l^{\text{th}}$ layer of the language or audio models with fMRI acquisition times using Lanczos interpolation. To account for temporal delays between stimulus presentation and neural responses, we concatenated representations from the four preceding timepoints (2, 4, 6, and 8 seconds prior) for each fMRI acquisition timepoint, yielding a feature vector for each TR (see Appendix \ref{appendix aligning detail}). Unless stated otherwise, we extracted semantic features from the 12th layer of LLAMA-7B and audio features from the final encoder layer of Whisper Large V1. The layers were selected based on our findings that performance scaling with increasing LLM size, as reported in \cite{huthscaling}, does not hold for LLAMA models of size $\geq$7B (see Appendix \ref{Scaling does not work}). %

\subsection{Representations for fMRI Data} \label{PCA}
We predicted PCA-reduced fMRI representations, rather than the full voxel space, motivated by three benefits. First, PCA is a common dimensionality reduction method in fMRI analysis that helps prevent overfitting and has been widely applied in neuroimaging studies \citep{PCA_cls,PCA_cls_2,PCA_cls_3,PCA_cls_4}. This was crucial as speech comprehension engages the whole cortex and hence a vast number of voxels 80 -- 90k \citep{huthdatasetpaper}, far more than vision encoding $\approx$ 15k \citep{NSD_dataset}. In fact, linearly mapping the semantic stimulus (4 × 4096) to subject S1 require 1.3 billion parameters, whereas utilizing PCA (512 components) reduced this to 8.4 million, preventing overfitting. Second, PCA is effective at untangling the redundancy in brain data, as fMRI voxels are highly correlated. Studies have shown that masking up to 90\% of voxels does not significantly impact fMRI decoding or reconstruction performance \cite{redundancy_cortex_mental_state,redundancy_NIPS} , suggesting that information is distributed and redundant. Lastly, PCA allows us to recover the original voxel space from predicted PCA embeddings, maintaining the interpretability of the model’s predictions. %

In detail, we applied PCA to the aggregate fMRI response matrix $Y_{\text{org}} \in \mathbb{R}^{N_{\text{TR}} \times N_{\text{voxels}}}$, reducing its dimensionality to $Y_{\text{PCA}} \in \mathbb{R}^{N_{\text{TR}}\times N_{\text{PCA}}}$. $N_{\text{TR}},~N_{\text{voxels}}$ each refers to the number of time points (TRs) and voxels respectively, and $N_\text{PCA}$ was set to 512. The encoding model was trained to predict these PCA-reduced representations, and during evaluation, the predicted $\hat{Y}_{\text{PCA}}^{\text{test}}$ was reconstructed back to the original voxel space using inverse projection. This reconstructed output was then compared to the actual fMRI responses, $Y^\text{test} \in \mathbb{R}^{N_{\text{TR-test}}\times N_{\text{voxels}}}$.  More details are provided in Appendix \ref{appendix pca detail}.

\subsection{Encoding Model} \label{Encoding model}

Previous research primarily employed linear regression to predict voxel responses from unimodal features (audio or semantic) \citep{huthsemanticdecoding, encoding_model_2_huth2016, encoding_model_3_hierarchical, encoding_model_8_lebel2021voxelwise, encoding_model_4_jain2018incorporating, encoding_model_7_schrimpf2021neural}. Our study expands upon this approach by systematically investigating a range of encoding models varying in complexity and input modality to capture more nuanced relationships between stimulus representations and brain responses. We explored combinations of different stimulus representations, encoding model architectures, and response representations (as in Table \ref{tab:ALl results}). The following encoding architectures were used to assess the impact of complexity and nonlinearity (Details are in Appendix \ref{appendix other detail}): %

\begin{itemize}[itemsep=0pt, topsep=0pt, parsep=0pt]
\item \textit{Linear Regression (Linear):} Following \cite{huthscaling}, we used ridge regression. 
\item \textit{Multi-Layer Perceptron (MLP):} MLP with a single hidden layer of 256 units.
\item \textit{Multi-Layer Linear (MLLinear):} MLP but without dropout, batch normalization, and with the identity activation function. This model serves as a reduced-rank linear regression, helping to isolate the effects of dimensionality reduction from nonlinearity.
\item \textit{Delayed Interaction MLP (DIMLP):} Used for multimodal cases, this MLP variant processes each modality through separate 256-unit hidden layers before concatenation and final linear projection. This allows nonlinear processing within each modality while limiting cross-modal interaction to be linear, revealing the effects of nonlinear fusion of modalities.
\end{itemize}

\subsection{Noise Ceiling and Normalized Correlation Coefficient}
Due to the inherent noise in fMRI data, there is a theoretical upper limit to the amount of explainable variance an ideal encoding model can achieve, known as the noise ceiling. The noise ceiling for each voxel was estimated by applying an  existing method \citep{noiseceiling} on ten responses of the same test story (see Appendix \ref{appendix : noise ceiling and ccmax}). Afterwards, by dividing $CC_{abs}$, the correlation coefficient of the encoding model's prediction with the ground truth fMRI signals (estimated as the average of test responses to the same stimuli), by $CC_{max}$, we obtain $CC_{norm}$, the normalized correlation coefficient. Due to the large number of voxels ($\approx$ 80,000), random noise can cause certain voxels to have $CC_{abs} > CC_{max}$, resulting in $CC_{norm} > 1$. To prevent this, we regularized for noisy voxels by setting those with $CC_{max}< 0.25$ to 0.25 when computing $CC_{norm}$.

\subsection{RED (Relative Error Difference)}
We introduce a novel metric called the Relative Error Difference (RED). For each voxel $v$ at time $t$, $\text{RED}(v,t) = |f_1(v,t) - y(v,t)| - |f_2(v,t) - y(v,t)|$, where $f_1(v,t)$ and $f_2(v,t)$ are the predictions from model 1 (LLAMA) and model 2 (Whisper), respectively, and $y(v,t)$ is the true fMRI response. A positive RED value indicates that model 2 outperforms model 1 at that voxel and timepoint, while a negative value indicates the opposite.

RED extends beyond traditional voxel-wise analyses ($f(v)$) that focus only on spatial patterns of brain activity. By preserving temporal information ($f(v,t)$), RED enables analysis of both spatial and temporal dynamics of neural processing. We leverage this spatio-temporal information in Section \ref{nonlinear brainwide and spatiotemporal} to develop a novel approach for clustering ROIs based on their semantic and audio processing dynamics.

\begin{table}[h]
\caption{Performance of encoding models across different modalities and architectures (sem=semantic, aud=audio). This table presents the average voxelwise $r^2$ and normalized correlation coefficient ($CC_{norm}$) values for various encoding models, comparing their ability to predict fMRI responses across different input modalities \textit{semantic, audio} or \textit{multimodal} and encoder architectures \textit{Linear, MLLinear, DIMLP} and \textit{MLP}. Notably, MLP encoders consistently outperform linear models and their variants, highlighting the importance of incorporating nonlinearity for accurate fMRI prediction. $r^2$ is computed as $|r|*r$.}
\label{tab:ALl results}
\begin{center}
\footnotesize
\begin{tabular}{l@{\hspace{2mm}}l@{\hspace{2mm}}l@{\hspace{2mm}}l@{\hspace{2mm}}l}
\hline
\textbf{Input} & \textbf{Encoder} & \textbf{Output} & \multicolumn{1}{c}{\textbf{$r^2$ (\%)}} & \multicolumn{1}{c}{\textbf{$CC_{norm}$ (\%)}} \\ %
& & & \multicolumn{1}{c}{\textbf{(Δ$r^2$ \%)}} & \multicolumn{1}{c}{\textbf{(Δ$CC_{norm}$ \%)}} \\ %
\hline
sem+aud & MLP & PCA & \textbf{4.29 (+17.2)} & \textbf{34.32 (+17.9)} \\
sem+aud & DIMLP & PCA & 4.18 (+14.2) & 32.59 (+11.9) \\
sem+aud & MLLinear & PCA & 4.10 (+12.0) & 32.41 (+11.3) \\
sem+aud & Linear & voxels & 4.10 (+12.0) & 31.36 (+7.7) \\
sem+aud & Linear & PCA & 3.87 (+5.7) & 28.92 (-0.7) \\
sem+aud & MLP & voxels & 3.83 (+4.6) & 31.11 (+6.8) \\
\hline
sem & MLP & PCA & 3.79 (+3.6) & 30.89 (+6.1) \\
sem & MLLinear & PCA & 3.67 (+0.3) & 29.95 (+2.8) \\
sem & Linear & voxels & \underline{3.66 (base)} & \underline{29.12 (base)} \\
sem & Linear & PCA & 3.56 (-2.7) & 26.88 (-7.7) \\
sem & MLP & voxels & 3.36 (-8.2) & 27.45 (-5.7) \\
\hline
aud & MLP & PCA & 3.01 (-17.8) & 29.01 (-0.4) \\
aud & MLP & voxels & 2.89 (-21.0) & 28.21 (-3.1) \\
aud & MLLinear & PCA & 2.89 (-21.0) & 27.50 (-5.6) \\
aud & Linear & PCA & 2.81 (-23.2) & 26.71 (-8.3) \\
aud & Linear & voxels & 2.77 (-24.3) & 25.20 (-13.5) \\
\hline
\end{tabular}
\end{center}
\end{table}

\section{Results}

\subsection{Nonlinear Encoders}
\subsubsection{Nonlinearity, Not Reduced Dimensionality Alone, Improves Encoding Performance}

To examine the benefits of nonlinearity across feature hierarchies, we compared MLP and linear encoders across different layers of LLAMA and Whisper models (Figure \ref{Fig MLP layerwise} (Appendix)). Our findings reveal that MLP consistently outperformed linear models across all feature hierarchies and layer depths, supporting the notion that nonlinear transformations capture richer and more complex relationships in brain activity than linear mappings alone.

To disentangle the role of nonlinearity from dimensionality reduction, we compared the MLP encoder with two linear control models: ``Linear," which uses linear regression on PCA-reduced data, and ``MLLinear," which mirrors the MLP architecture without nonlinear activation functions. As shown in Table \ref{tab:ALl results}, both Linear and MLLinear models performed similarly to or worse than linear regression on the full voxel space (baseline model). 
These findings highlight the MLP's ability to capture nonlinear relations drives its superior performance, not merely dimensionality reduction. Additionally, PCA proves essential for leveraging nonlinearity. MLP models predicting all voxels directly performed poorly, likely due to overfitting from the large voxel space (80–90k voxels compared to 512 PCA components).

\begin{figure}[h]
\begin{center}
\includegraphics[width=\linewidth]{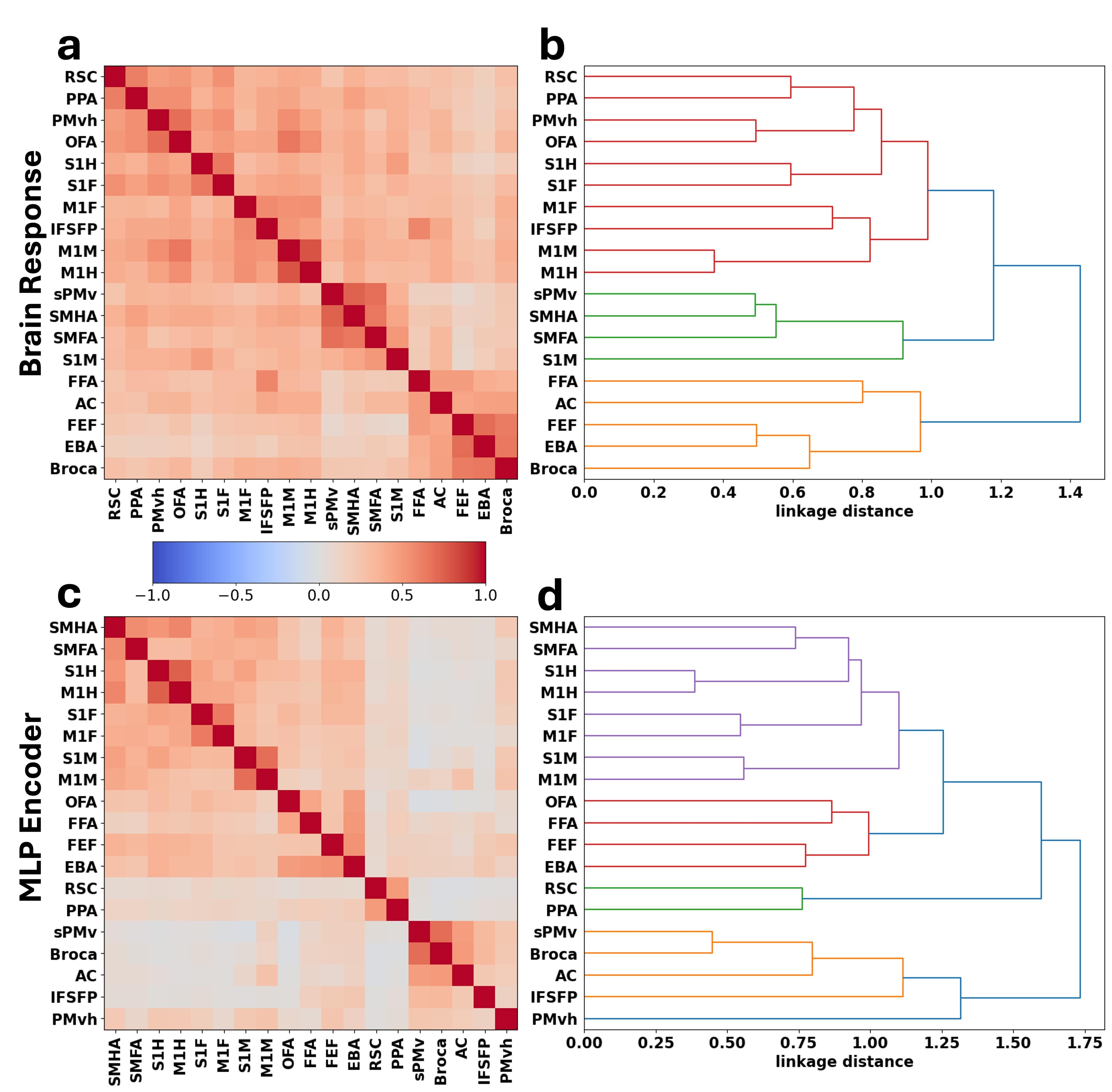}
\end{center}
\caption{Spatio-temporal clustering analysis: \textbf{(a,b)} Functional connectivity matrix and hierarchical clustering dendrogram from raw fMRI correlations. \textbf{(c,d)} Correlation matrices and dendrograms from Relative Error Difference (RED) between semantic and audio encoding models using MLP encoders. Matrix values indicate regional similarity. Hierarchical clustering reveals brain region organization by response profiles. The nonlinear models \textbf{(d)} show clearer functional groupings than standard connectivity \textbf{(b)}, quantified by higher modularity scores (see main text).}

\label{Fig main dendograph}
\end{figure}

\subsubsection{Nonlinearity Improves Brain-Wide Predictions and Spatio-Temporal Clustering}\label{nonlinear brainwide and spatiotemporal}

Nonlinear MLP models provide a crucial advantage over linear models by effectively capturing the complex nonlinear relationships inherent in brain activity during speech comprehension. Comparative brain maps in 
Appendix \ref{nonlinear improvement-CCnorm} 
illustrate the superior performance of MLP encoders over linear encoders, with improvements in prediction accuracy distributed across the cortex. The MLP model shows significant gains in regions associated with semantic and auditory processing, such as the medial prefrontal cortex (mPFC), precuneus (PrCu), and lateral temporal cortex (LTC). These gains highlight the  role of nonlinear interactions in accurately modeling brain activity, particularly in areas involved in higher-order language processing.

Furthermore, our hierarchical clustering analysis based on the Relative Error Difference (RED) between Whisper and LLAMA encoding models (Figure~\ref{Fig main dendograph} and Appendix~\ref{SpatioTemporal}) reveals two key insights: (1) RED enables functional clustering of brain regions where traditional functional connectivity fails, and (2) nonlinear encoders achieve superior functional grouping over linear ones, as shown by higher modularity $Q$ values (nonlinear: 0.155, linear: 0.145, FC: 0.068). The MLP encoder's clustering (Figure \ref{Fig main dendograph} (d)) demonstrates clear hierarchical functional organization: primary motor and somatosensory regions (M1/S1) first pair by body part (e.g., M1M/S1M) before forming broader motor and somatosensory clusters; higher visual regions cluster by function (face processing: OFA/FFA; scene processing and spatial navigation: PPA/RSC); and speech-language regions (sPMv/Broca/AC) form a cluster aligned with the dorsal stream pathway. This underscores RED’s potential and shows that nonlinear models capture structured spatiotemporal relationships in brain responses, aligning with known functional organization principles.

\begin{figure*}[t]
\begin{center}
\includegraphics[width=0.95\linewidth]{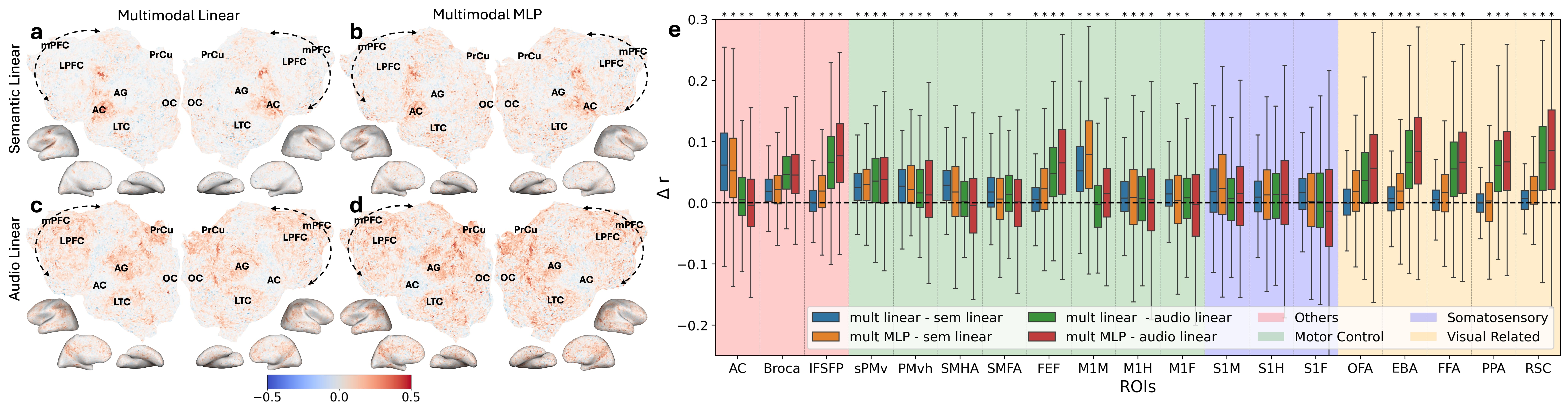}
\end{center}
\caption{Multimodality improvement in encoding models. 
Panels (a)-(d) display voxelwise $\Delta r$ values of a single subject, with warmer colors indicating regions where multimodal models outperform linear models. Each panel corresponds to the difference between voxel-wise predictions of the model in the corresponding column and the model in the corresponding row. E.g., panel (a) shows the difference between the Multimodal Linear and Semantic Linear models. (e) Box plot showing $\Delta r$ across different regions of interest (ROIs), where the $\Delta r$ values are aggregated over all subjects. \textit{mult} and \textit{sem} each refer to multimodal and semantic encoders. Asterisk* indicate ROI where $\Delta r > 0$ is statistically significant (p $< 0.05$).
ROIs are grouped and color-coded by their functions. The boxes represent the range between the 25th and 75th percentiles, with the line inside showing the median. Whiskers extend to 1.5 times this range. (A complete list of ROI abbreviations are at Appendix \ref{appendix abbreviations}. Voxelwise and ROI-wise plots for each subject are in Figure \ref{Fig all subs multimodal voxelwise}, \ref{Fig all subs multimodal voxelwise_2}, and \ref{Fig2 ROI subwise} in the Appendix).} %
\label{Fig multimodal voxelwise}
\end{figure*}

\subsection{Multimodal Encoders}

\subsubsection{Multimodality Predominantly Contributes to and Improves Brain-Wide Predictions}%

Improvements from multimodal encoding are cortex-wide and extend beyond modality-specific processing regions. Voxelwise analyses in Figure \ref{Fig multimodal voxelwise} (a,b) show that adding audio features improves not only auditory areas but also primary motor somatosensory regions. These improvements extend beyond expected auditory areas, with enhancements observed in the paracentral lobule, situated medially between the mPFC and Precuneus (PrCu), and in the occipital cortex (OC), reflecting the widespread impact of auditory information on cortical processing. Similarly, Figure \ref{Fig multimodal voxelwise} (c,d) shows that the addition of semantic features leads to improvements in most cortical areas, except certain parts of the auditory cortex (AC). These improvements are more pronounced in $CC_{norm}$ visualizations (Appendix \ref{multimodal improvement-CCnorm}), reinforcing the broad influence of semantic processing on neural activity beyond classical language areas.

This widespread improvement from multimodality is further amplified by nonlinearity. Comparing Figure \ref{Fig multimodal voxelwise} (b) with (a), and (d) with (c) reveals that MLP models not only enhance performance in regions that were already well predicted by multimodal linear encoders, but also in regions not initially benefiting from the added modality, such as the LTC when adding audio features, and LTC, mPFC, and OC when adding semantic features. This suggests multimodal MLP models can better exploit the additional modality through nonlinearity, which we discuss in Section \ref{sec:nonlinear interactions and multimodality}.

To understand the source of these improvements, we conducted variance partitioning analysis, decomposing each voxel's explained variance into modality-specific and joint components. Our analysis (Appendix \ref{VP various models}) reveals two key findings about brain-wide processing, with some regional exceptions such as the auditory cortex: first, the majority of explained variance comes from joint audio-semantic processing rather than from either modality alone; second, both semantic and audio features make unique contributions to explaining brain activity, though audio's unique contribution is much smaller in magnitude across most regions. These results suggest that, for the majority of cortical areas, auditory and semantic models contain largely overlapping information, with semantic models providing additional unique predictive information beyond what audio models capture.

By assigning each voxel to its dominant predictive modality, we found that joint audio-semantic features dominate cortical representations (Figure \ref{Fig variance partitinoing main figure} (a), shown for subject S1; plots for all subjects in Appendix \ref{App : Largest variance partitioning for each voxel}). This pattern is consistent across subjects, with semantic, audio, and joint features being the most attributable source for approximately 21.4\%, 10.1\%, and 68.5\% of significantly predicted voxels across the cortex, respectively, as shown by ROI-wise analysis in Figure \ref{Fig variance partitinoing main figure} (b) (subject-wise analyses in Appendix \ref{App Variance partitioning Venn diagram}).

Our findings align with and diverge from prior multimodal language studies. The cortex-wide improvement contrasts with \citep{huthscaling}, which reported localized enhancements in AC and M1M with auditory features. Methodological differences may explain this: First, they used multiple Whisper layers, potentially adding redundancy, whereas our approach focuses on the final layer. Second, their linear stacked regression model averaged predictions from separate encoders, limiting modality interaction, while our concatenation allows direct interaction (Appendix \ref{Appendix Stacked Regression}).

Our analysis also provides new insights into the nature of modality-specific representations. While our results align with \cite{oota2023speech_LM_lack_imp_brain_relevant_semantics} in that semantic models contain information beyond low-level features present in audio models, our results reveal a more nuanced picture - audio models, though showing smaller unique contributions, provide meaningful complementary information across multiple brain regions. This is evidenced by both improved prediction performance and non-zero unique variance contribution in our voxel-wise analysis. This apparent discrepancy may be because our voxel-wise approach might have  captured finer-grained patterns of unique audio contributions that might be averaged out in their ROI-level analyses.

These patterns of distributed joint processing align with the Convergence-Divergence-Zone theory \citep{nature_ref_2_convergence_zone_ORG}, which posits that semantic information is integrated from multiple modalities across the cortex.

\begin{figure*}[t]
\begin{center}
\includegraphics[width=0.95\linewidth]{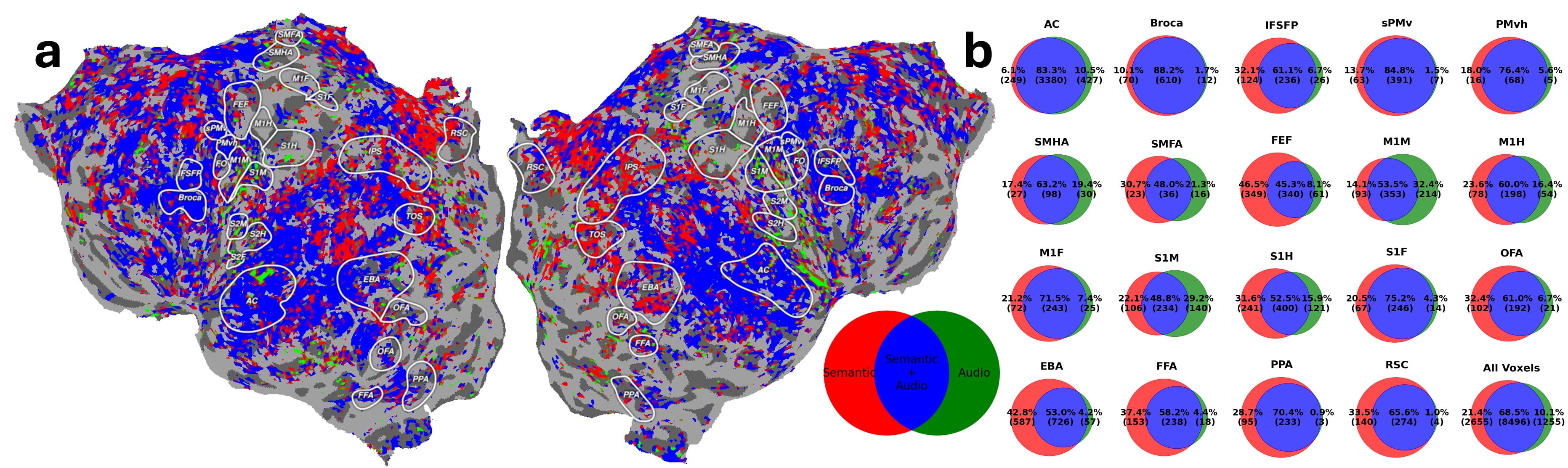}
\end{center}
\caption{Visualization of most dominant feature type in brain activity predictions from variance partitioning analysis. (a) Voxel-wise plots from a single subject (S1) and (b) ROI-wise Venn diagrams showing which feature type (semantic: red, audio: green, joint: blue) explains the largest variance for each significantly predicted voxel ($q(\text{FDR})<0.01$) using MLP encoders. ROI results are aggregated across subjects with numbers indicating voxel percentages and counts.
}
\label{Fig variance partitinoing main figure}
\end{figure*}

\subsubsection{Multimodal Fusion Supports and Extends Neurolinguistic Theories}
\label{Subsection:Multimodal motor and sensory area predictions}

Building on the brain-wide improvements observed, regions of interest (ROI) analyses reveal how multimodal integration supports and extends multiple neurolinguistic theories.

\noindent\textbf{Speech Related Regions (AC, Broca, sPMv, M1M)}

Our analysis reveals a systematic organization of speech processing that follows the auditory dorsal pathway, a key component of the dual-stream model of language processing \citep{dual_stream_model}. This pathway, extending from the auditory cortex (AC) through Broca’s area and sPMv to the primary motor cortex, exhibits distinct patterns of multimodal integration at each stage. 

In early AC, voxel-wise variance partitioning shows that unique contributions from audio features dominate (Figure \ref{Fig variance partitinoing main figure}), reinforcing its role in processing low-level acoustic information. However, processing in broader AC regions shows a shift to joint audio-semantic representations, with 83.3\% of significantly predicted voxels showing joint audio-semantic representation. The improved performance from adding auditory features (Figures \ref{Fig multimodal voxelwise} (a,b)) supports this hierarchical pattern, with earlier AC areas showing greater gains.

Moving along the dorsal pathway to Broca's area and sPMv (superior ventral premotor speech area), we find predominant joint feature attribution (88.2\% and 84.8\% of voxels respectively) with improved predictions from the addition of either modality. This multimodal integration aligns with these regions' role in speech planning and articulatory control—processes that require integrating acoustic targets with semantic content and motor programs \citep{wiki_borca_18,wiki_broca_17, encoding_model_3_hierarchical,sPMvMultimodal}.

At the terminus of the dorsal pathway, M1M shows a strong contribution from auditory features, exceeding even AC, consistent with its role in executing speech articulation (32.4\% of voxels) (Figure \ref{Fig variance partitinoing main figure} (b)). This strong auditory presence in motor areas is further supported by substantial performance improvements when adding auditory features, reinforcing previous findings from \cite{motor_theory_motor_auditory_interaction_review} that highlight the coupling between auditory and motor processes in speech production.

These findings extend our understanding of speech model representations. Our variance partitioning results align with previous findings that semantic models primarily predict AC activity by capturing low-level speech features \citep{oota2023speech_LM_lack_imp_brain_relevant_semantics}. Our analysis also reveal some voxels show unique semantic contributions, and audio models capture distinct brain features beyond the typical scope of language models. The observed semantic contribution in AC, sPMv and Broca's area aligns with prior findings \citep{encoding_model_3_hierarchical} and may be a general mechanism for language processing.

\noindent\textbf{Motor and Somatosensory Areas: Embodied Speech Processing}

The addition of audio or semantic features consistently improved predictions in cortical regions associated with motor control (green) and somatosensory processing (blue) (Figure \ref{Fig multimodal voxelwise} (e)). These improvements vary across ROIs: some benefit from the addition of semantic features (e.g., frontal eye field (FEF)), others from audio features (e.g., primary mouth motor cortex (M1M)), and some from both. Furthermore, variance partitioning analysis reveals that motor and somatosensory regions show unique contributions from both modalities - for instance, in M1M, audio features uniquely explain 32.4\% of the variance while semantic features explain 14.1\%, with 53.5\% jointly explained. Similar patterns emerge across motor areas (SMHA, SMFA, FEF, M1H, M1F) and somatosensory regions (S1M, S1H, S1F), suggesting these regions process unique auditory and semantic information absent from their overlapping features.

These findings align with the long-standing Motor Theory of Speech Perception \citep{motortheoryofspeechperction_org_paper_1, motortheoryofspeechperction_org_paper_2, poeppel2020speech}, which posits that motor regions actively participate in simulating the articulatory movements necessary for speech production, thereby aiding comprehension. In particular, improvements from the addition of and the unique contribution from auditory features align with research that discovered a tight coupling between auditory and motor-sensory processing \citep{motor_theory_audio_only_small_amount, motor_theory_motor_auditory_interaction_review, audio_motor_speech_sPMV}.

These findings also suggests that semantic information plays a critical role in shaping activity within somatosensory regions. This suggests a broader involvement of these areas in speech comprehension than previously recognized. This aligns with the concept of embodiment of semantic memory, where the understanding of concepts is grounded in sensory and motor experiences and their memory in the neocortex \citep{motor_theory_semantic_memory_all_review}. Our results align with \cite{motor_theory_ECoG_semantic_motor_nagata2022spatiotemporal}, who showed that the sensorimotor cortex is engaged in processing both concrete and abstract word semantics.

The enhancements in motor and sensory area predictions are more pronounced with MLP models, underscoring the importance of nonlinear interactions between auditory and semantic information. We explore this in more details in Section \ref{sec:nonlinear interactions and multimodality}. See Appendix \ref{Appendix : Improvements from multimodality} for subject-wise plots. 

\noindent\textbf{Higher-Order Visual Areas: Multimodal Semantic Representations}

Adding semantic features significantly enhances fMRI prediction accuracy in high-level visual areas like OFA \citep{OFA_whatitdoes}, EBA \citep{EBA_whatitdoes}, FFA \citep{FFA_whatitdoes}, PPA \citep{PPA_whatitdoes}, and RSC \citep{RSC_whatitdoes} (Figure \ref{Fig multimodal voxelwise}(e)). Variance partitioning (Figure \ref{Fig variance partitinoing main figure} (b)) shows that these ROIs have largest contributions from semantic and joint features, suggesting text-derived semantics provide substantial predictive information for visual regions beyond audio features alone.

This finding aligns with prior studies demonstrating that visual and linguistic stimuli with similar semantic content elicit similar brain responses \citep{huth_brain_multimodal,nature_ref_20_deniz_reading_linstening,huth_brain_multimodal_16_shared_semantic,huth_brain_multimodal_17_brain_amodal_semantic, huth_gallant_visual_border,nature_ref_18_huth_continuous,encoding_model_2_huth2016}. These studies, along with our results, support the convergence-divergence-zone theory \citep{huth_gallant_visual_border, nature_ref_11_convergence_zone_ORG,nature_ref_12_convergence_zone_ORG,nature_ref_2_convergence_zone_ORG}, which posits semantic information from multiple modalities is integrated at points across the cortex, leading to a unified representation of semantic meaning. This model suggests that the brain constructs a modality-independent representation of semantics, drawing on information from vision, language, and other senses \citep{motor_theory_semantic_memory_all_review,huthsemanticdecoding,huth_brain_multimodal,huth_brain_multimodal_16_shared_semantic,huth_brain_multimodal_17_brain_amodal_semantic,huth_brain_multimodal_18_brain_amodal_semantic}.

Our study also provides novel evidence for the auditory modality's contribution to this unified semantic representation. Variance partitioning (Figure \ref{Fig variance partitinoing main figure} (b)) shows that auditory information accounts for approximately 5\% of voxels in higher visual area ROIs. Adding audio features to our multimodal models resulted in a statistically significant performance increase in these ROIs (yellow) (Figure \ref{Fig multimodal voxelwise}(e)), suggesting auditory information, such as tone of voice and environmental sounds, may provide unique semantic context not fully captured by visual or linguistic features alone.

The consistent observation that multimodal fusion, particularly with nonlinear models, enhances prediction accuracy emphasizes the brain's use of complex, nonlinear computations to combine information from different modalities for a holistic understanding of language.  Subject-wise ROI prediction differences are visualized in Figure \ref{Fig2 ROI subwise} (Appendix).

\subsection{Nonlinear and Multimodal Encoders}
\label{sec:nonlinear interactions and multimodality}
\subsubsection{Nonlinear Interactions Between Modalities Enhance fMRI Predictions}
To assess the role of nonlinear cross-modal interactions, we developed a Delayed Interaction MLP (DIMLP), which processes audio and semantic features separately before a final linear fusion stage. This contrasts with MLP, which allows full nonlinear interactions across modalities. This design enables a direct comparison of within-modality nonlinearity (DIMLP) vs. cross-modal nonlinear interactions (MLP).

Both DIMLP and MLP outperform linear models (Table \ref{tab:ALl results}). DIMLP, incorporating only within-modality nonlinearity, yields a 2.0\% gain over the linear model (from 4.10\% average r² to 4.18\%). But the standard MLP, allowing full nonlinear interactions, achieves a further 2.6\% gain (from 4.18\% to 4.29\%). These results suggest that both forms of nonlinearity enhance encoding performance, but cross-modal nonlinear interactions contribute most significantly,

This conclusion is further supported by voxelwise analysis (Appendix \ref{DIMLP subwise appendix} and Figure \ref{Fig all subs nonlinear and multimodal voxelwise ccnorm} in Appendix). While DIMLP improves prediction accuracy across brain regions compared to the linear model, the transition to a standard MLP leads to further, cortex-wide enhancements. This suggests that nonlinear interactions between audio and semantic features are essential for modeling the complex, distributed neural representations underlying speech comprehension (see Appendix \ref{DIMLP subwise appendix}).

ROI-wise analysis (Appendix Figure \ref{Fig multimodal nonlinear}) shows regional variation in nonlinearity’s benefits. Multimodal MLP consistently matches or outperforms DIMLP and often surpasses linear models. Notably, motor (e.g., M1M) and somatosensory regions (e.g., S1M) benefit significantly from nonlinear cross-modal interactions, showing their role in complex multimodal processing during speech comprehension.

\section{Discussion and Conclusion}
This study underscores the transformative potential of nonlinear, multimodal encoding models for advancing our understanding of speech comprehension in the brain. By introducing a nonlinear Multi-layer Perceptron (MLP) and integrating audio and linguistic features, we achieved a 14.4\% increase in mean normalized correlation across the cortex compared to previous state-of-the-art \citep{huthscaling}, predicting 34.3\% of the brain's explainable variance. %

A key finding is that nonlinearity is fundamental to neural speech processing - nonlinear models outperformed linear approaches across all network layers, with improvements stemming from nonlinearity rather than dimensionality reduction alone as shown by linear control models. This cortex-wide enhancement reveals the brain's reliance on nonlinear computation, further supported by our novel RED analysis showing improved hierarchical clustering of brain regions, with higher modularity (0.155) than linear models (0.145) and traditional connectivity measures (0.068).

Our second key finding is that speech comprehension involves inherent multimodal fusion across the cortex. Adding either audio or semantic features improved predictions cortex-wide, while variance partitioning showed 68.5\% of significantly predicted voxels are best explained by joint audio-language processing rather than either modality alone. Through ROI-wise analyses of both variance partitioning and performance improvements, we provide support for key neurolinguistic theories including the Motor Theory of Speech Perception \citep{motortheoryofspeechperction_org_paper_1}, Convergence-Divergence Zone model \citep{nature_ref_2_convergence_zone_ORG}, and embodied semantics \citep{embodied}, highlighting the brain's reliance on distributed multimodal fusion.

Our nonlinear encoding approach has two main limitations. First, insufficient dataset size currently constrains model complexity, leading to overfitting when adding hidden layers or using RNNs and Transformers (Appendix \ref{appendix more complex models}). Given data scaling benefits in linear encoders \citep{huthscaling} and how a large dataset such as the Natural Scenes Dataset \citep{NSD_dataset} enabled deep learning breakthroughs in visual encoding and decoding \citep{trans_vision_encoding,mindseye_visiondecoding}, larger language fMRI datasets are needed to fully harness the potential of deep learning and drive further advancements. Second, while nonlinear encoders offer strong performance gains, they create new interpretability challenges. While variance partitioning and RED-based clustering offer preliminary insights, further innovations such as RSA \citep{RSA_paper} and novel feature attribution \citep{oota2023speech_LM_lack_imp_brain_relevant_semantics} are necessary. Moreover, nonlinear models offer unique interpretative possibilities, as shown by \citep{memory_vision_encoding} in memory vision encoding.

In conclusion, our study demonstrates that nonlinear, multimodal encoding models are crucial for understanding brain speech comprehension. Addressing dataset size and model interpretability limitations will be key to advancing brain aligned AI, enabling models that better reflect the hierarchical and distributed nature of neural processing. These insights have implications for neural representation learning, deep learning interpretability, and brain computer interfaces.

\section{Impact Statement}
This paper presents work whose goal is to advance the field of Neuroscience and Machine Learning. There are many potential societal consequences of our work, none which we feel must be specifically highlighted here.

\bibliography{example_paper}
\bibliographystyle{arxiv2025}

\newpage
\appendix

\section{Abbreviations of Brain Areas and Regions of Interest (ROIs)} \label{appendix abbreviations}
Brain Areas are abbreviated as follows : 
\begin{itemize}
    \item \textbf{AC}: Auditory Cortex
    \item \textbf{AG}: Angular Gyrus
    \item \textbf{LPFC}: Lateral Prefrontal Cortex
    \item \textbf{LTC}: Lateral Temporal Cortex
    \item \textbf{mPFC}: Medial Prefrontal Cortex
    \item \textbf{OC}: Occipital Cortex
    \item \textbf{PrCu}: Precuneus
\end{itemize}

The ROIs are abbreviated as follows : 
\begin{itemize}
    \item \textbf{AC}: Auditory Cortex
    \item \textbf{AG}: Angular Gyrus
    \item \textbf{Broca}: Broca's Area
    \item \textbf{EBA}: Extrastriate Body Area
    \item \textbf{FFA}: Fusiform Face Area
    \item \textbf{FEF}: Frontal Eye Field
    \item \textbf{IFSFP}: Inferior Frontal Sulcus Face Patch
    \item \textbf{LPFC}: Lateral Prefrontal Cortex
    \item \textbf{LTC}: Lateral Temporal Cortex
    \item \textbf{M1F}: Primary Motor Cortex - Foot
    \item \textbf{M1H}: Primary Motor Cortex - Hand
    \item \textbf{M1M}: Primary Motor Cortex - Mouth
    \item \textbf{mPFC}: Medial Prefrontal Cortex
    \item \textbf{OC}: Occipital Cortex
    \item \textbf{OFA}: Occipital Face Area
    \item \textbf{PMvh}: Ventral Premotor Hand Area
    \item \textbf{PPA}: Parahippocampal Place Area
    \item \textbf{PrCu}: Precuneus
    \item \textbf{RSC}: Retrosplenial Cortex
    \item \textbf{S1F}: Primary Somatosensory Cortex - Foot
    \item \textbf{S1H}: Primary Somatosensory Cortex - Hand
    \item \textbf{S1M}: Primary Somatosensory Cortex - Mouth
    \item \textbf{sPMv}: Superior Ventral Premotor Speech Area
    \item \textbf{SMFA}: Supplementary Motor Foot Area
    \item \textbf{SMHA}: Supplementary Motor Hand Area
\end{itemize}

\section{Details of Implementation} \label{appendix implementation detail}

\subsection{LLAMA Feature Extraction Strategy}\label{appendix : llama featuree dynamical size window extraceiotn}

LLAMA feature extraction was done in a dynamical window size manner for efficiency. Initially, the context window grew incrementally as tokens were added, up to a maximum of 512 tokens, after which the window was reset to a new context of 256 tokens. This approach avoided memory overheads associated with processing the entire tokenized text while maintaining sufficient contextual information for accurate semantic representation.

\subsection{Noise Ceiling ($CC_{max}$) and Normalized Correlation ($CC_{norm}$) Calculation}\label{appendix : noise ceiling and ccmax}
For each voxel, the maximum correlation  coefficient is estimated as $CC_{max} = (\sqrt{1+\frac{NP}{SP\times N}})^{-1}$, where $N$ is the number of repeats (10 in our case), $NP$ is the noise power or unexplainable variance, and $SP$ is the amount of variance that could be explained by an ideal predictive model.

\subsection{Resampling the hidden state of LLMs to fMRI time points } \label{appendix aligning detail}
After giving the language/audio model the same input as the subject, we temporally aligned the hidden states of its $l^{\text{th}}$ layer corresponding to a given $i^{\text{th}}$ token (last token of the $i^{\text{th}}$ word for language models), $H_l^i(S_{\{k | k\leq i\}})\in \mathbb{R}^{d_{\text{model}}^l}$ (aggregate shape of $\mathbb{R}^{N_{\text{token}}\times d_{\text{model}}^l}$ for the whole story where $N_{\text{token}}$ is the number of tokens/words), to the fMRI acquisition times (TR times) using Lanczos interpolation, obtaining an extracted feature of size $\mathbb{R}^{N_{\text{TR}}\times d_{\text{model}}^l}$, where $N_{\text{TR}}$ is the number of tokens (or number of words for language models) for each story and $d_{\text{model}}^l$ is the dimension of the $l^{\text{th}}$ hidden layer. We constructed the feature corresponding to a given $n^{\text{th}}$ TR ($2n$ seconds in physical time) by concatenating the representations from four previous TRs (2, 4, 6, 8 seconds before $t$ in physical time) to get a vector of shape $\mathbb{R}^{4d_{\text{model}}^l}$ for every $n^{\text{th}}$ TR, which we denote as ${H'}_l^{n}(S_{\{t | t\leq 2n\}})$. $H'$ denotes the additional resampling and concatenation done after applying the model, $H$. We used four previous time delays (2, 4, 6, 8 seconds) to account for the delay between the stimuli and brain response and to provide past stimuli information to the model. 

\subsection{Representations for fMRI response using PCA}
\label{appendix pca detail}
To an aggregate fMRI response, $Y_{\text{org}} \in \mathbb{R}^{N_{\text{TR}}\times N_{\text{voxels}}}$,  we applied PCA with 8192 maximum components along the voxel dimension using scikit-learn \citep{scikit-learn}, yielding an approximate projection matrix, $W\in \mathbb{R}^{N_{\text{voxels}}\times N_{8192}}$. Given $N_{\text{PCA}}$ number of principal components to consider, we take the top $N_\text{PCA}$ components to get $W_\text{PCA} \in \mathbb{R}^{N_{\text{voxels}}\times N_{\text{PCA}}}$, and train the encoding model to predict the reduced dimension PCA projection of the data, $Y_{\text{PCA}} = Y_{\text{org}}W_\text{PCA}\in \mathbb{R}^{N_{\text{TR}}\times N_{\text{PCA}}}$. 
During evaluation, the trained model outputs a reduced dimension representation of the data, $\hat{Y}_\text{PCA}^{\text{test}} \in \mathbb{R}^{N_{\text{TR-test}}\times N_{\text{PCA}}}$, where $N_{\text{TR-test}}$ denotes the number of timepoints (TRs) in the test story. This is reconstructed back the the original voxel space by applying an inverse of the projection matrix, $\hat{Y}^\text{test} = \hat{Y}_\text{PCA}^{\text{test}} W^T_{PCA} \in \mathbb{R}^{N_{\text{TR-test}}\times N_{\text{voxels}}}$, which is later compared with the ground truth, $Y^\text{test} \in \mathbb{R}^{N_{\text{TR-test}}\times N_{\text{voxels}}}$.

It should be noted that due to the high dimensionality of the data, incremental PCA was used, in place of regular PCA.

\subsection{Details of encoding models} \label{appendix other detail}

The encoding model architecture is as follows:

\begin{itemize}
\item \textit{Linear Regression (Linear):} Ridge regression. Following \cite{huthscaling}, ridge regression with bootstrapping ($n=3$) was used to estimate the optimal regularization parameters (alphas) for each voxel. The training data was divided into chunks of length 20, with 25\% used for held-out validation in each bootstrap iteration. The best alpha values were averaged across iterations, and the final model was trained on the full training dataset using these alphas.
\item \textit{Multi-Layer Perceptron (MLP):} MLP with a single hidden layer of 256 units, applying batch normalization and dropout to prevent overfitting. The hyperbolic tangent ($\tanh$) was used as the activation function.
\item \textit{Multi-Layer Linear (MLLinear):} MLP but without dropout, batch normalization, and with the identity activation function.
\item \textit{Delayed Interaction MLP (DIMLP):} MLP variant processes. Each modality through separate 256-unit hidden layers before concatenation and final linear projection.
\end{itemize}

We implemented encoding models using PyTorch. We employed the AdamW optimizer \citep{adamW} with a batch size of 128 and Mean Absolute Error (MAE) as the loss function to mitigate excessively penalizing random signal fluctuations. Our training regime consisted of 200 epochs with early stopping (patience = 10) based on validation loss, and we applied batch normalization with a momentum of 0.1. For robust evaluation, we implemented 5-fold cross-validation, averaging predictions across the five models for our final results. Hyperparameter optimization was conducted using Optuna \citep{optuna}, which performed 70 trials to determine optimal values for the dropout rate (0.1 to 0.3), learning rate ($10^{-5}$ to $10^{-1}$), and weight decay ($5 \times 10^{-5}$ to $10^{-1}$). 

Ridge regression was performed using a CPU node with 96 cores (Intel(R) Xeon(R) Gold 6240R CPU @ 2.40GHz) and 512 GB of RAM. Running the audio and language models and training encoding models was done using a GPU node with 8 H100 80GB GPUs.

\section{Comparison with stacked regression model of \cite{huthscaling}}\label{Appendix Stacked Regression}

\begin{table*}[ht!]
\caption{Comparing encoding performance across different models using the single test story evaluation protocol. Values show normalized correlation coefficient ($CC_{norm}$) and story-specific $r^2$ (\textbf{Avg $r^2$ (story)})(distinguishing from Table \ref{tab:ALl results}'s three-story evaluation (\textbf{Avg $r^2$})). SR refers to the previous state-of-the-art stacked regression model \citep{huthscaling}, which combines LLM and audio predictions through weighted averaging. Two masking approaches are used: 1) ``$\text{mask}_\text{A}$'' - their pre-computed validation-based voxel selection mask, and 2) ``mask'' - our computed masks that retain voxels showing validation improvements. For ``mask'', Linear+Mask indicates creating and applying a mask based on multimodal linear vs semantic linear performance, while MLP+Mask does the same using MLP models. $\text{semantic}_\text{A}$ denotes features from LLAMA-30B's 18th layer used in SR, while our models uses features from the 12th layer of LLAMA-7B. All approaches are evaluated using identical test data for fair comparison and $r^2$ is computed as $|r|*r$.}
\label{tab:appendix_results}
\begin{center}
\begin{tabular}{llllll}
\hline
\textbf{modality 1} & \textbf{modality 2} & \textbf{encoder} & \textbf{response} & \textbf{Avg $r^2$ (single story)} & \textbf{Avg $CC_{norm}$} \\
\hline
semantic & audio & MLP & PCA & \textbf{5.13\%} (+7.7\%) & \textbf{34.32\%} (+14.4\%)  \\
semantic & audio & MLP + mask & PCA &  5.02\% (+5.5\%)  & 33.33\% (+11.0\%) \\
semantic & audio & DIMLP & PCA &  4.93\% (+3.6\%)  & 32.59\% (+8.6\%) \\
semantic & audio & MLLinear & PCA &  5.00\% (+5.1\%)  & 32.41\% (+8.0\%) \\
semantic & audio & MLP + mask\_A & PCA &  4.77\% (+0.2\%)  & 31.70\% (+5.6\%) \\
semantic & audio & Linear & all voxels &  4.92\% (+3.4\%)  & 31.36\% (+4.5\%) \\
semantic & audio & MLP & all voxels &  4.54\% (-4.5\%)  & 31.11\% (+3.6\%) \\
semantic & audio & Linear + mask & all voxels &  4.90\% (+2.9\%)  & 31.09\% (+3.6\%) \\
semantic\_A & audio & SR + mask\_A & all voxels & \underline{4.76\% (Baseline)} & \underline{30.02\% (Baseline)} \\
semantic & audio & Linear & PCA & 4.48\% (-5.8\%) & 28.92\% (-3.7\%)  \\
\hline
semantic & - & MLP & PCA & 4.58\% (-3.7\%)  & 30.89\% (+2.9\%) \\
semantic & - & MLLinear & PCA & 4.59\% (-3.6\%)  & 29.95\% (-0.2\%) \\
semantic\_A & - & Linear & all voxels & 4.60\% (-3.3\%)  & 29.84\% (-0.6\%) \\
semantic & - & Linear & all voxels & 4.50\% (-5.4\%)  & 29.12\% (-3.0\%) \\
semantic & - & MLP & all voxels & 3.97\% (-16.6\%)  & 27.45\% (-8.6\%) \\
semantic & - & Linear & PCA & 4.15\% (-12.8\%)  & 26.88\% (-10.4\%) \\
\hline
audio & - & MLP & PCA & 3.83\% (-19.6\%)  & 29.01\% (-3.4\%) \\
audio & - & MLP & all voxels & 3.67\% (-22.8\%)  & 28.21\% (-6.0\%) \\
audio & - & MLLinear & PCA & 3.66\% (-23.1\%)  & 27.50\% (-8.4\%) \\
audio & - & Linear & PCA & 3.54\% (-25.6\%)  & 26.71\% (-11.0\%) \\
audio & - & Linear & all voxels & 3.46\% (-27.3\%)  & 25.20\% (-16.0\%) \\
\hline
\end{tabular}
\end{center}
\end{table*}

To establish the effectiveness of our nonlinear multimodal approach, we conduct a detailed comparison with the current state-of-the-art stacked regression model \citep{huthscaling}. Their method combines semantic and audio predictions through stacked regression followed by voxel-selection, where they decide what model to use (stacked regression or semantic linear) for each voxel based on a validation dataset. Their results are compared here and not in Table \ref{tab:ALl results} due to their use of only parts of the test stories as validation, barring computation of the ``\textbf{Avg $r^2$}'' value in Table \ref{tab:ALl results}. For accurate comparison, we obtain and use their published model weights and features.

The evaluation protocols differ specifically for the stacked regression (SR) model: while all models (including those in \citet{huthscaling}) primarily report performance using three test stories (Table~\ref{tab:ALl results}), SR uniquely requires using two of these test stories for validation-based voxel selection and only using the story ``wheretheressmoke'' for final testing. 

Also, following the identification of an error in the original evaluation protocol through community feedback, we corrected the methodology for fair comparison. Note that $\text{CC}_{\text{norm}}$ values remain consistent with Table \ref{tab:ALl results} as they were originally computed using only the ``wheretheressmoke'' story due to the unavailability of test repeats for the other two stories. 

To ensure fair comparison with SR, we additionally evaluate all models using their single-story protocol in Table~\ref{tab:appendix_results}, reporting both $\text{CC}_{\text{norm}}$ and a story-specific \textbf{Avg $r^2$ (single story)} metric to distinguish from our three-story evaluation. We found $\text{CC}_{\text{norm}}$ provides more stable comparisons than $r^2$ in this context, as the reduced number of timepoints (251 versus 790) makes $r^2$ more susceptible to noisy voxels compared to $\text{CC}_{\text{norm}}$ that accounts for these noisy voxels. This stability is reflected in the closer alignment between $\text{CC}_{\text{norm}}$ and $r^2$ rankings in Table~\ref{tab:ALl results} compared to Table~\ref{tab:appendix_results}. Therefore, we sort Table \ref{tab:appendix_results} with respect to the $\text{CC}_{\text{norm}}$.

Also, while their approach uses LLAMA-30B's 18th layer (denoted as $\text{semantic}_\text{A}$), we demonstrate competitive performance using LLAMA-7B features, consistent with our finding that encoding performance roughly plateaus beyond 7B parameters (Appendix \ref{Scaling does not work}). For comprehensive comparison, we implement both their pre-computed validation-based voxel selection mask (``$\text{mask}_\text{A}$'', created using an unspecified significance threshold) and our simpler approach (``mask'') that retains voxels showing any validation set improvement.  %

Table~\ref{tab:appendix_results} demonstrates several key results about our multimodal nonlinear approach. Our multimodal MLP achieves 34.32\% $\text{CC}_{\text{norm}}$ without masking, representing a 14.4\% improvement over the baseline stacked regression model, though the $\text{Avg }r^2\text{ (story)}$ improvement is more modest at 7.7\%. 

Our multimodal linear encoder also outperforms stacked regression by 4.5\%, supporting our hypothesis that direct concatenation enables more effective modality interaction compared to weighted averaging of unimodal predictions. The performance hierarchy (MLP $>$ Linear $>$ SR) suggests that both architectural choices - direct multimodal fusion and nonlinearity - contribute independently to improved predictions.

Interestingly, validation-based masking did not improve performance for either our linear or MLP models, regardless of whether using our mask or the precomputed $\text{mask}_\text{A}$ from previous work. This suggests our models learn effective feature selection implicitly, determining when to leverage or ignore audio features for specific voxels without explicit masking. The benefit of removing masking also likely stems from our models' ability to learn voxel-specific feature importance through direct access to input data, combined with the inherent noise in validation masks due to the limited number of timepoints.

These results demonstrate that enabling direct interaction between modalities through concatenation, combined with nonlinear processing, provides a more robust approach than previous methods relying on weighted averaging and explicit feature selection.

\begin{table*}[t]
\centering
\caption{Encoding performance of various nonlinear semantic encoders compared to other models. The table presents the average \( r^2 \) and normalized correlation coefficients (\( CC_{norm} \)) along with percentage changes relative to the baseline Linear model. Deep MLP refers to an MLP with two hidden layers, while MLP is an MLP with one hidden layer.}
\label{tab:deeper_models}
\begin{tabular}{llllll}
\hline
\textbf{modality 1} & \textbf{modality 2} & \textbf{encoder} & \textbf{response} & \textbf{Avg $r^2$} & \textbf{Avg $CC_{norm}$} \\
\hline
semantic & - & MLP & PCA & 3.79\% (+3.6\%) & 30.89\% (+6.1\%) \\
\textbf{semantic} & - & \textbf{Linear} & \textbf{all voxels} & \textbf{3.66\% (Baseline}) & \textbf{29.12\% (Baseline)} \\
semantic & - & LSTM & PCA & 3.33\% (-9.0\%) & 26.95\% (-7.46\%) \\
semantic & - & GRU & PCA & 3.21\% (-12.3\%) & 26.15\% (-10.2\%) \\
semantic & - & DeepMLP & PCA & 3.05\% (-16.7\%) & 27.45\% (-5.73\%) \\
semantic & - & RNN & PCA & 2.99\% (-18.0\%) & 25.42\% (-12.7\%) \\
semantic & - & Transformer & PCA & 2.82\% (-23.0\%) & 27.97\% (-3.95\%) \\
\hline
\end{tabular}
\end{table*}

\section{Results of more complex nonlinear models}\label{appendix more complex models}

We explored a range of more complex nonlinear models, as detailed in Table \ref{tab:deeper_models}. Specifically, we evaluated LSTM, GRU, RNN, and Transformer architectures, each configured with a single layer. The hidden dimensions for these models were determined by experimenting with sizes of 256, 512, 768, and 1024, selecting the dimension that yielded the best performance.

All models received inputs consisting of four timepoints, consistent with the MLP model, which concatenates these timepoints. For the recurrent models (LSTM, GRU, RNN), the final predictions were generated by applying a linear projection to a weighted pooling of the outputs corresponding to the four input timepoints. In the case of the Transformer model, we utilized learnable positional embeddings along with full self-attention mechanisms, and the final prediction was obtained by linearly projecting the output of the last token.

Additionally, we examined the DeepMLP model, an extension of the standard MLP with two hidden layers instead of one.

Our results indicate that while the MLP with a single hidden layer outperforms linear models, introducing greater complexity—such as recurrenct models or additional hidden layers—leads to overfitting and decreased performance.

\section{Performance of multimodal MLP model when mixing different layers}
\begin{figure}[h]
\begin{center}
\includegraphics[width=\linewidth]{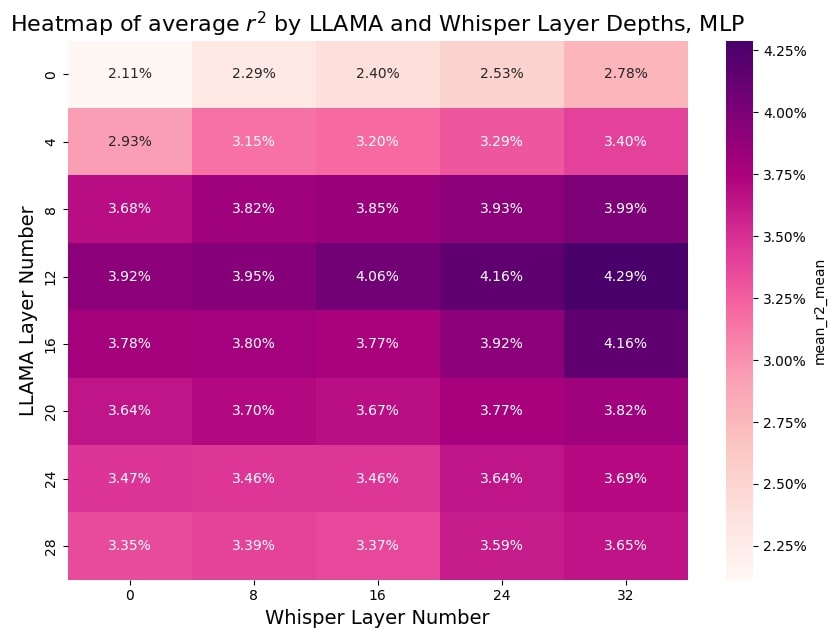}
\end{center}
\caption{Heatmap showing average $r^2$  values for different combinations of LLAMA and Whisper layer depths using an MLP encoder. Darker colors represent higher performance, with the best results obtained when the best layers in the respective uni-modal encoding models were used.}
\label{Multimodal heatmap}
\end{figure}

We observe in Figure \ref{Multimodal heatmap} that integrating the best performing layers from each modality results in the best performing multimodal model.

\section{Scaling LLM and audio models does not necessarily lead to better encoders} \label{Scaling does not work}
\begin{figure}[h]
\begin{center}
\includegraphics[width=\linewidth]{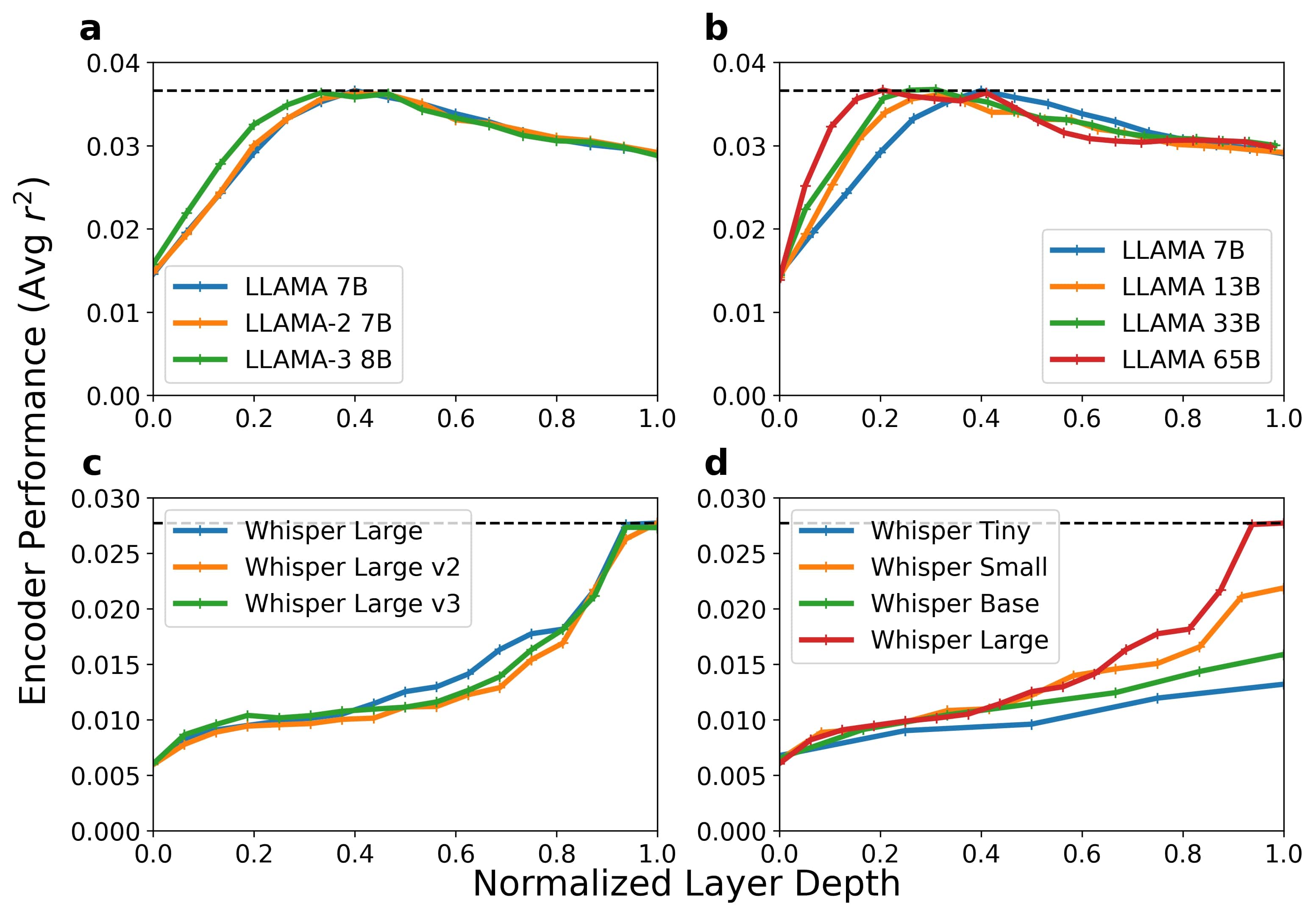}
\end{center}
\caption{Encoder performance across different LLAMA and Whisper model variants, using linear regression applied to the full set of voxels. Panel (a) compares LLAMA models of various architectures (LLAMA-2 and LLAMA-3) with 7B and 8B parameters. Panel (b) presents performance across different LLAMA models of increasing sizes, from 7B to 65B. Panels (c) and (d) show the performance for different Whisper model variants, including comparisons between Whisper Large versions (c) and different model sizes (d), from Whisper Tiny to Whisper Large. Performance is measured in terms of average \(r^2\), plotted against normalized layer depth.}
\label{Fig scaling}
\end{figure}

Previous research by \citet{huthscaling} found that increasing the size of large language models (LLMs) and audio models, such as scaling OPT from 125M to 175B parameters or Whisper from 8M to 637M parameters, enhanced encoding performance. However, performance gains plateaued for larger models like LLAMA-33B and OPT-175B, which they attributed to overfitting from larger hidden sizes.

Building on these findings, our study delves deeper into the scaling trends and offers a refined perspective on their implications for brain encoding models. For audio models, we confirm a positive correlation between model size and performance, as shown in Figure \ref{Fig scaling} (d). However, this scaling effect does not hold for language models. Specifically, LLAMA-7B, LLAMA-13B, LLAMA-33B, and LLAMA-65B exhibit comparable encoding performance, as shown in Figure \ref{Fig scaling} (b). This suggests diminishing returns beyond 7 billion parameters, a finding consistent with prior work by \citet{bonnasse2024fmri_scaling}, which reported performance plateaus for LLMs larger than 3 billion parameters.

We also evaluated the impact of scaling training data by examining newer versions of LLAMA and Whisper (e.g., LLAMA-1, LLAMA-2, LLAMA-3; Whisper v1, v2, v3). Despite larger datasets, newer versions did not yield significant performance improvements for either audio or semantic encoding models. This indicates that advancements in self-supervised learning (SSL) tasks, such as better next-token prediction, do not necessarily translate to more effective features for brain encoding. In essence, SSL improvements do not directly enhance brain-aligned representations.

In conclusion, our findings highlight two key points: (1) scaling language models beyond 7 billion parameters does not substantially improve encoding performance, and (2) increasing training data or using newer model versions does not enhance brain encoding feature extractors. These results challenge the assumption that simply scaling feature extractors, as proposed by \citet{huthscaling}, will lead to better encoding models.

\section{Context size speech models influence encoder performance}

\begin{figure}[h]
\begin{center}
\includegraphics[width=0.8\linewidth]{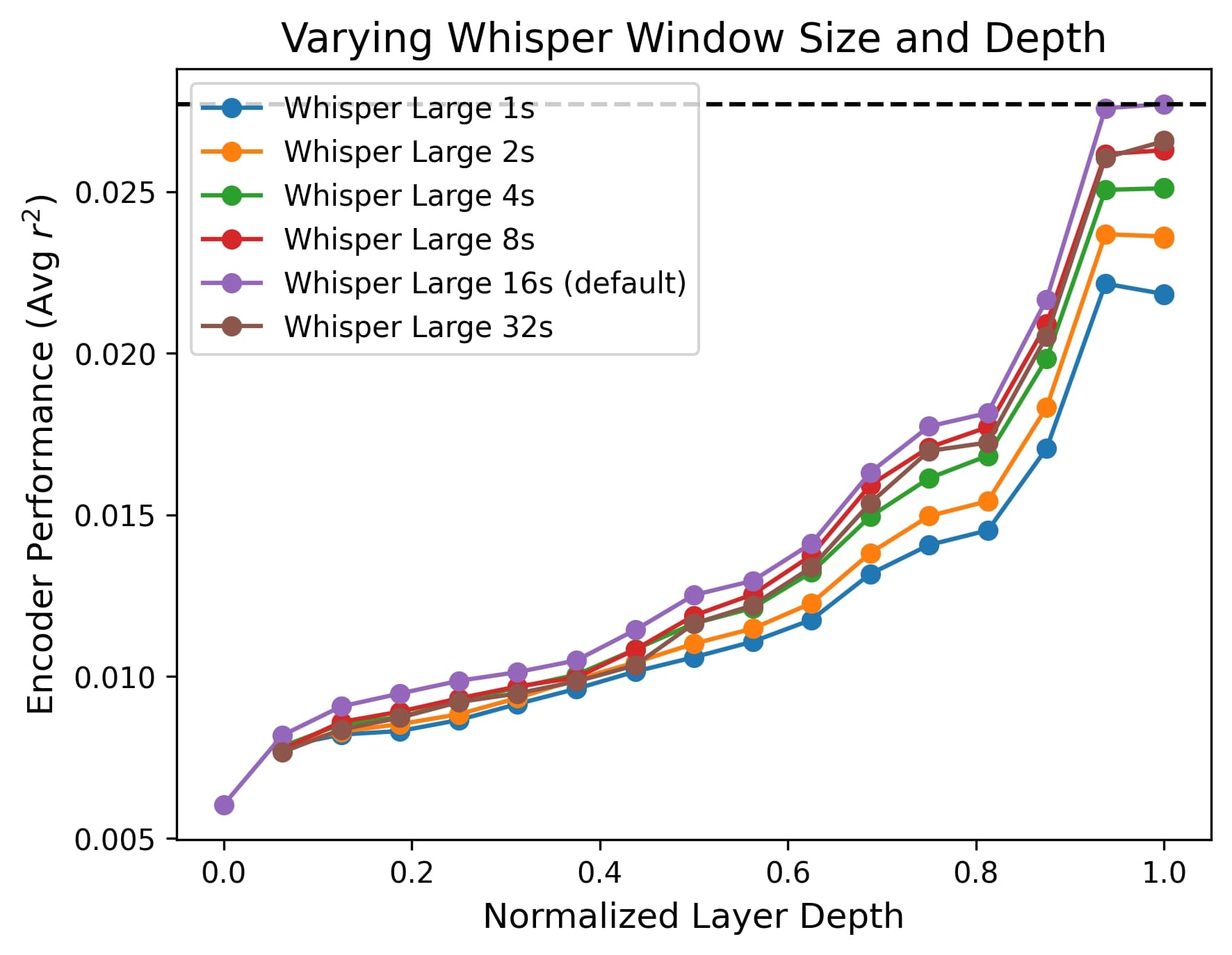}
\end{center}
\caption{Encoder performance across different Whisper Large models with varying window size, using linear regression applied to the full set of voxels.}
\label{Fig whisper stim size}
\end{figure}

Figure \ref{Fig whisper stim size} illustrates the impact of varying the context size (window size) of the Whisper model on encoding performance when using linear encoders, as explored in \cite{oota2023speech_LM_lack_imp_brain_relevant_semantics}. The results indicate that a 16-second window size, which was used as the default throughout our study, delivers the best performance. This outcome aligns with expectations, as the selected window size is consistent with the recommendations from \cite{huthscaling}.

\section{Performance of various encoding models using different inputs}

\subsection{Voxelwise $r$ values from different encoding mdoels and stimuli}
\begin{figure}[h]
\begin{center}
\includegraphics[width=\linewidth]{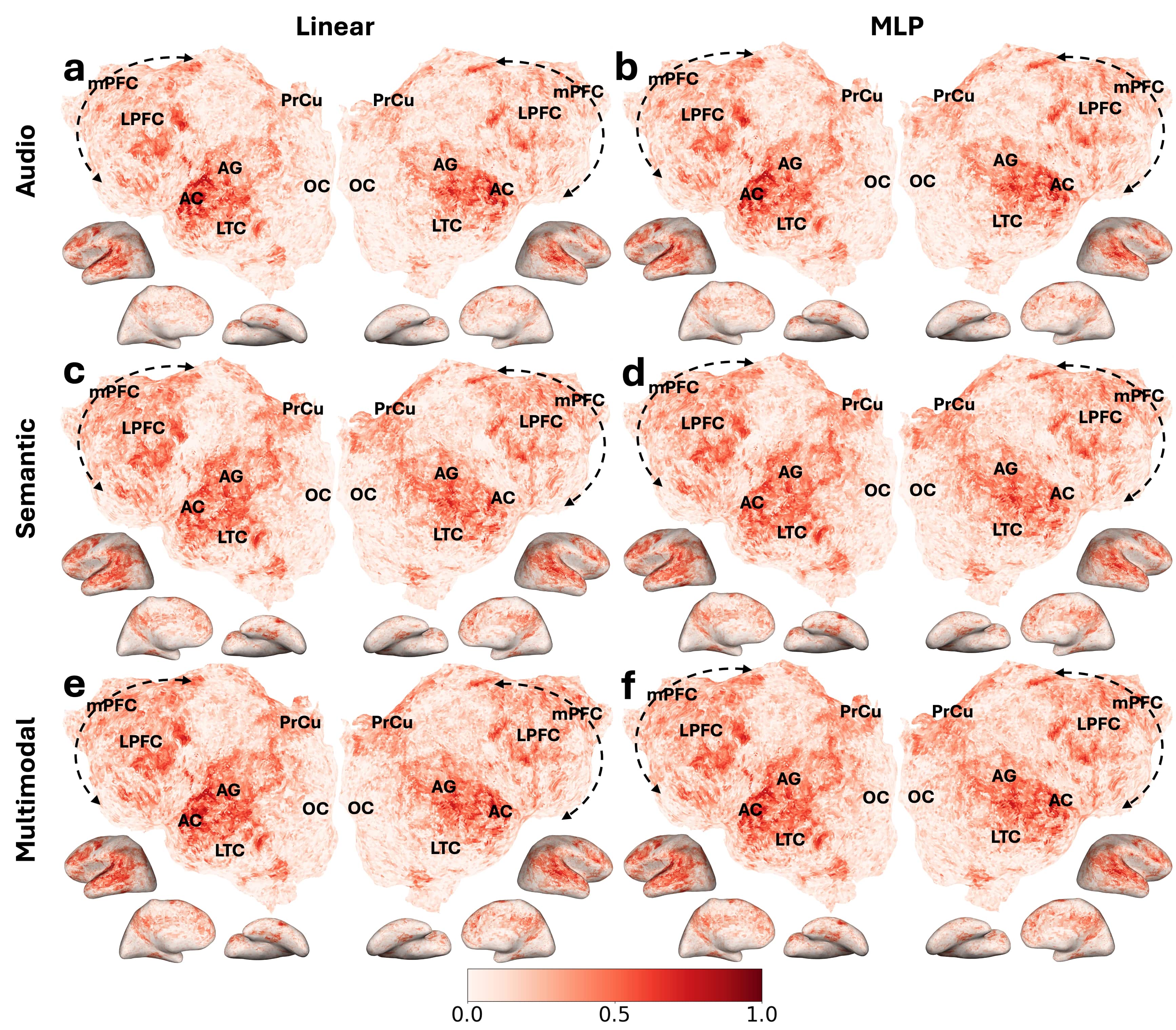}
\end{center}
\caption{Voxelwise $r$ values for Subject S1 across different input modalities and encoding models. Rows show audio-only (a,b), semantic-only (c,d), and multimodal (e,f) inputs. Columns compare Linear (left) and MLP (right) encoders. Warmer colors indicate higher prediction accuracy.}
\label{Fig r2-voxwise S1}
\end{figure}

Figures \ref{Fig r2-voxwise S1} represent the voxelwise correlation ($r$) values using various encoders and inputs for subject S1. Due to file size constraints, the plots for other subjects have been moved to the supplementary materials.

\subsection{ROI-wise $r$ values from different encoding models and stimuli}

\begin{figure*}[t]
\begin{center}
\includegraphics[width=\linewidth]{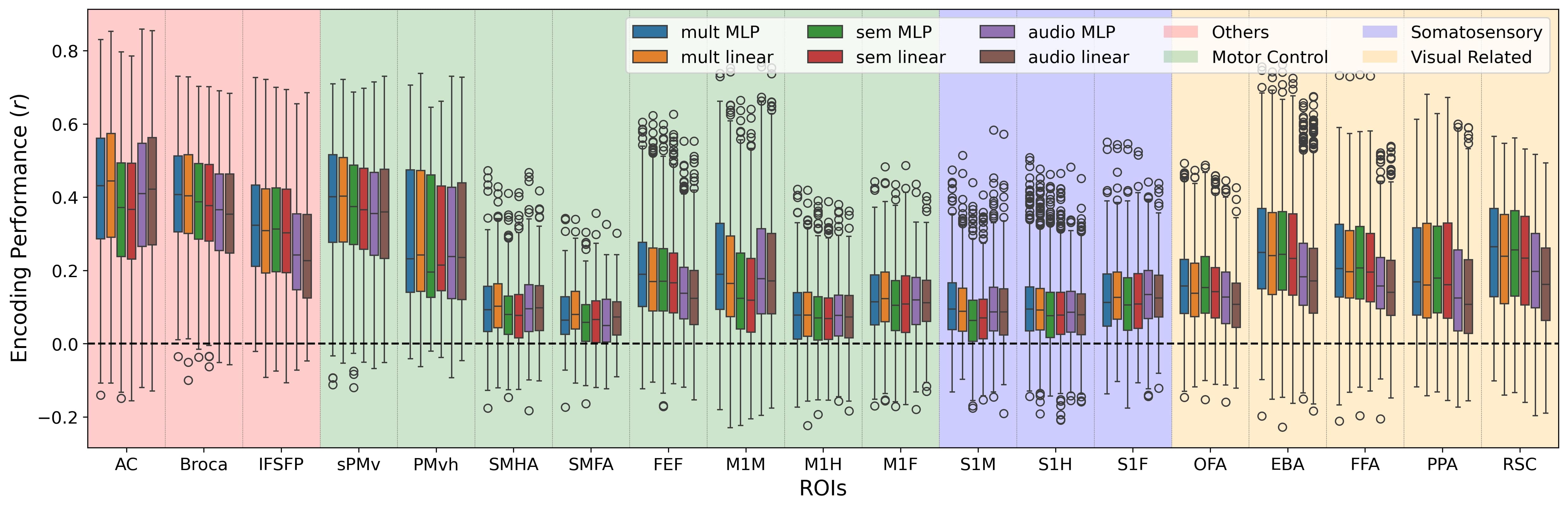}
\end{center}
\caption{Box plot showing $r$ across different regions of interest (ROIs), where the $r$ values are aggregated over all subjects. \textit{multi} refers to multimodal, and \textit{sem} refers to semantic encoders. ROIs are grouped and color-coded by their functions. }
\label{Fig roiwise rvalue}
\end{figure*}

Figure \ref{Fig roiwise rvalue} shows the $r$ value for different encoding models and stimuli averaged across subjects. 

\FloatBarrier

\section{Improvements from nonlinearity}
\subsection{Layer-wise performance increases from MLP}
\begin{figure}[h]
\begin{center}
\includegraphics[width=\linewidth]{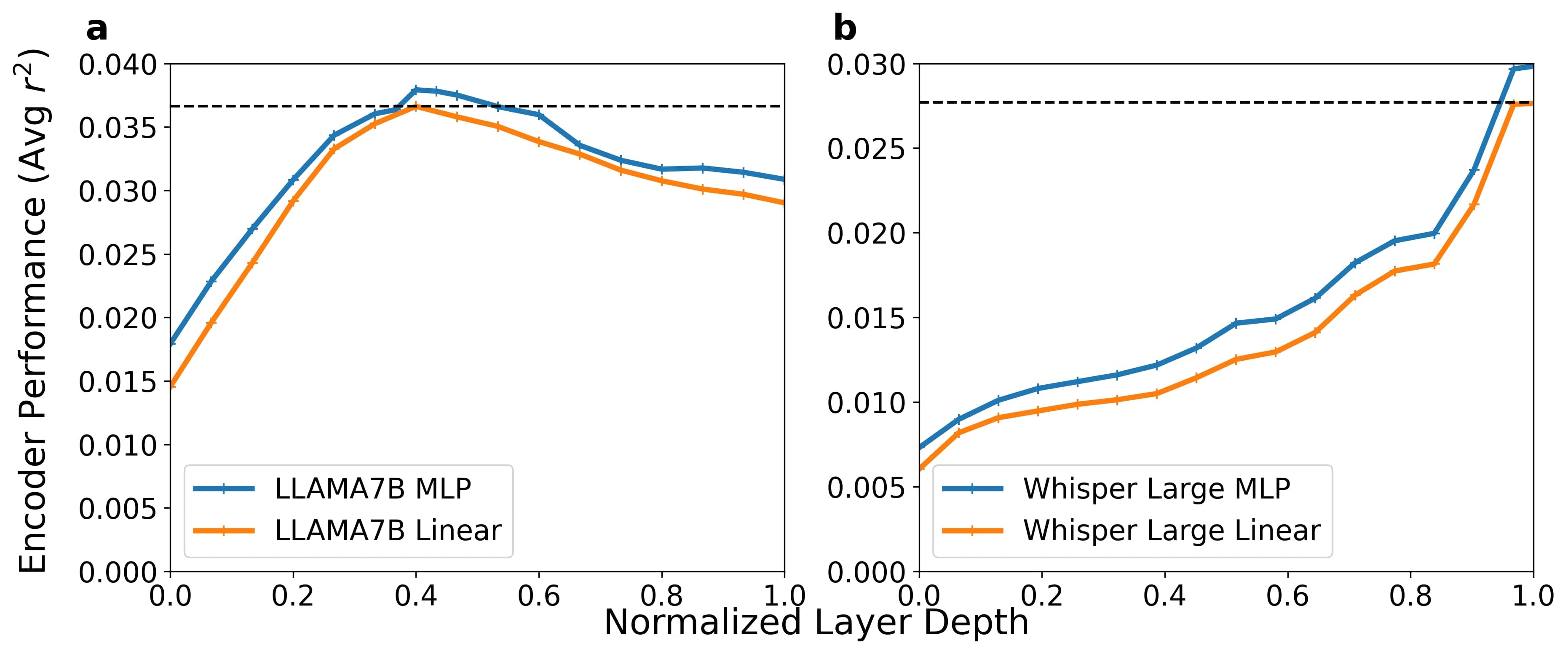}
\end{center}
\caption{Average voxel-wise $r^2$ values, computed as the mean across three subjects, for each layer of the (a) language (LLAMA7B) and (b) audio (Whisper Large) models. Comparisons are shown between the MLP and linear encoders, and dashed black lines indicate the best performance for linear encoders}
\label{Fig MLP layerwise}
\end{figure}

Figure \ref{Fig MLP layerwise} shows that MLP improves encoding performance for both language and audio models, regardless of what layer is used for the MLP encoding model.

\subsection{Voxelwise improvements from MLP ($r$ analysis)}\label{nonlinear improvement-r}
\begin{figure}[h]
\begin{center}
\includegraphics[width=\linewidth]{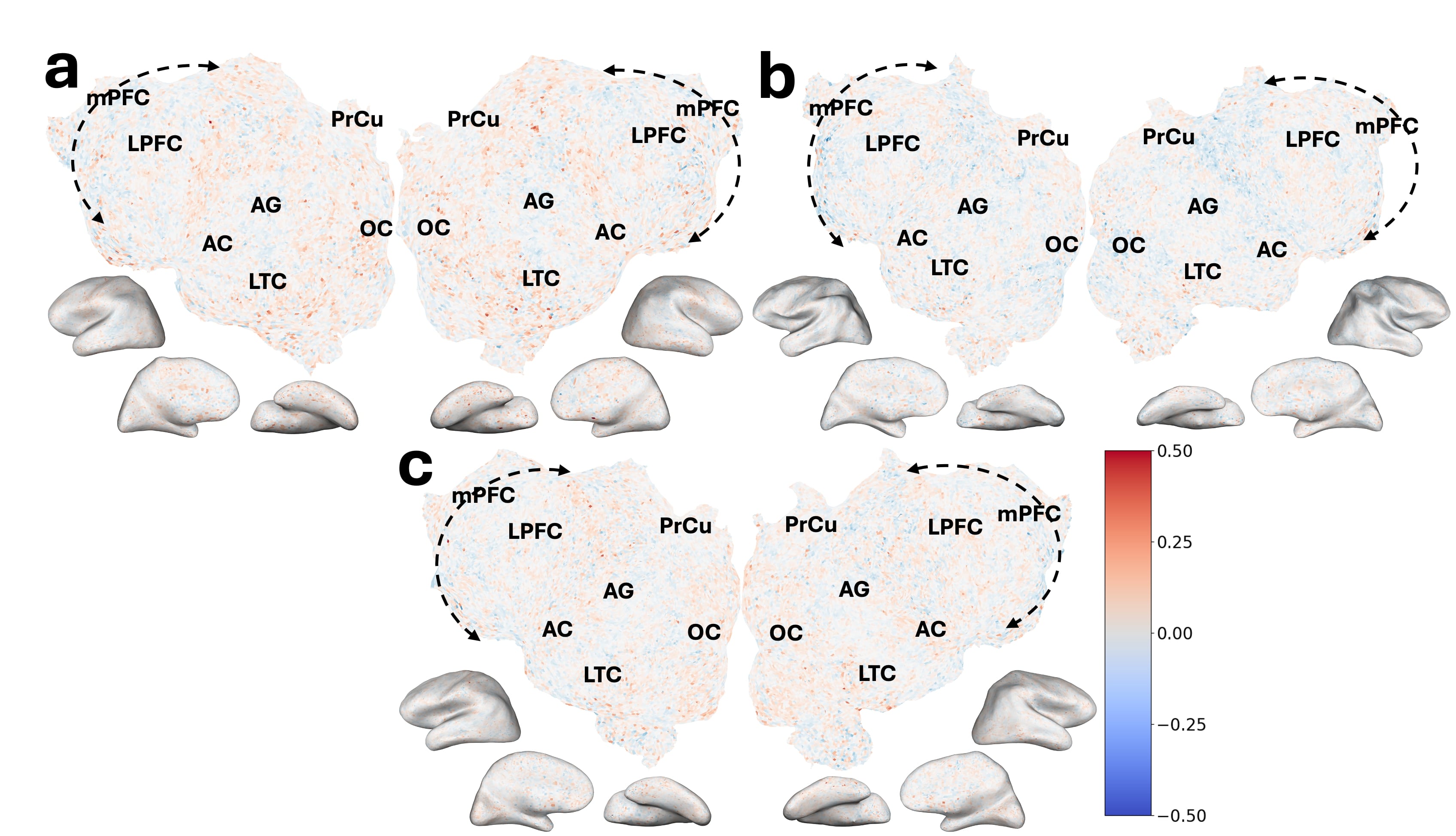}
\end{center}
\caption{Encoding model performance improvements. (a-c) Voxelwise $\Delta r$ (MLP performance minus linear performance) for semantic input for subjects S1, S2, S3, respectively. Positive values indicate MLP outperformance.}
\label{Fig nonlinear voxelwise semantic}
\end{figure}

\begin{figure}[h]
\begin{center}
\includegraphics[width=\linewidth]{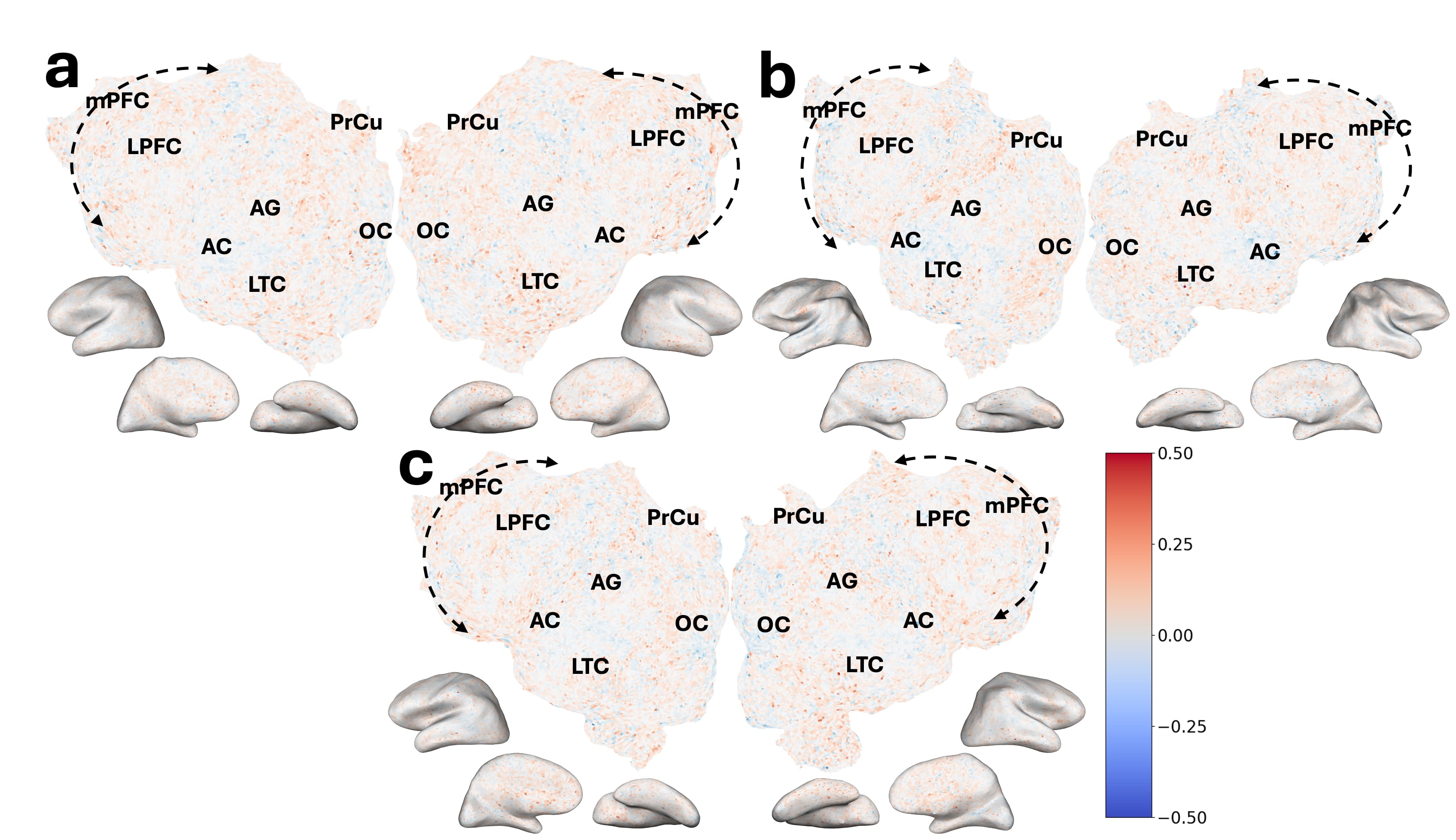}
\end{center}
\caption{Encoding model performance improvements. (a-c) Voxelwise $\Delta r$ (MLP performance minus linear performance) for audio input for subjects S1, S2, S3, respectively. Positive values indicate MLP outperformance.}
\label{Fig nonlinear voxelwise audio}
\end{figure}

\begin{figure}[h]
\begin{center}
\includegraphics[width=\linewidth]{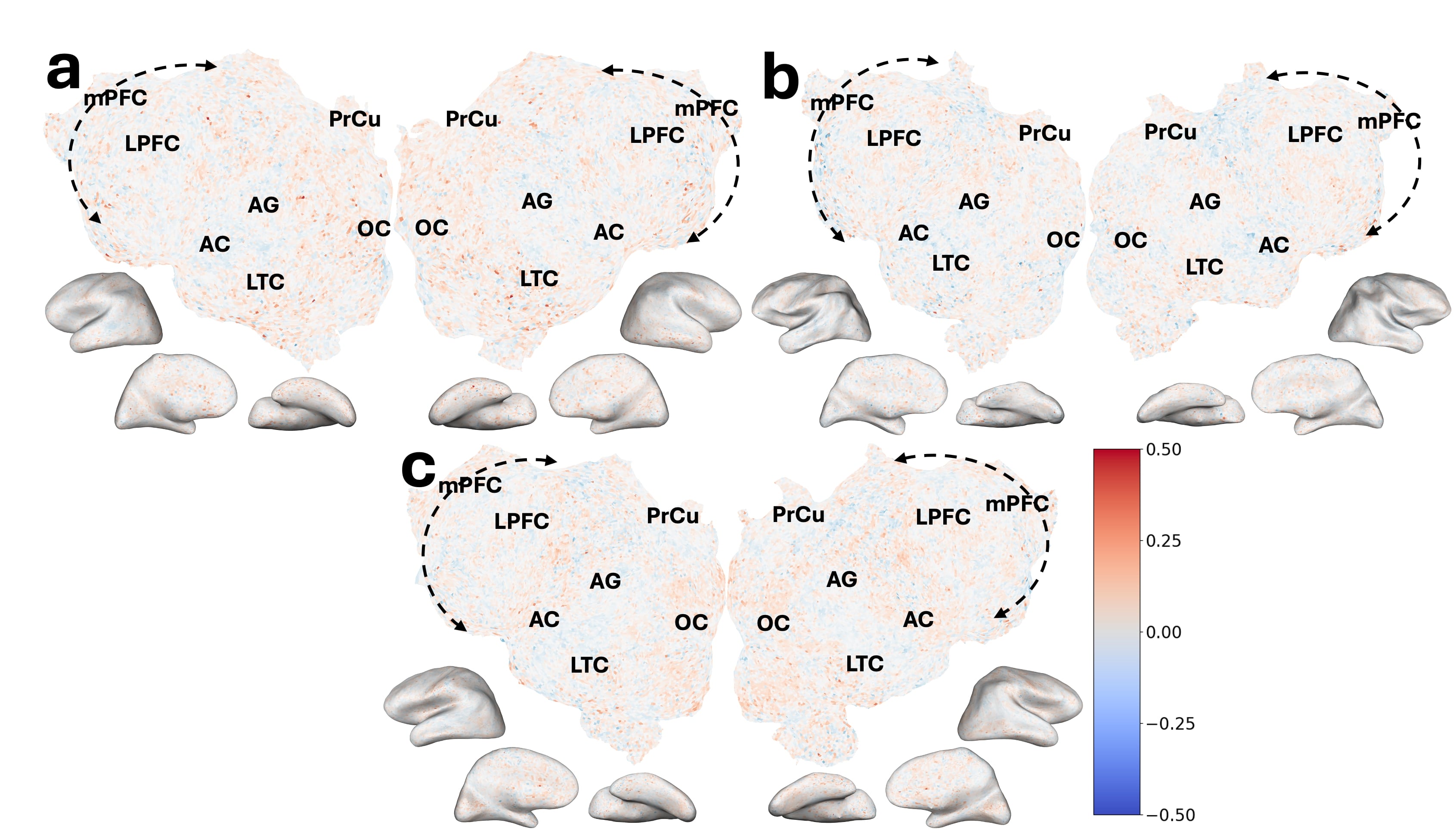}
\end{center}
\caption{Encoding model performance improvements. (a-c) Voxelwise $\Delta r$ (MLP performance minus linear performance) for multimodal input for subjects S1, S2, S3, respectively. Positive values indicate MLP outperformance.}
\label{Fig nonlinear voxelwise multi}
\end{figure}

Figures \ref{Fig nonlinear voxelwise semantic}, \ref{Fig nonlinear voxelwise audio}, and \ref{Fig nonlinear voxelwise multi} each represent the performance improvements in voxelwise correlation values for semantic, audio, and multimodal inputs, respectively, for each subject.

\FloatBarrier

\subsection{Voxelwise improvements from MLP ($CC_{norm}$ analysis)}\label{nonlinear improvement-CCnorm}

\begin{figure}[h]
\begin{center}
\includegraphics[width=\linewidth]{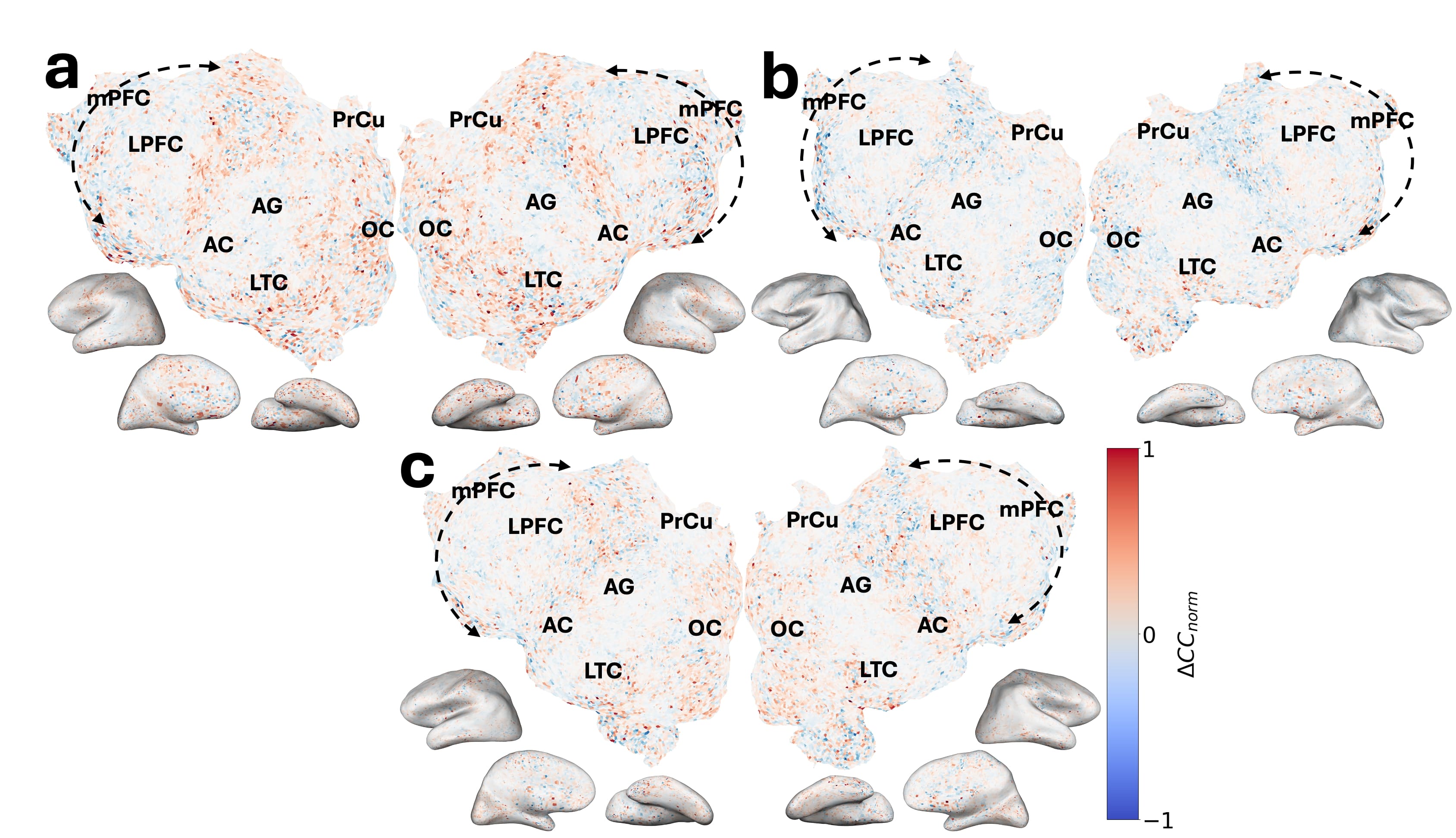}
\end{center}
\caption{Encoding model performance improvements. (a-c) Voxelwise $\Delta CC_{norm}$ (MLP performance minus linear performance) for semantic input for subjects S1, S2, S3, respectively. Positive values indicate MLP outperformance.}
\label{Fig nonlinear voxelwise semantic ccnorm}
\end{figure}

\begin{figure}[h]
\begin{center}
\includegraphics[width=\linewidth]{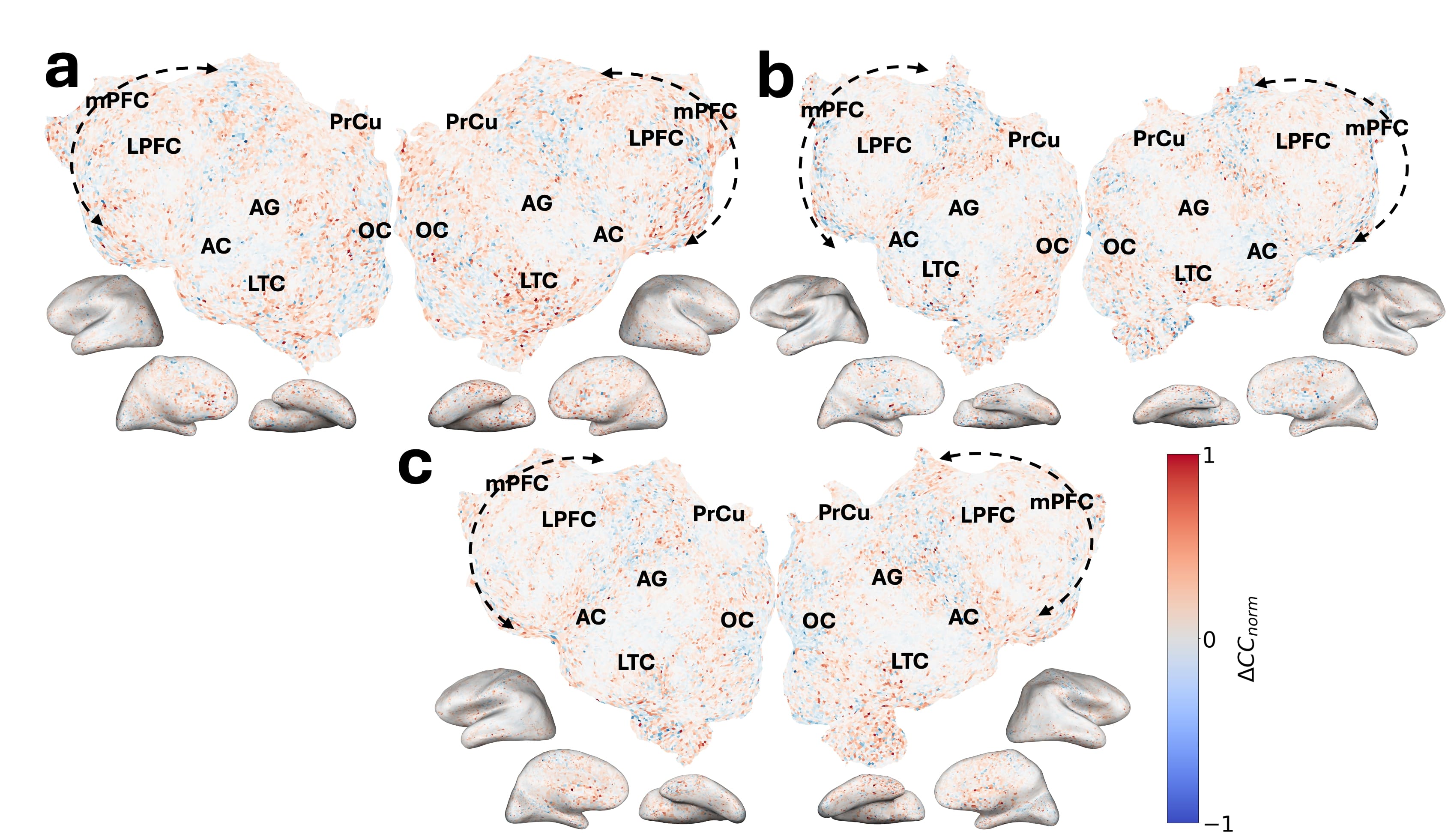}
\end{center}
\caption{Encoding model performance improvements. (a-c) Voxelwise $\Delta CC_{norm}$ (MLP performance minus linear performance) for audio input for subjects S1, S2, S3, respectively. Positive values indicate MLP outperformance.}
\label{Fig nonlinear voxelwise audio ccnorm}
\end{figure}

\begin{figure}[h]
\begin{center}
\includegraphics[width=\linewidth]{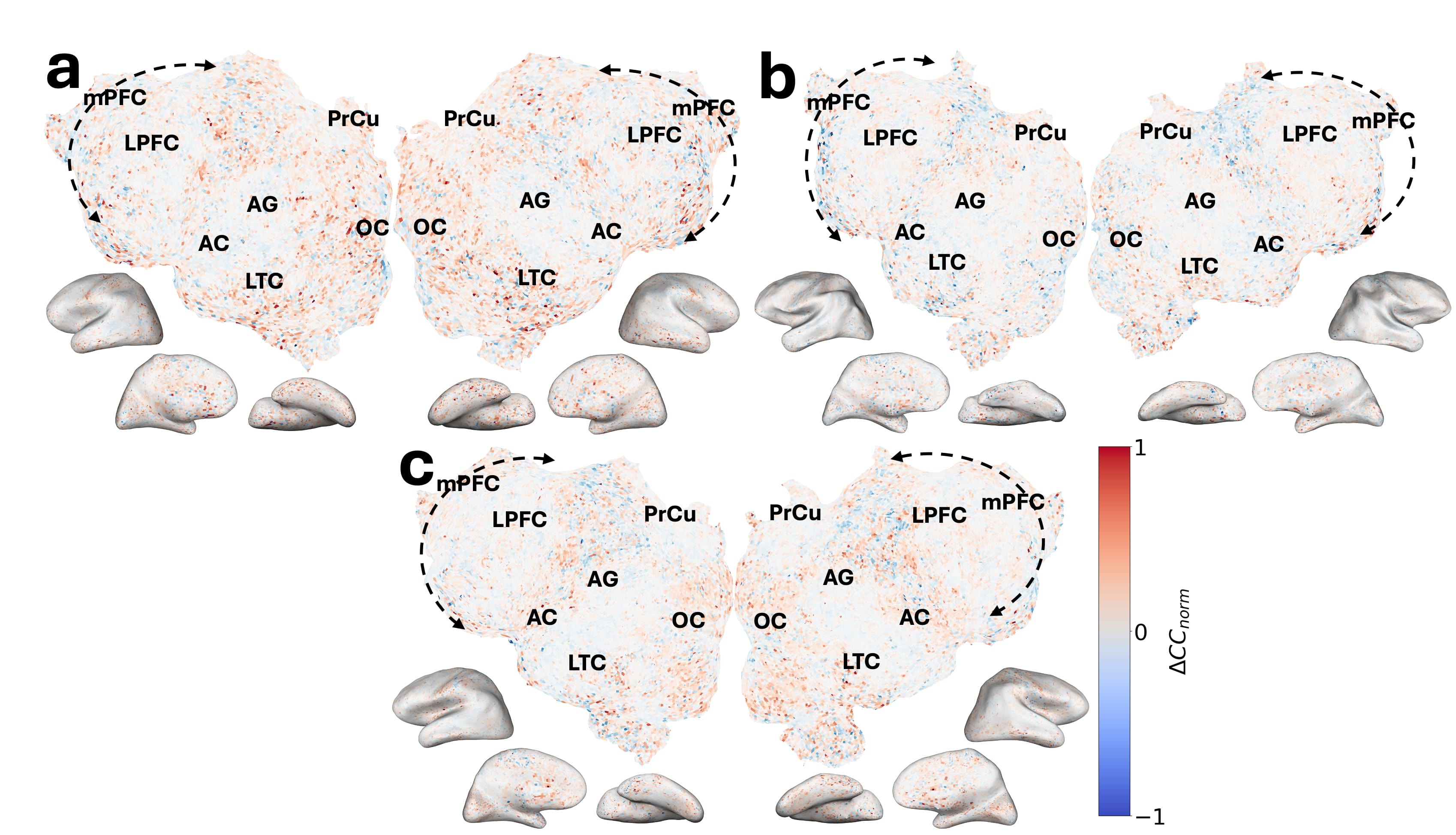}
\end{center}
\caption{Encoding model performance improvements. (a-c) Voxelwise $\Delta CC_{norm}$ (MLP performance minus linear performance) for multimodal input for subjects S1, S2, S3, respectively. Positive values indicate MLP outperformance.}
\label{Fig nonlinear voxelwise multi ccnorm}
\end{figure}

Figures \ref{Fig nonlinear voxelwise audio ccnorm}, \ref{Fig nonlinear voxelwise semantic ccnorm}, and \ref{Fig nonlinear voxelwise multi ccnorm}  each represent the performance improvements in voxelwise $CC_{norm}$ values for semantic, audio, and multimodal inputs, respectively, for each subject. The improvements are more pronounced with $CC_{norm}$ compared to $r$ as noise is taken into account. 

\FloatBarrier

\subsection{Better spatio-temporal compartmentalization of brain function}\label{SpatioTemporal}

To compare the performance between Whisper and LLAMA models, we define the Relative Error Difference (RED) for each voxel \( v \) at time \( t \) as:

$$
\text{RED}(v,t) = \left| f_{\text{semantic}}(v,t) - y(v,t) \right| - \left| f_{\text{audio}}(v,t) - y(v,t) \right|
$$

where \( f_{\text{semantic}}(v,t) \) is the prediction from the semantic encoding model for voxel \( v \) at time \( t \), \( f_{\text{audio}}(v,t) \) is the prediction from the audio encoding model for voxel \( v \) at time \( t \), and \( y(v,t) \) represents the true value at voxel \( v \) and time \( t \). A positive RED value indicates that the audio model outperforms the semantic model at that specific voxel and time, while a negative value indicates that the semantic model performs better.

In this analysis, we computed the RED between Whisper and LLAMA models for each voxel $v$ at a given time $t$. For each region of interest (ROI), the average RED is calculated as:

$$
\text{RED}_{\text{ROI}}(t) = \frac{1}{N} \sum_{{v \in \text{ROI}}} \text{RED}(v,t)
$$

\begin{figure*}[t]
\begin{center}
\includegraphics[width=0.9\linewidth]{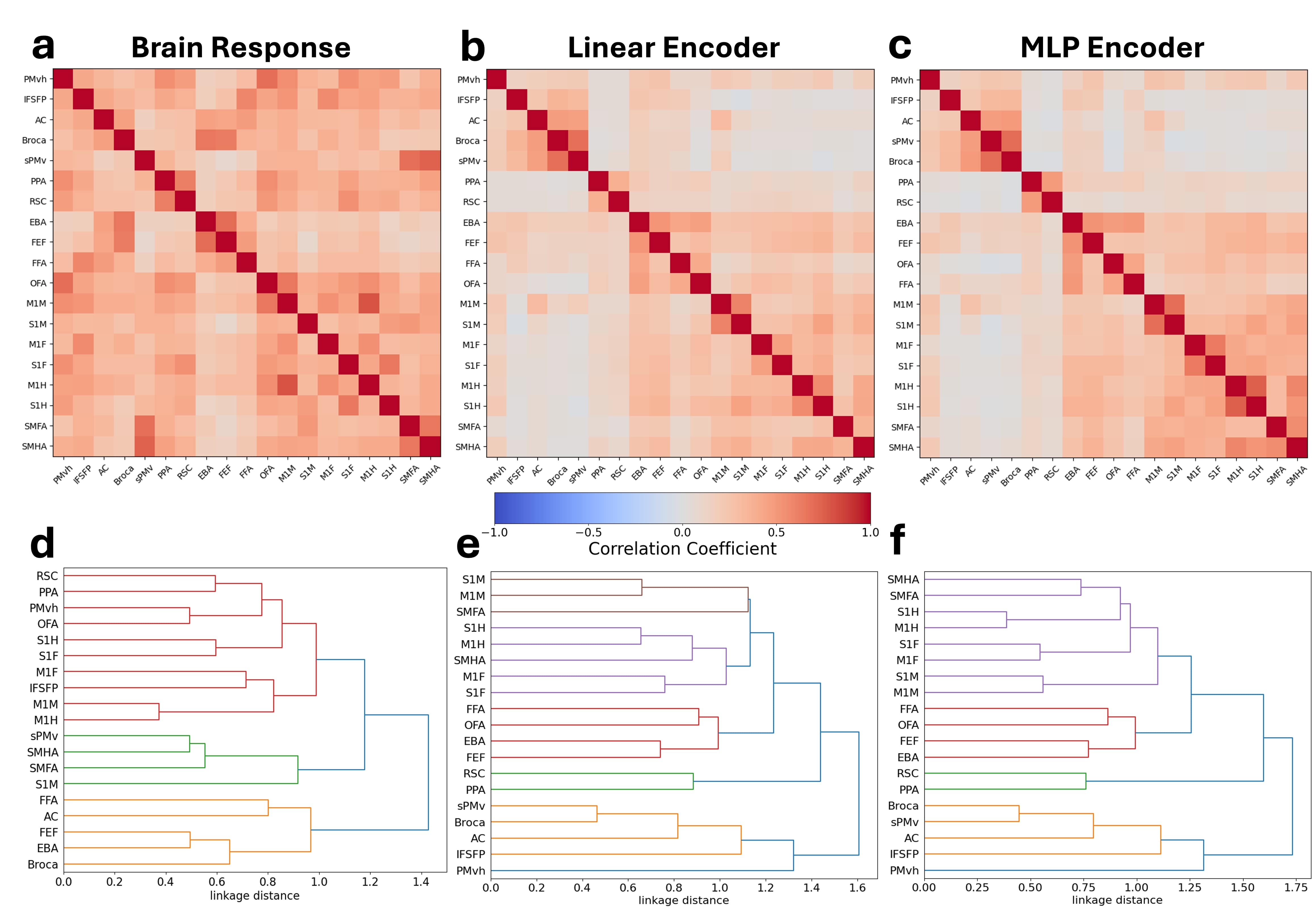}
\end{center}
\caption{Spatio-temporal clustering based on Relative Error Difference (RED) between semantic and audio encoding models. Panels (a-c) display correlation matrices representing the temporal relationships between regions of interest (ROIs). For consistency, all the ROIs in (a,b,c) are ordered according to the most optimal ordering for (c). Panel (a) shows the functional connectivity (FC) matrix, calculated from the average fMRI signals. Panel (b) presents the correlation matrix from Relative Error Difference between Whisper and LLAMA using linear encoders, while panel (c) uses nonlinear (MLP) encoders, showing better functional compartmentalization with stronger block-diagonal structures. Panels (d-f) depict hierarchical clustering dendrograms derived from the correlation matrices in panels (a-c). Panel (d), based on FC, shows no clear compartmentalization of ROIs. Panel (e), based on linear encoders, show almost perfect functional clustering, though with inaccuracies (e.g., SMFA clustered with S1M/M1M). Panel (f), based on nonlinear (MLP) encoders, achieves better functional clustering, correctly grouping motor-related regions. The modularity  Q  values confirm this improvement: FC (a) scored 0.068, linear encoders (b) scored 0.145, and nonlinear encoders (c) scored 0.155, highlighting the advantage of nonlinear encoders for functional organization.}
\label{Fig spatio temporal}
\end{figure*}

Where  $N$  is the number of voxels in the ROI. The correlation matrices were then computed over these ROI time series for both linear and nonlinear (MLP) encoders (Figure \ref{Fig spatio temporal} (b, c)). A high correlation between two ROIs indicates that their semantic/audio processing temporal dynamics are similar over time.

For comparison, functional connectivity (FC) was also computed using the average fMRI signal for each voxel (Figure \ref{Fig spatio temporal} a). Hierarchical clustering was then performed on the correlation matrices, producing the dendrograms in panels (d-f).

As shown in Figure \ref{Fig spatio temporal}, panel (d) does not exhibit meaningful compartmentalization, indicating that the ROIs are not functionally clustered based on FC. However, the correlation matrices derived from RED (panels b, c) demonstrate clear block-diagonal structures, suggesting better functional compartmentalization. The dendrograms in panels (e, f) show that the ROIs cluster according to their functional roles, where the somatosensory and motor areas, visual areas, and auditory areas are grouped (even lower levels are grouped well (M1H/S1H, M1M/S1M, M1F/S1F, SMHA/SMFA, Broca/sPMv are grouped)) with nonlinear (MLP) models (f) achieving more accurate clustering than linear models (e). Specifically, panel (e) incorrectly clusters SMFA with S1M and M1M, whereas panel (f) correctly clusters SMHA and SMFA together before clustering them with other sensory and motor-related regions.

This study presents a novel approach, as it is the first to use fMRI language encoding models to group ROIs based not only on spatial dynamics but also on their temporal processing dynamics. Traditionally, voxel-wise functional classification or grouping has been the norm in fMRI analysis, focusing solely on spatial relationships. However, here with the help of fMRI encoders, we incorporate both spatial and temporal information, allowing for a more comprehensive view of brain function, especially in the context of semantic and auditory encoding.

In summary, using nonlinear (MLP) models leads to better functional compartmentalization. In fact, modularity  Q  values further confirm this: FC (a) scored 0.068, linear encoders (b) scored 0.145, and nonlinear encoders (c) scored 0.155, highlighting the improved functional clustering achieved with better encoders.

\FloatBarrier

\section{Improvements from multimodality}\label{Appendix : Improvements from multimodality}
\subsection{Voxelwise improvements from multimodality ($r$ analysis)}\label{multimodal improvement-r}

\begin{figure}[ht]
\begin{center}
\includegraphics[width=\linewidth]{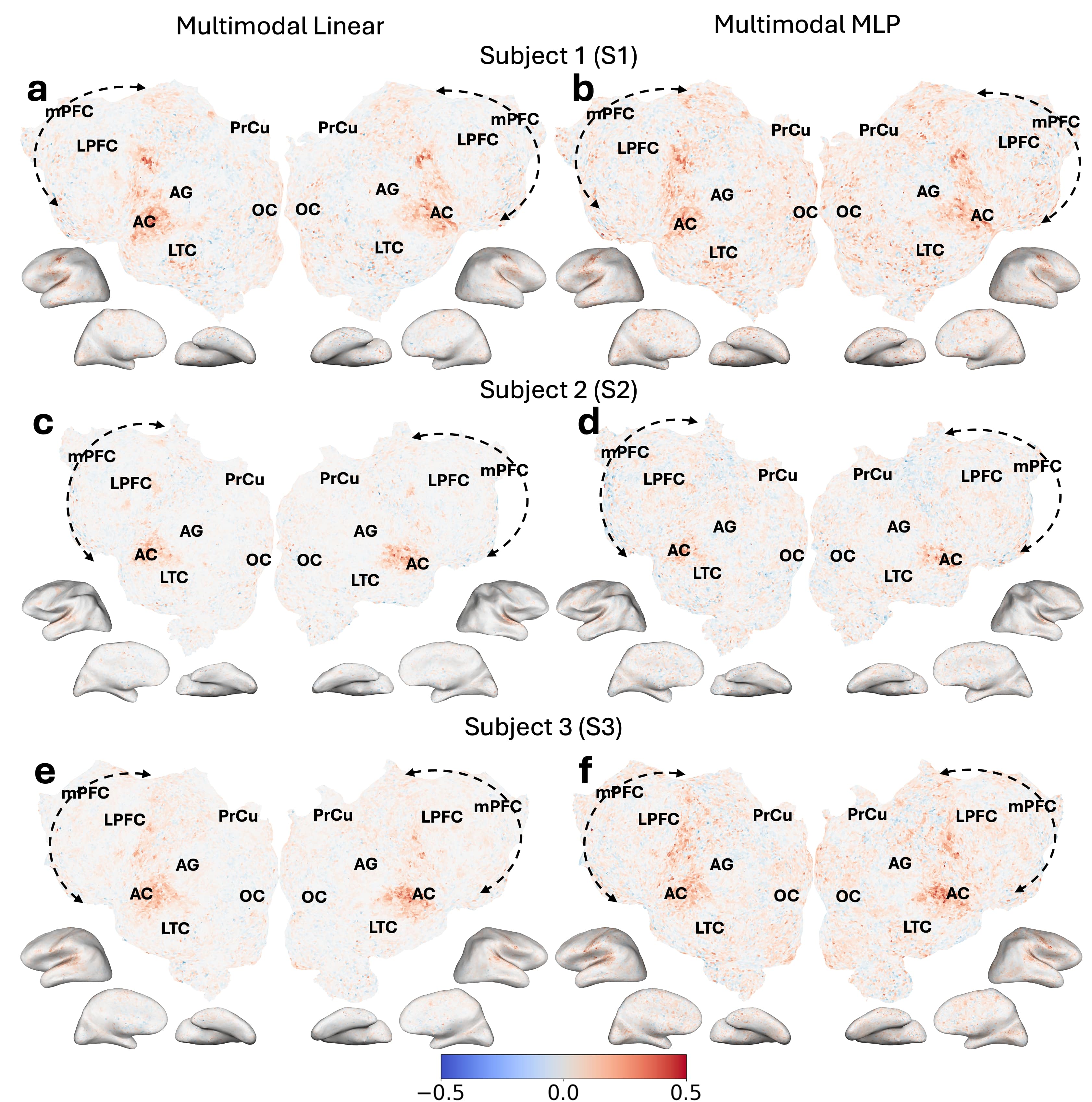}
\end{center}
\caption{Subject-wise voxelwise  $\Delta r$  plots of multimodal models compared to semantic models. Panels (a-f) display voxelwise  $\Delta r$  values comparing multimodal and unimodal models across three subjects. Panels a, c, e show the difference between multimodal linear and semantic linear models, while panels b, d, f compare multimodal MLP and semantic linear models. Each row represents a different subject: Subject 1 (S1) in panels a-b, Subject 2 (S2) in panels c-d, and Subject 3 (S3) in panels e-f. Warmer colors indicate regions where the multimodal models outperform the unimodal linear models in prediction accuracy. The spatial patterns highlight enhanced encoding performance in key areas associated with semantic and auditory processing, such as the medial prefrontal cortex (mPFC), angular gyrus (AG), precuneus (PrCu), and lateral temporal cortex (LTC), emphasizing the benefits of multimodal models in capturing complex brain activity.}
\label{Fig all subs multimodal voxelwise}
\end{figure}

\begin{figure}[h]
\begin{center}
\includegraphics[width=\linewidth]{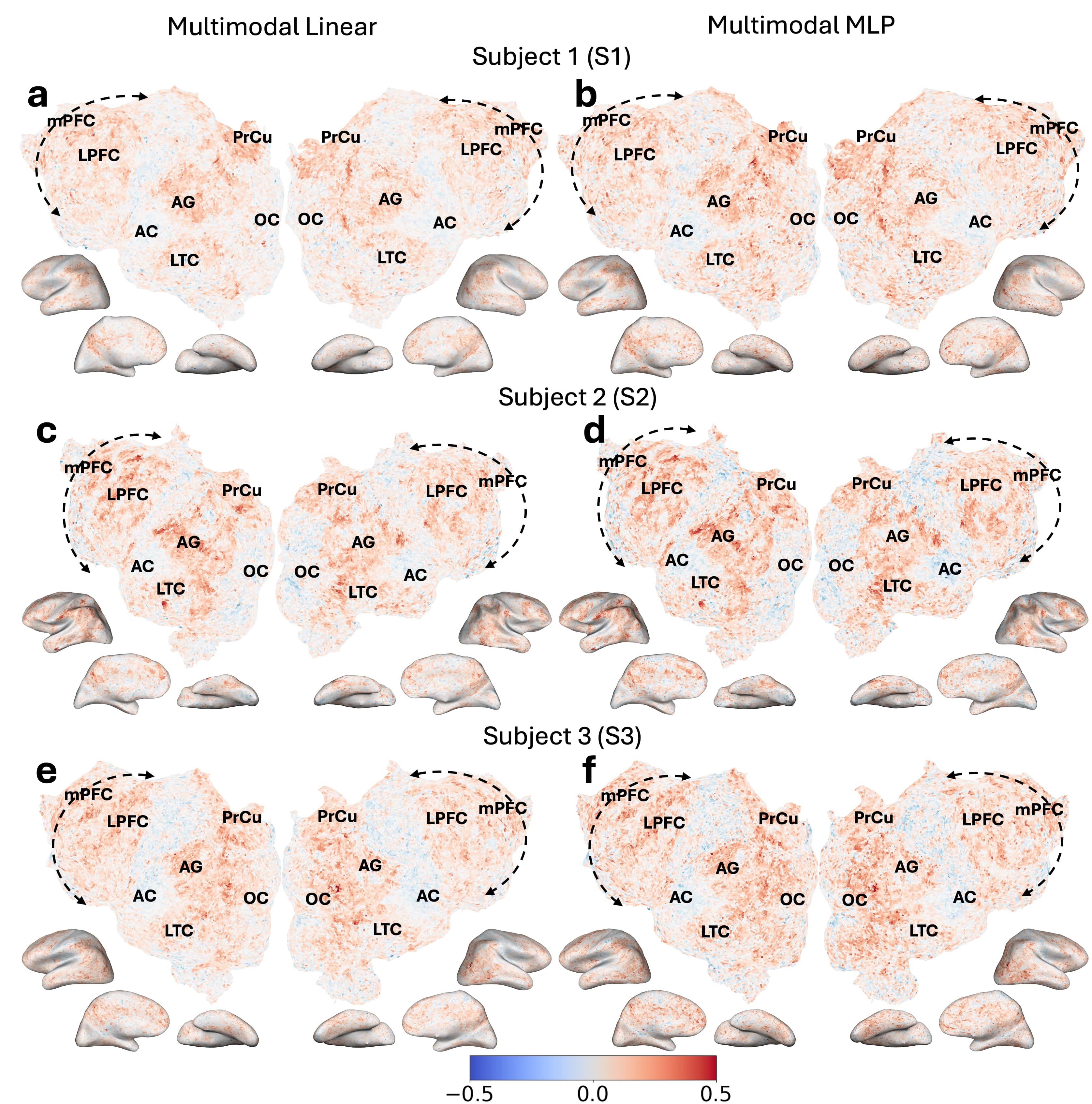}
\end{center}
\caption{Subject-wise voxelwise  $\Delta r$  plots of multimodal models compared to audio models. Panels (a-f) display voxelwise  $\Delta r$  values comparing multimodal and unimodal models across three subjects. Panels a, c, e show the difference between multimodal linear and audio linear models, while panels b, d, f compare multimodal MLP and audio linear models.  Each row represents a different subject: Subject 1 (S1) in panels a-b, Subject 2 (S2) in panels c-d, and Subject 3 (S3) in panels e-f. Warmer colors indicate regions where the multimodal models outperform the unimodal linear models in prediction accuracy.}
\label{Fig all subs multimodal voxelwise_2}
\end{figure}

This section shows the subject-wise plots of voxelwise $\Delta r$ between multimodal linear/MLP and semantic/audio linear models (Figure \ref{Fig all subs multimodal voxelwise}, Figure \ref{Fig all subs multimodal voxelwise_2}). We observe consistent patterns of improvement when using multimodal models.

\FloatBarrier

\subsection{Voxelwise improvements from multimodality ($CC_{norm}$ analysis)}\label{multimodal improvement-CCnorm}
This section shows the subject-wise plots of voxelwise $\Delta CC_{norm}$ between multimodal linear/MLP and semantic/audio linear models (Figure \ref{Fig all subs multimodal voxelwise_2 ccnorm}, Figure \ref{Fig all subs multimodal voxelwise_2 ccnorm}). We observe consistent patterns of improvement when using multimodal models. The improvements are more noticable with $CC_{norm}$ compared to $r$ as noise is taken into account. 

\begin{figure}[h]
\begin{center}
\includegraphics[width=\linewidth]{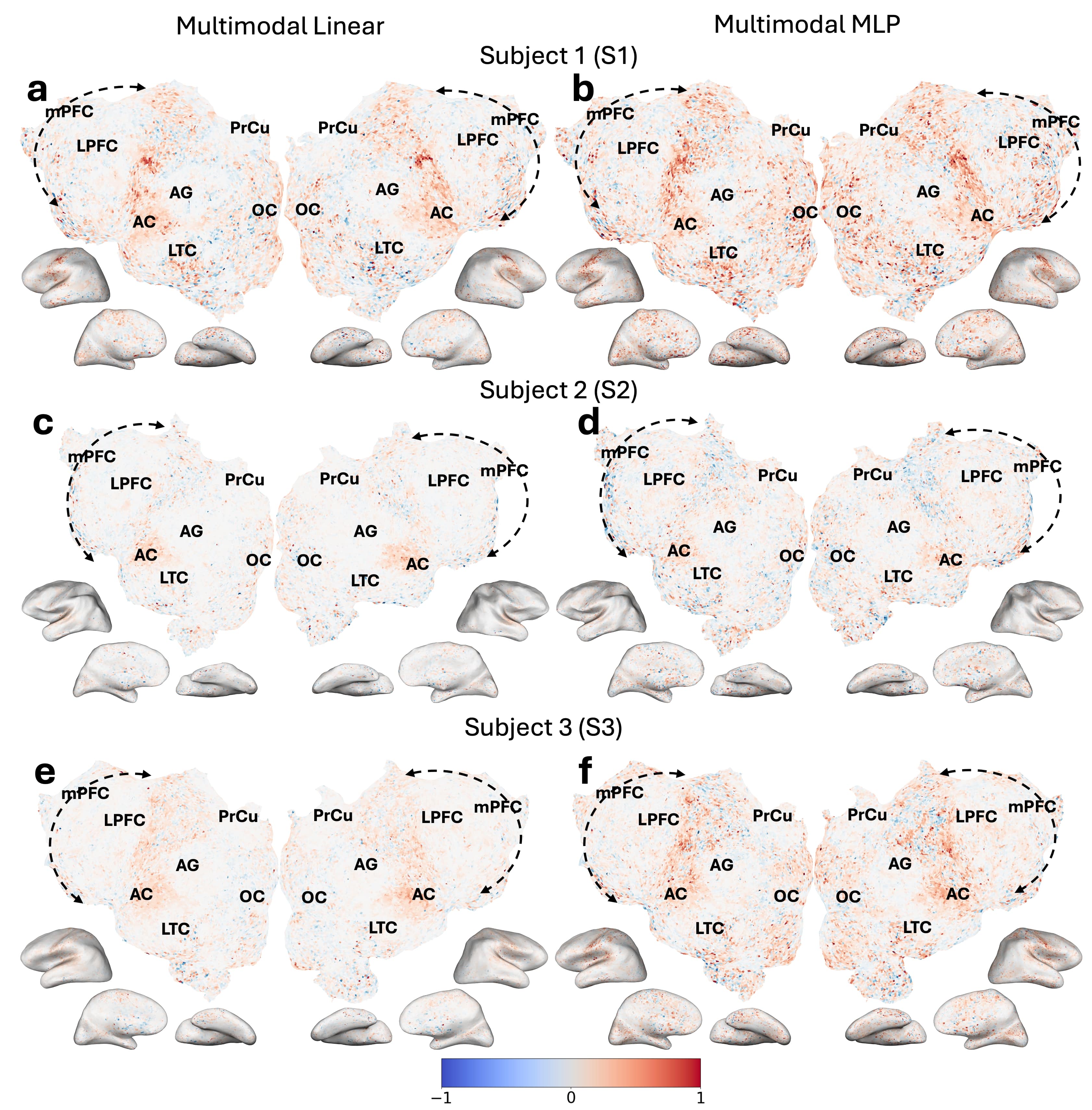}
\end{center}
\caption{Subject-wise voxelwise  $\Delta CC_{norm}$  plots of multimodal models compared to semantic models. Panels (a-f) display voxelwise  $\Delta CC_{norm}$  values comparing multimodal and unimodal models across three subjects. Panels a, c, e show the difference between multimodal linear and semantic linear models, while panels b, d, f compare multimodal MLP and semantic linear models.  Each row represents a different subject: Subject 1 (S1) in panels a-b, Subject 2 (S2) in panels c-d, and Subject 3 (S3) in panels e-f. Warmer colors indicate regions where the multimodal models outperform the unimodal linear models in prediction accuracy.}
\label{Fig all subs multimodal voxelwise_2 ccnorm}
\end{figure}

\begin{figure}[h]
\begin{center}
\includegraphics[width=\linewidth]{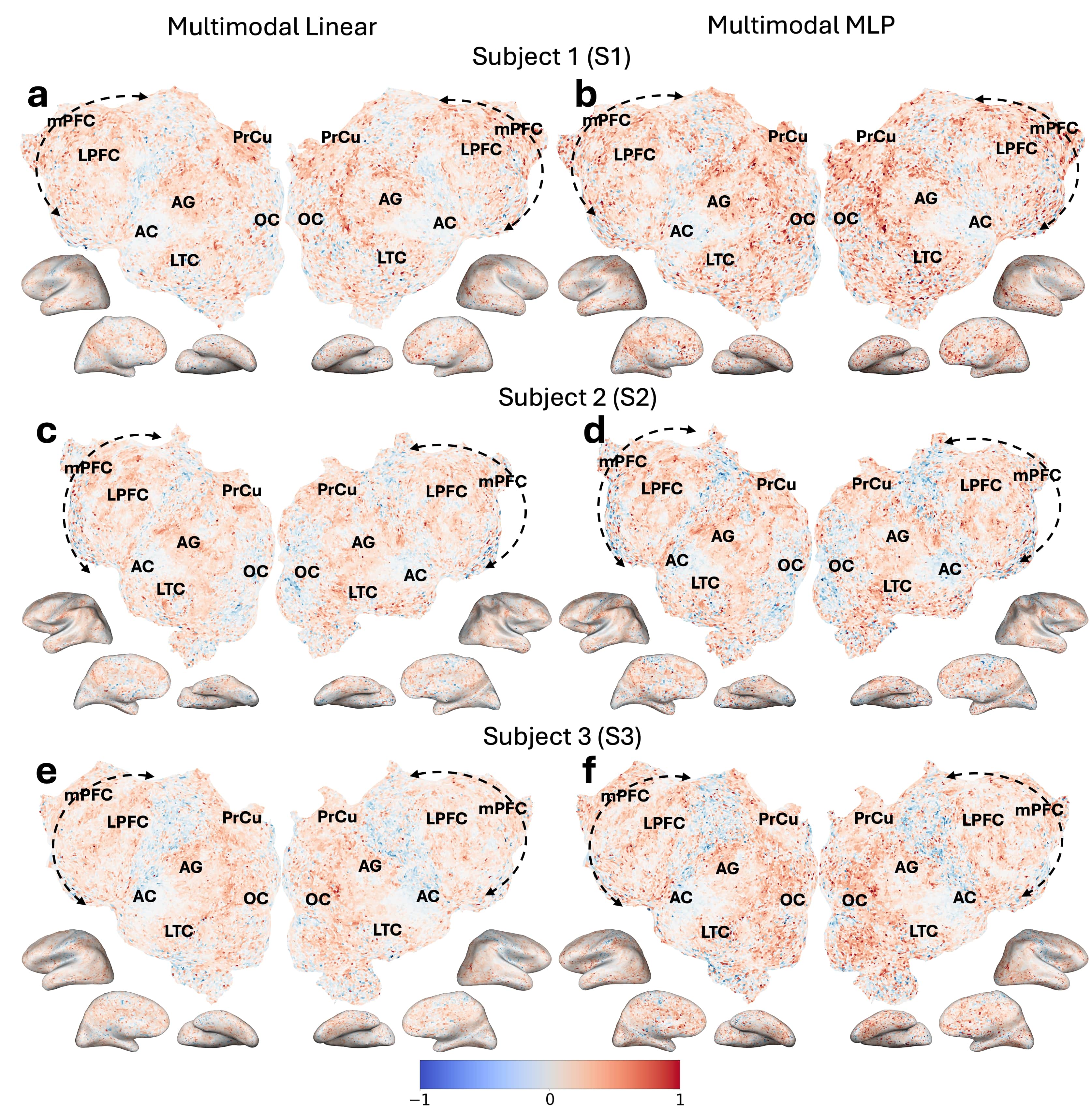}
\end{center}
\caption{Subject-wise voxelwise  $\Delta CC_{norm}$  plots of multimodal models compared to audio models. Panels (a-f) display voxelwise  $\Delta CC_{norm}$  values comparing multimodal and unimodal models across three subjects. Panels a, c, e show the difference between multimodal linear and audio linear models, while panels b, d, f compare multimodal MLP and audio linear models.  Each row represents a different subject: Subject 1 (S1) in panels a-b, Subject 2 (S2) in panels c-d, and Subject 3 (S3) in panels e-f. Warmer colors indicate regions where the multimodal models outperform the unimodal linear models in prediction accuracy.}
\label{Fig all subs multimodal voxelwise_2 ccnorm}
\end{figure}

\FloatBarrier

\subsection{ROI predictions improvements from multimodality}
\begin{figure}[h]
\begin{center}
\includegraphics[width=\linewidth]{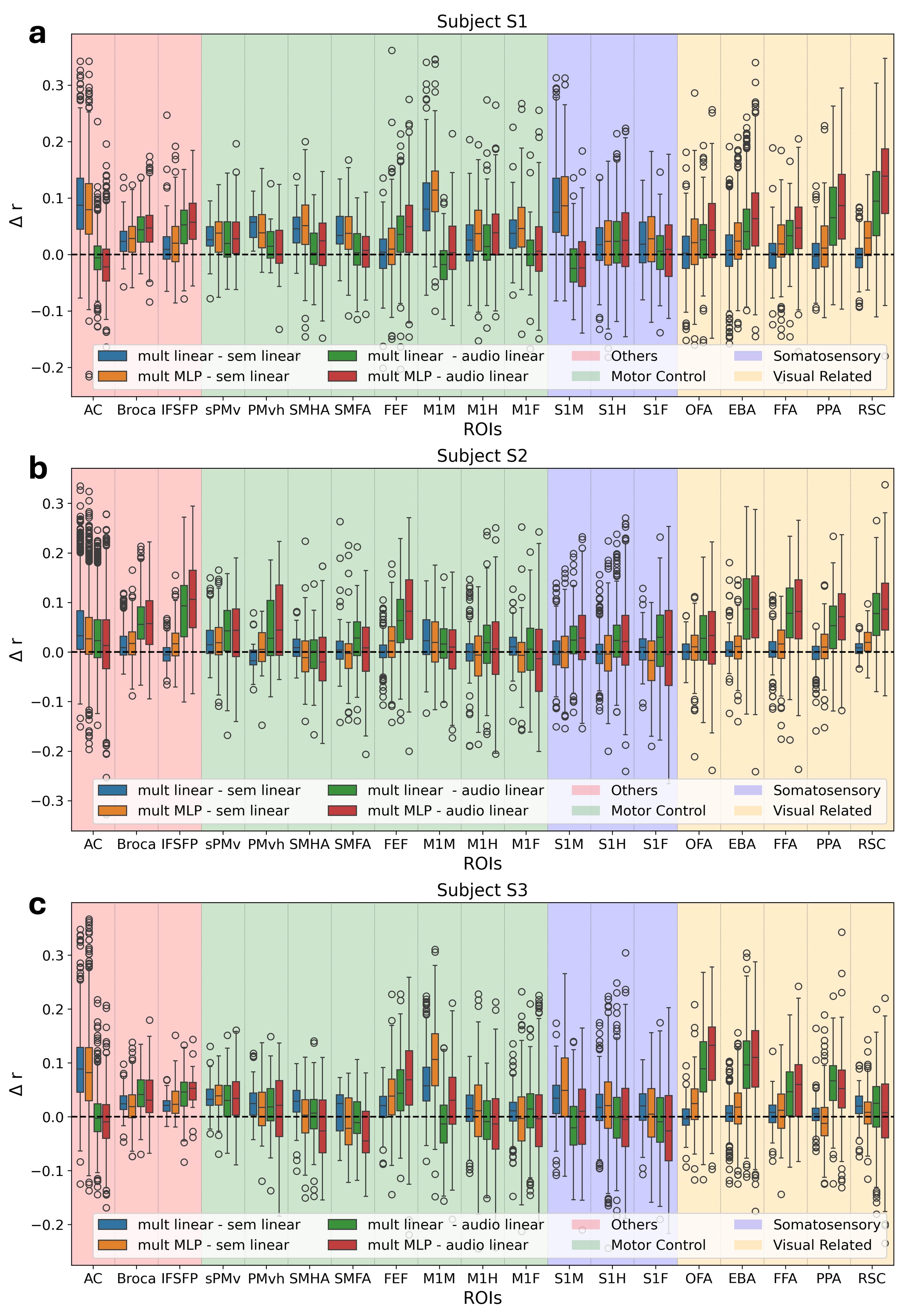}
\end{center}
\caption{Subject-wise boxplots of performance differences (\( \Delta r \)) across different ROIs. The comparisons are made between different stimuli and encoding models: multimodal linear and multimodal MLP (mult MLP) models are compared against semantic (sem) and audio linear models. The ROIs are grouped into functional categories.}
\label{Fig2 ROI subwise}
\end{figure}

This section shows the ROI-wise improvements from using multimodal models (Figure \ref{Fig2 ROI subwise})

\FloatBarrier

\section{Improvements from nonlinearity and multimodality}\label{DIMLP subwise appendix}
\subsection{Voxelwise improvements from DIMLP, and additional improvements from MLP ($r$ analysis)}

\begin{figure}[h]
\begin{center}
\includegraphics[width=\linewidth]{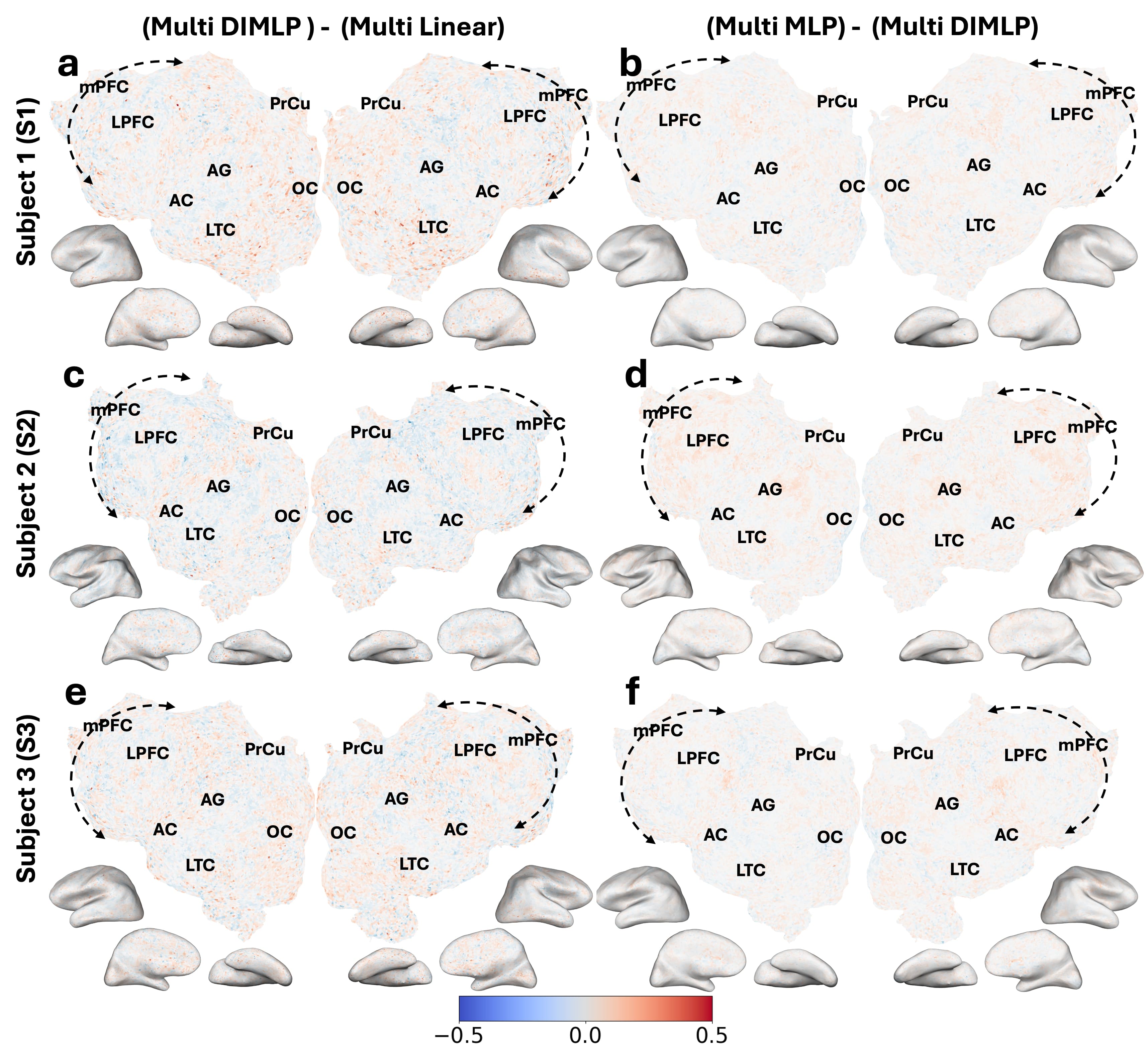}
\end{center}
\caption{Nonlinearity Enhances Multimodal fMRI Predictions. Panels (a, c, e) show the voxelwise  $\Delta r$  values (DIMLP minus linear model), illustrating the improvements achieved through nonlinear processing within each modality, while largely limiting cross-modal interactions. Panels (b, d, f) display voxelwise  $\Delta r$  values (Multi MLP minus Multi DIMLP), highlighting the additional benefits of allowing nonlinear interactions between modalities (``Multi" denotes Multimodal). Each row represents the same subject: Subject 1 (S1) in panels a-b, Subject 2 (S2) in panels c-d, and Subject 3 (S3) in panels e-f. Warmer colors indicate regions where the nonlinear models outperform linear models.}
\label{Fig all subs nonlinear and multimodal voxelwise}
\end{figure}

\FloatBarrier

\subsection{Voxelwise improvements from DIMLP, and additional improvements from MLP ($CC_{norm}$ analysis)}

\begin{figure}[h]
\begin{center}
\includegraphics[width=\linewidth]{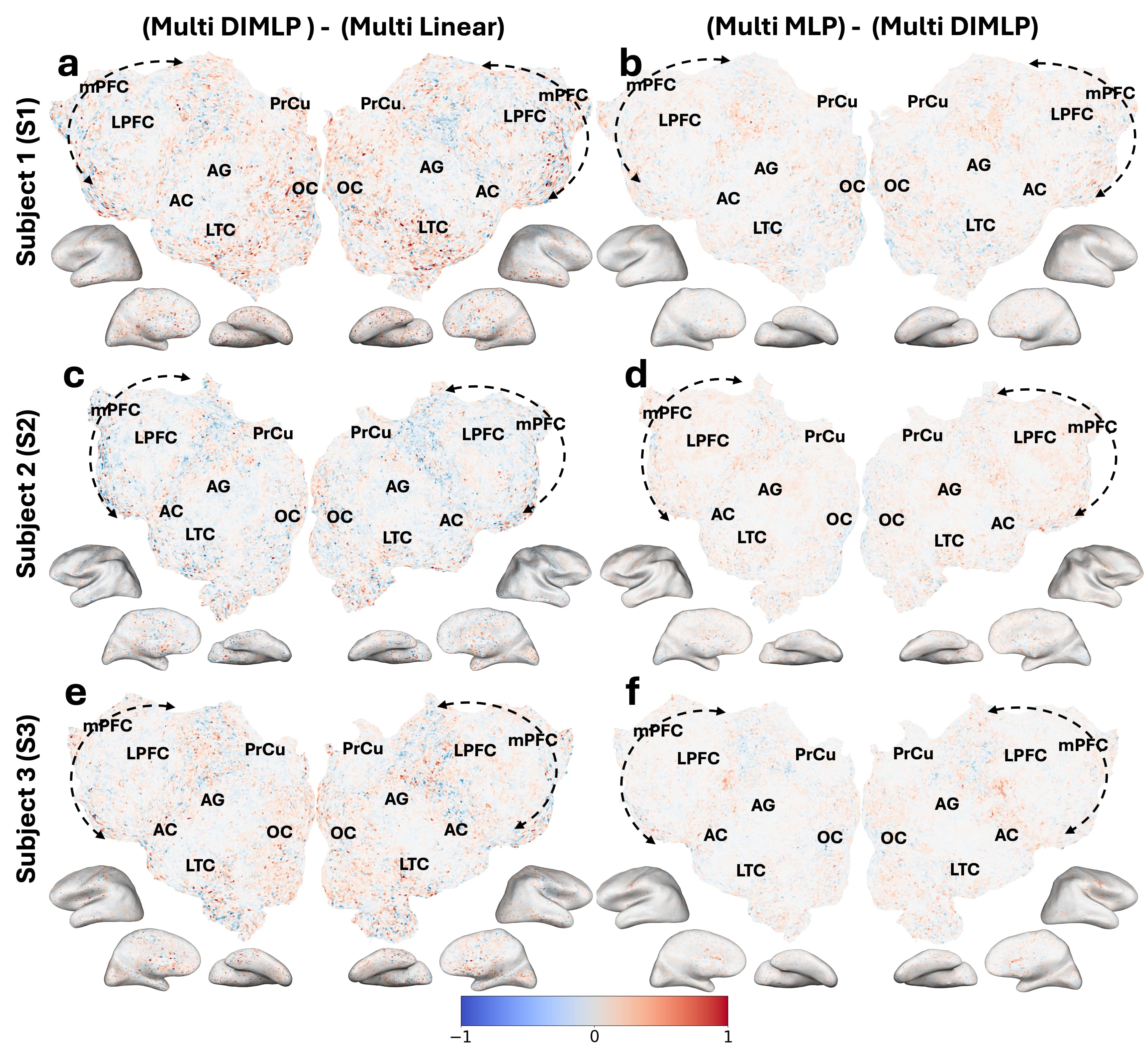}
\end{center}
\caption{Nonlinearity Enhances Multimodal fMRI Predictions. Panels (a, c, e) show the voxelwise  $\Delta CC_{norm}$  values (DIMLP minus linear model), illustrating the improvements achieved through nonlinear processing within each modality, while largely limiting cross-modal interactions. Panels (b, d, f) display voxelwise  $\Delta CC_{norm}$  values (Multi MLP minus Multi DIMLP), highlighting the additional benefits of allowing nonlinear interactions between modalities (``Multi" denotes Multimodal). Each row represents the same subject: Subject 1 (S1) in panels a-b, Subject 2 (S2) in panels c-d, and Subject 3 (S3) in panels e-f. Warmer colors indicate regions where the nonlinear models outperform linear models.}
\label{Fig all subs nonlinear and multimodal voxelwise ccnorm}
\end{figure}

Figure \ref{Fig all subs nonlinear and multimodal voxelwise ccnorm} shows the voxel-wise performance improvements in voxelwise $CC_{norm}$ values when incorporating nonlinear interactions. The improvements are more pronouned with $CC_{norm}$ compared to $r$ as noise is taken into account. 

\FloatBarrier

\subsection{ROI-wise improvements of multimodal DIMLP and MLP from multimodal linear model}

\begin{figure}[h]
\begin{center}
\includegraphics[width=\linewidth]{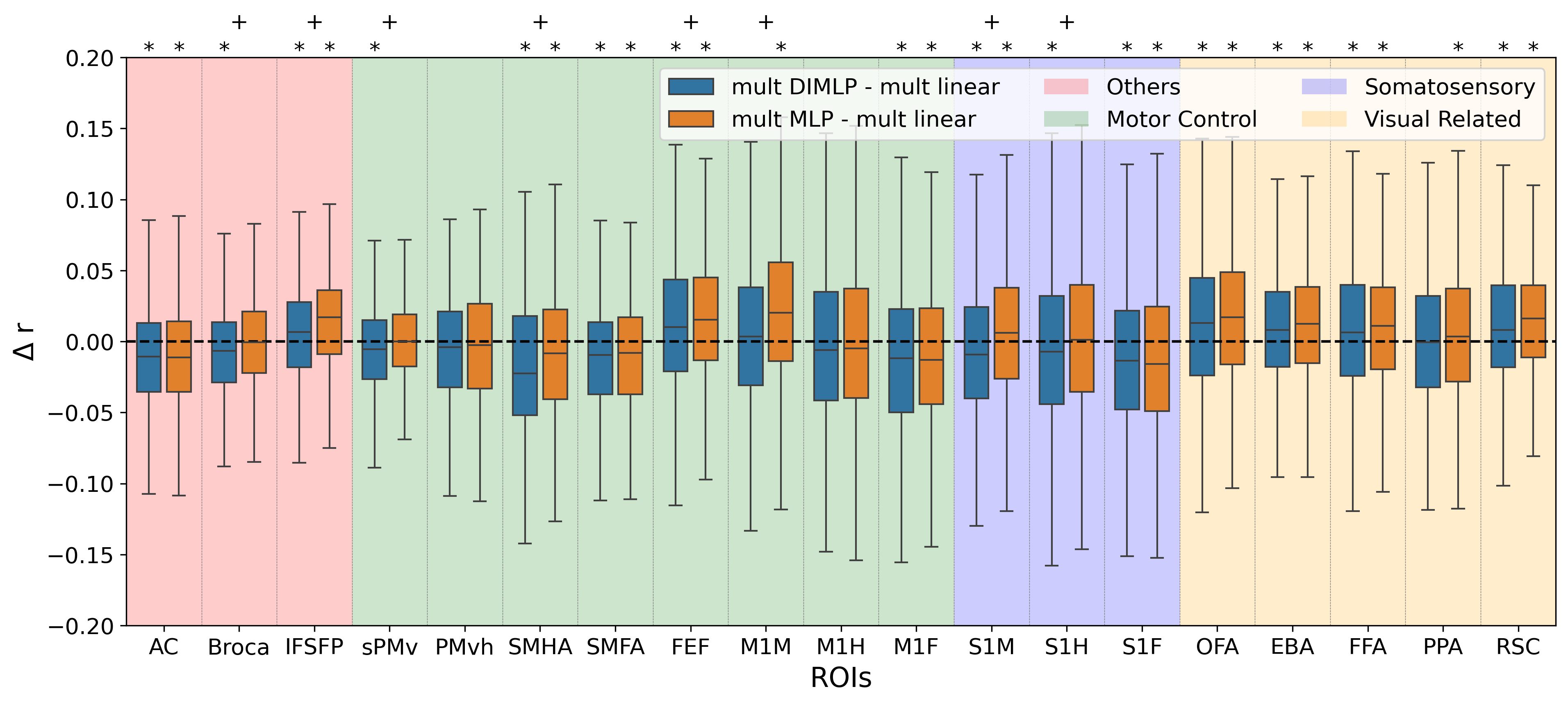}
\end{center}
\caption{Box plot showing $\Delta r$ across ROIs, where the $\Delta r$ values are aggregated over all subjects. \textit{multi} refers to multimodal, and \textit{sem} refers to semantic encoders, and \textit{DIMLP} refers to Delayed Interaction MLP, where only a \textit{linear} interaction between modalities is allowed. The ROIs are color-coded by function. Regions where $\Delta r > 0$ with a p-value less than 0.05 are indicated by * symbols. Additionally, + symbols denote ROIs where there is a statistically significant difference (p-value $<$ 0.05) between the two models based on a pairwise t-test. Voxelwise and ROI-wise plots for each subjects can be found in Figure \ref{Fig all subs nonlinear and multimodal voxelwise} (Appendix), and Figure \ref{Fig3 ROI subwise} (Appendix), respectively.}
\label{Fig multimodal nonlinear}
\end{figure}

\begin{figure}[h]
\begin{center}
\includegraphics[width=\linewidth]{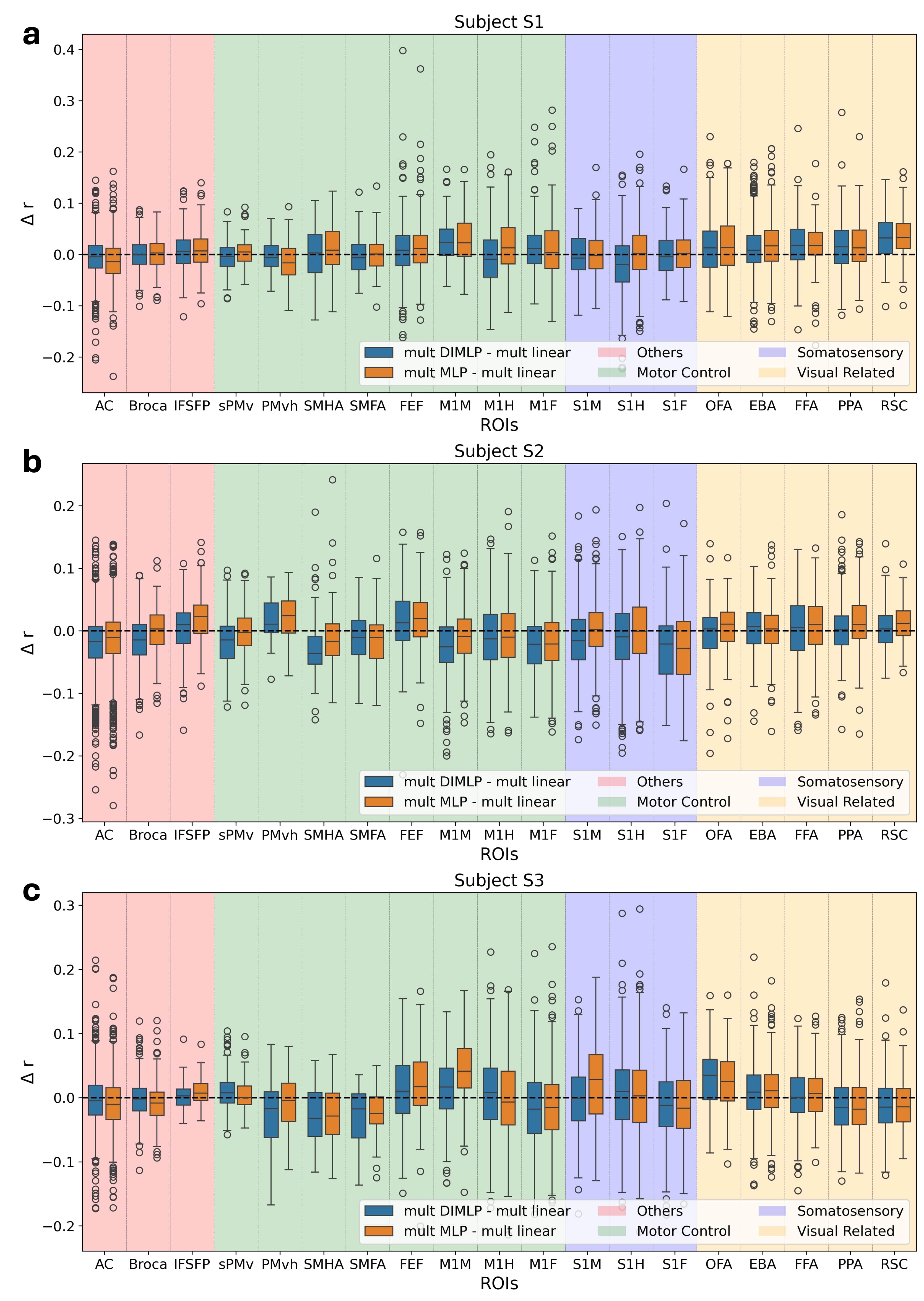}
\end{center}
\caption{Subject-wise boxplots of voxel-wise differences (\( \Delta r \)) across different ROIs. The comparisons are made between different encoding models: multimodal MLP and multimodal DIMLP models are compared against multimodal linear models. The ROIs are grouped into functional categories.}
\label{Fig3 ROI subwise}
\end{figure}

\FloatBarrier

\section{Variance partitioning analysis} \label{Appendix, ALL VARIANCE PARTITINOING}
To quantify the unique contributions of different feature spaces in our nonlinear multimodal encoding models, we employed a variance partitioning analysis similar to \cite{encoding_model_3_hierarchical}. This approach allowed us to determine how much variance could be uniquely explained by each feature versus that explained by a multiple features. We estimated both the fraction of variance explained by each feature space individually and the fraction that might be equally well explained by combinations of feature spaces.

We show our variance partitioning analysis results in three complementary ways: 1) voxel-wise variance partition results (Appendix \ref{VP various models}), 2) voxel-wise plots showing the largest variance partition for each voxel (Appendix \ref{App : Largest variance partitioning for each voxel}), and 3) ROI-wise Venn diagrams illustrating the distribution of variance explained across different brain regions (Appendix \ref{App Variance partitioning Venn diagram}).

For this analysis, we fit models with all possible combinations of feature spaces: two single-feature models (audio and semantic), one model combining both features (semantic-audio), and examined the distribution of variance explained within brain regions. This allowed us to decompose the total explained variance into three components: variance uniquely explained by audio features, variance uniquely explained by semantic features, and variance jointly explained by both feature spaces.

\subsection{Summary of variance partitioning results}

Looking at the results of Appendix \ref{VP various models}, we observe that joint variance dominates across most cortical regions, contrasting with \cite{encoding_model_3_hierarchical} where semantic only features showed greater dominance. This difference likely stems from our feature choices - whereas \cite{encoding_model_3_hierarchical} used spectral and articulatory features that primarily contained information relevant mostly only to auditory cortex, our use of Whisper features provides richer auditory representations that enable better predictions beyond traditional auditory regions. This finding aligns with our earlier argument (Section \ref{Subsection:Multimodal motor and sensory area predictions}) that multiple modalities jointly contribute to neural computations across the cortex rather than having one modality dominate.

The dominance pattern of joint variance is consistent both within and near AC, with a notable exception in early auditory regions where audio features show unique contributions. This hierarchical organization suggests that while early AC predominantly processes pure acoustic information, later AC regions integrate both semantic and auditory features for higher-level speech processing. The unique contribution of audio features in early AC is noteworthy as it suggests preservation of modality-specific processing at early sensory stages despite using rich Whisper features.

Also, Appendix \ref{App : Largest variance partitioning for each voxel} reveals distinct spatial patterns in feature representation across cortical regions. The prefrontal cortex exhibits mixed dominance patterns, showing both joint semantic-audio representation and semantic-only areas. While early auditory cortex shows expected unique audio contributions, we also observe audio-specific representation in motor-sensory mouth areas (M1M, S1M), though this pattern varies across subjects. 

The ROI-wise analysis in Appendix \ref{App Variance partitioning Venn diagram} reveals that joint semantic-audio features dominate cortical representation, accounting for approximately 65\% of significantly predicted voxels across the entire cortex. Core language-processing regions (AC, Broca's area, sPMv) show particularly strong joint representation (around 80 to 90\%), supporting our hypothesis that speech comprehension relies on integrated multimodal processing. This integration is consistently observed across subjects, though some ROIs (e.g., PMvh in Subject S2 with only 14 voxels) have insufficient data for reliable interpretation. The transition from linear to MLP encoders increases the total number of significantly predicted voxels while maintaining similar representation patterns, indicating that nonlinear encoding primarily enhances prediction accuracy rather than fundamentally altering feature representation structure.

\subsection{Variance partitioning of various models} \label{VP various models}

\begin{figure}[h]
\begin{center}
\includegraphics[width=\linewidth]{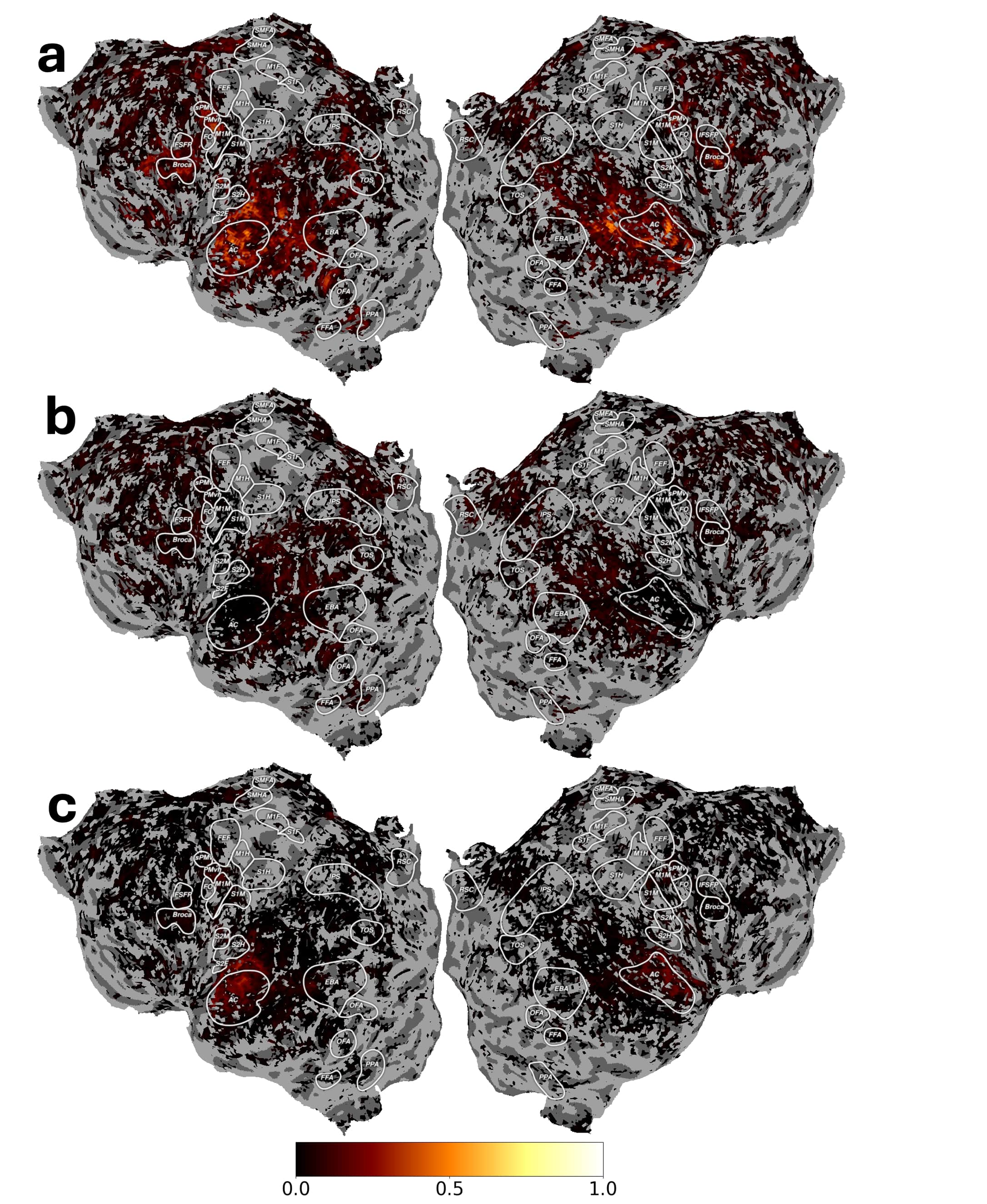}
\end{center}
\caption{Voxelwise variance partitioning analysis showing the contributions of different feature types to prediction accuracy for a subject S1 using linear models. The flatmaps display (a) variance jointly explained by audio and semantic features, (b) variance uniquely explained by semantic features, and (c) variance uniquely explained by audio features. Values shown are normalized correlations ($CC_{norm}$) for voxels where the joint model achieved significant prediction ($q(\text{FDR})<0.01$).} 
\label{Fig S1 VP Linear}
\end{figure}

Due to file size constraints, we only show the voxel-wise variance partitioning result of subject S1 using linear encoders, Figure \ref{Fig S1 VP Linear}. The rest have been moved to the supplementary material.

\begin{figure}[h]
\begin{center}
\includegraphics[width=\linewidth]{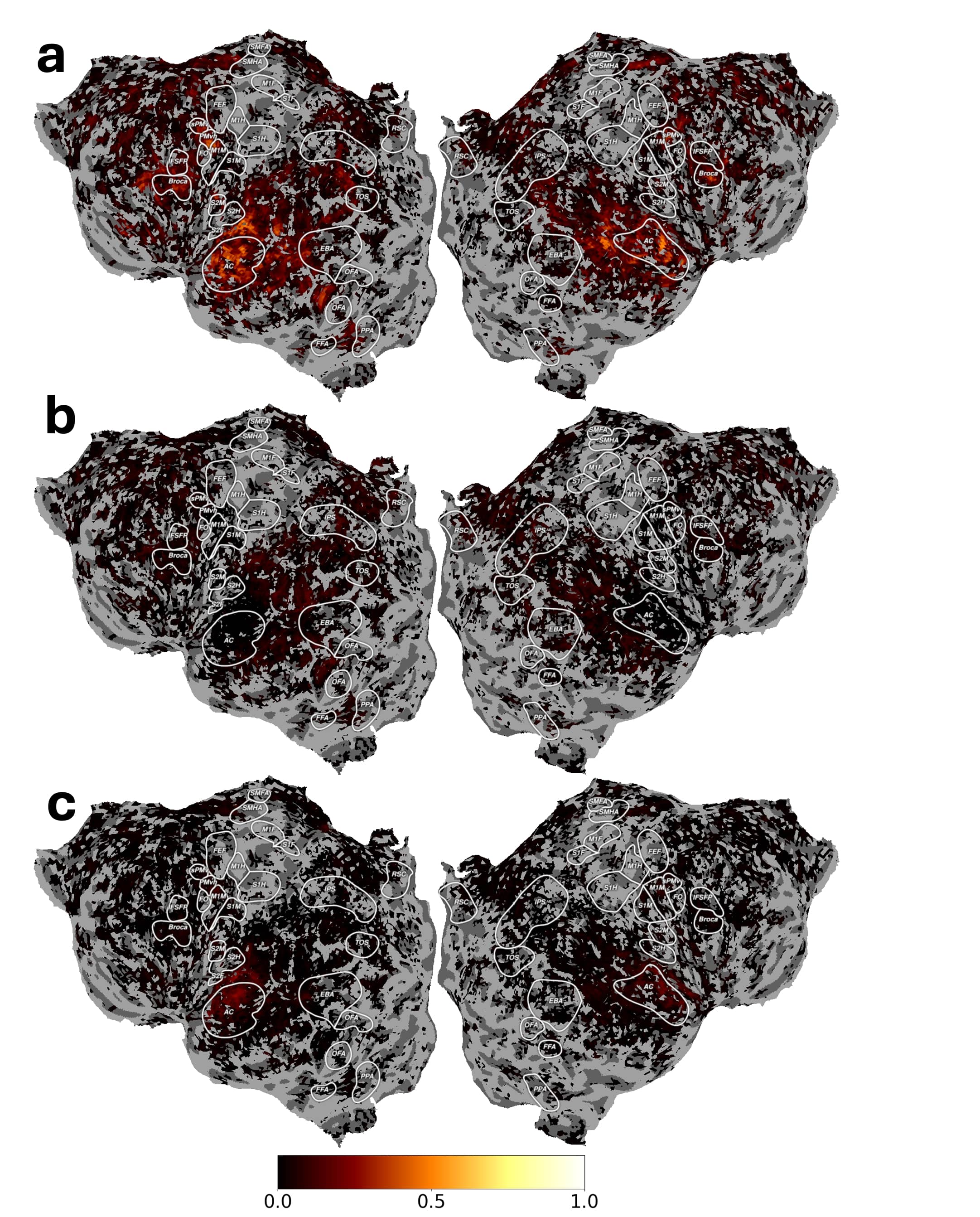}
\end{center}
\caption{Voxelwise variance partitioning analysis showing the contributions of different feature types to prediction accuracy for a subject S1 using MLP models. The flatmaps display (a) variance jointly explained by audio and semantic features, (b) variance uniquely explained by semantic features, and (c) variance uniquely explained by audio features. Values shown are normalized correlations ($CC_{norm}$) for voxels where the joint model achieved significant prediction ($q(\text{FDR})<0.01$).} 
\label{Fig S1 VP MLP}
\end{figure}

\begin{figure}[h]
\begin{center}
\includegraphics[width=\linewidth]{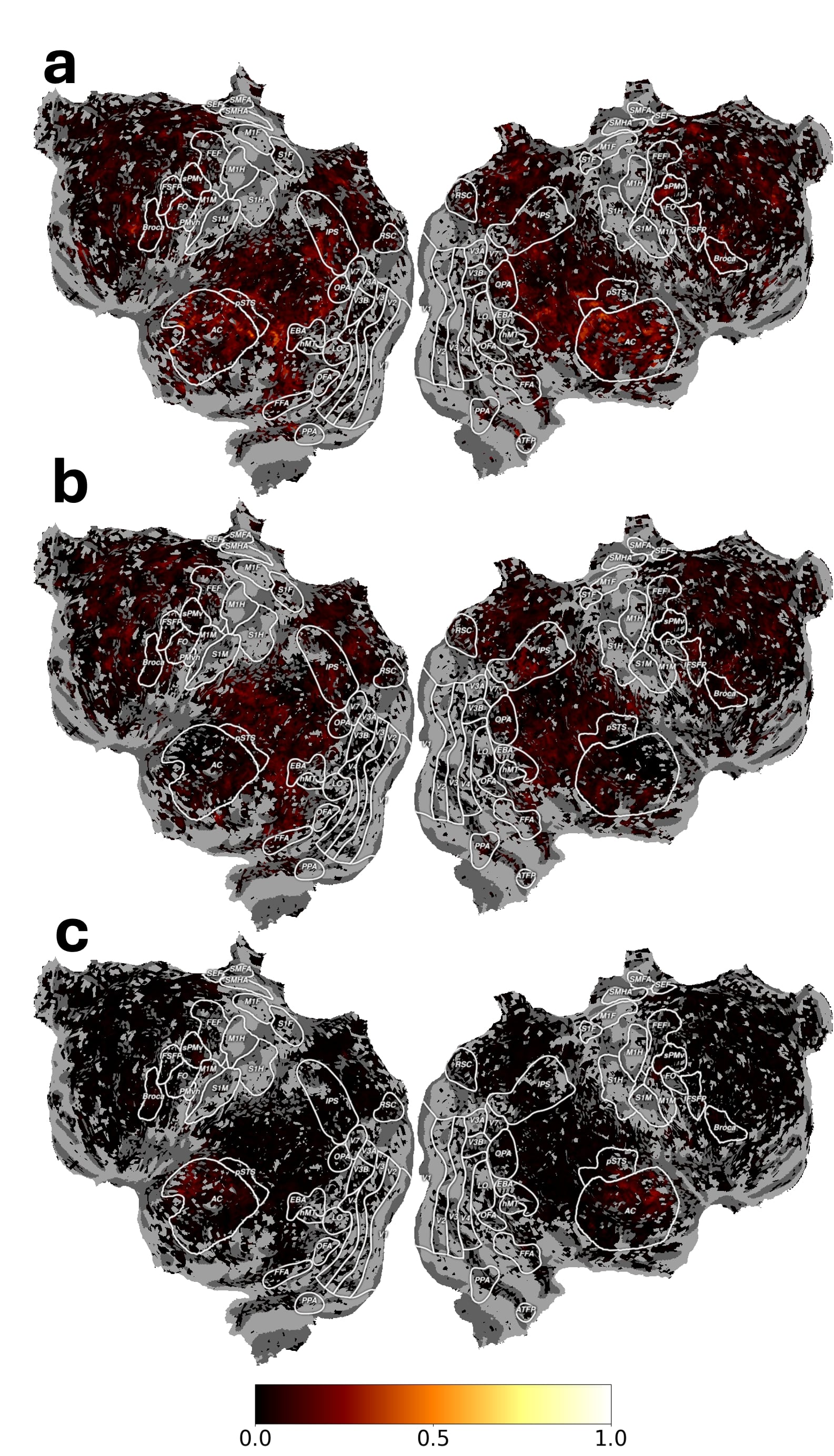}
\end{center}
\caption{Voxelwise variance partitioning analysis showing the contributions of different feature types to prediction accuracy for a subject S2 using linear models. The flatmaps display (a) variance jointly explained by audio and semantic features, (b) variance uniquely explained by semantic features, and (c) variance uniquely explained by audio features. Values shown are normalized correlations ($CC_{norm}$) for voxels where the joint model achieved significant prediction ($q(\text{FDR})<0.01$).} 
\label{Fig S2 VP Linear}
\end{figure}

\begin{figure}[h]
\begin{center}
\includegraphics[width=\linewidth]{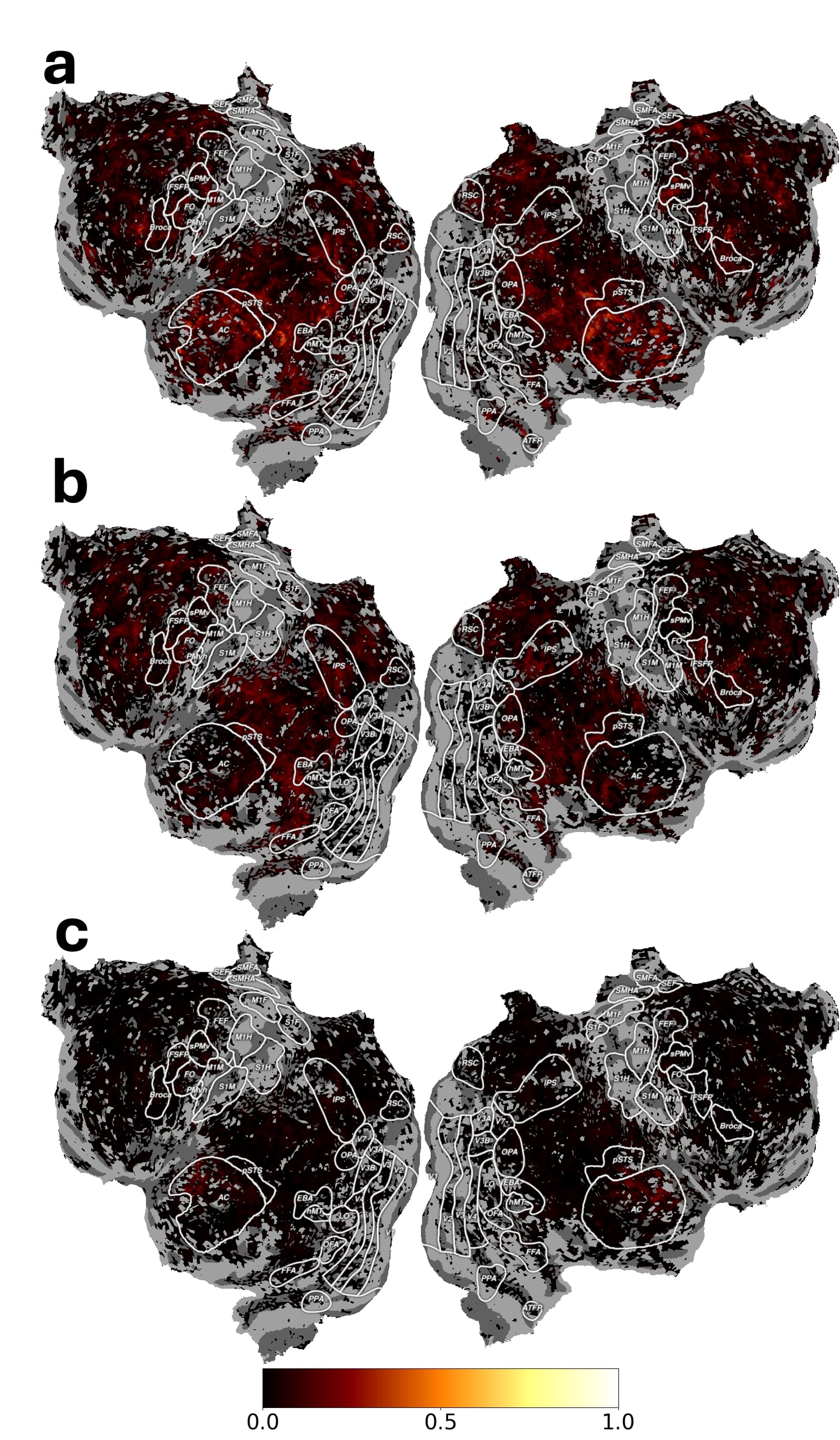}
\end{center}
\caption{Voxelwise variance partitioning analysis showing the contributions of different feature types to prediction accuracy for a subject S2 using MLP models. The flatmaps display (a) variance jointly explained by audio and semantic features, (b) variance uniquely explained by semantic features, and (c) variance uniquely explained by audio features. Values shown are normalized correlations ($CC_{norm}$) for voxels where the joint model achieved significant prediction ($q(\text{FDR})<0.01$).} 
\label{Fig S2 VP MLP}
\end{figure}

\begin{figure}[h]
\begin{center}
\includegraphics[width=\linewidth]{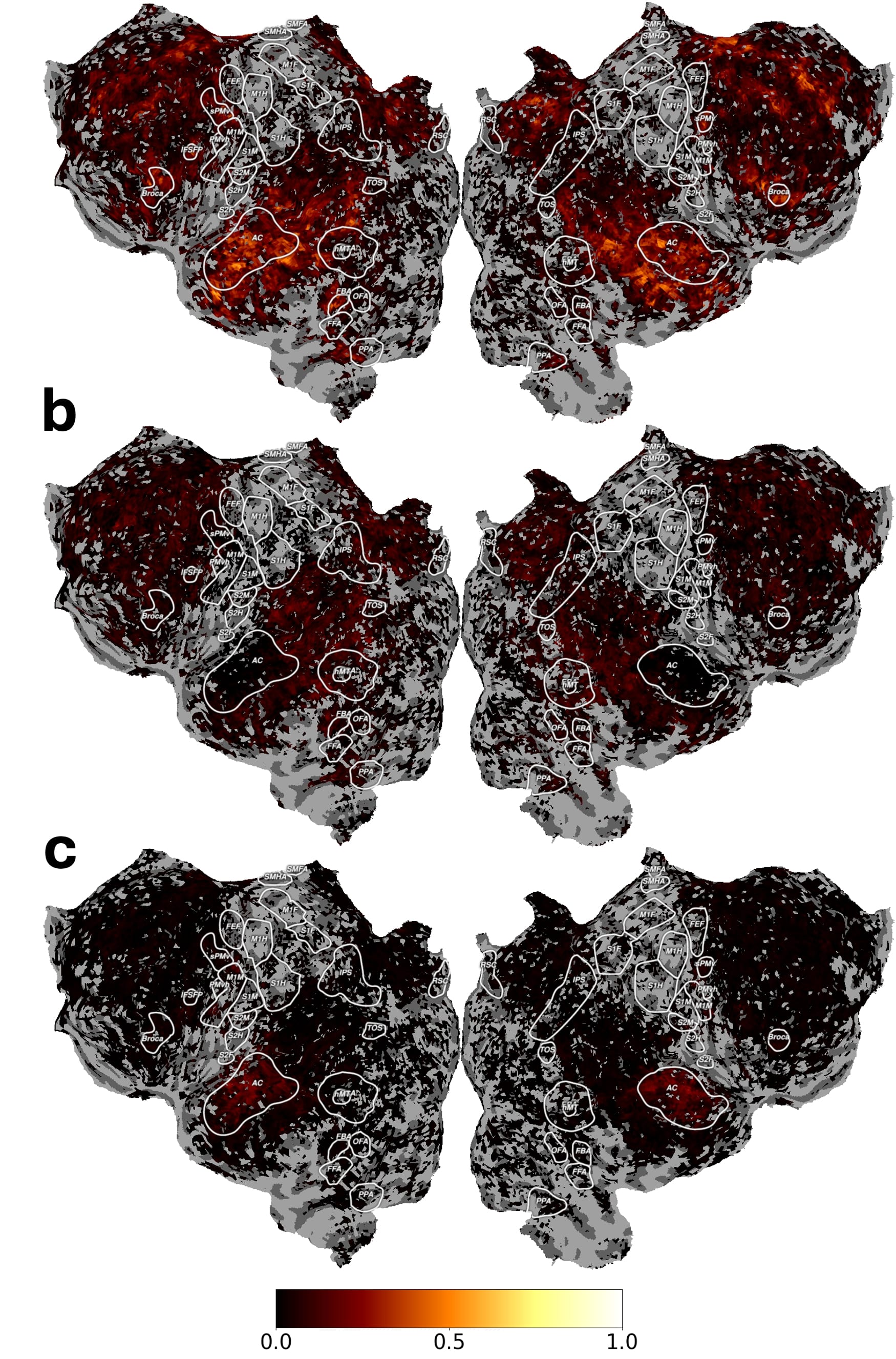}
\end{center}
\caption{Voxelwise variance partitioning analysis showing the contributions of different feature types to prediction accuracy for a subject S3 using linear models. The flatmaps display (a) variance jointly explained by audio and semantic features, (b) variance uniquely explained by semantic features, and (c) variance uniquely explained by audio features. Values shown are normalized correlations ($CC_{norm}$) for voxels where the joint model achieved significant prediction ($q(\text{FDR})<0.01$).} 
\label{Fig S3 VP Linear}
\end{figure}

\begin{figure}[h]
\begin{center}
\includegraphics[width=\linewidth]{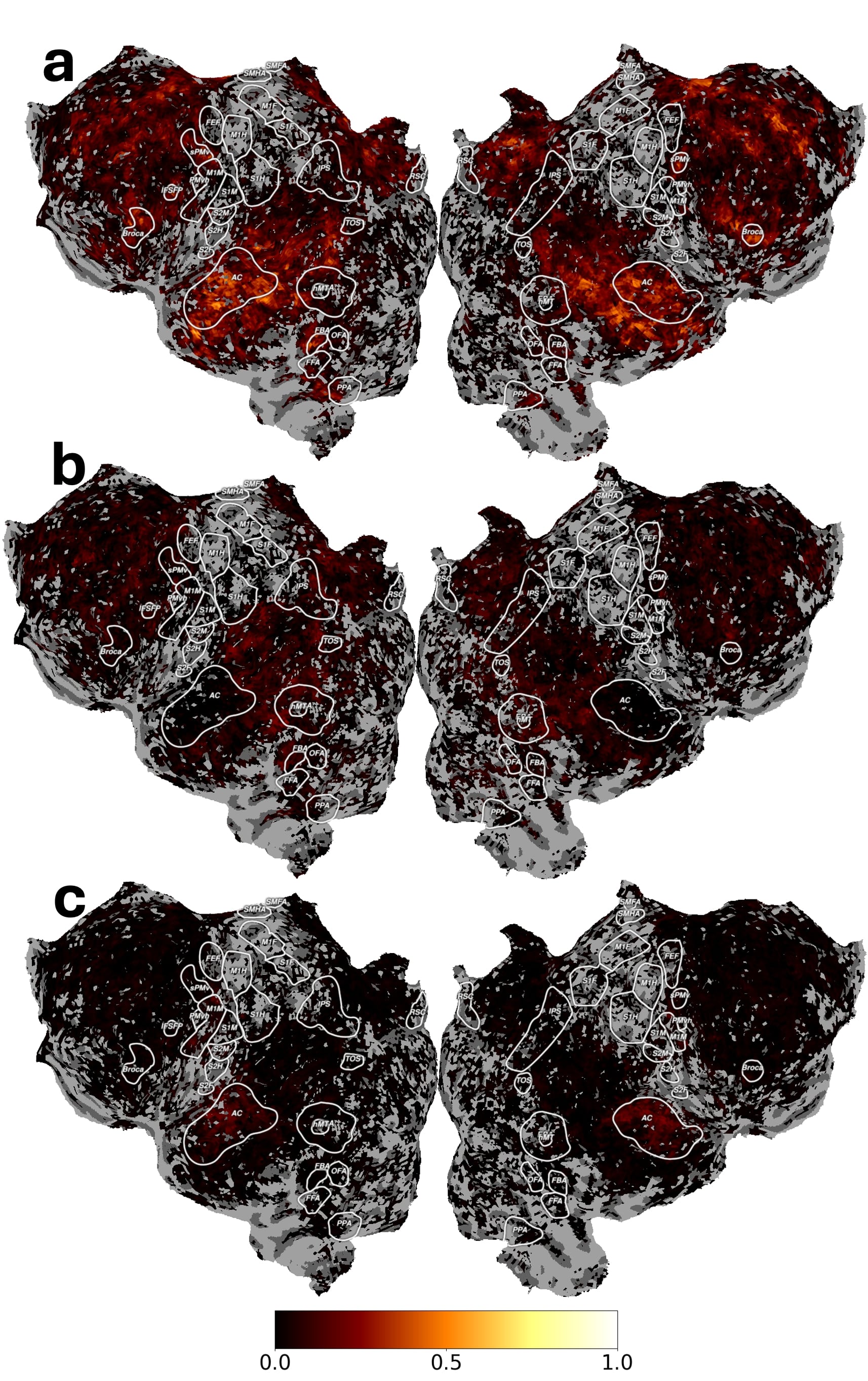}
\end{center}
\caption{Voxelwise variance partitioning analysis showing the contributions of different feature types to prediction accuracy for a subject S3 using MLP models. The flatmaps display (a) variance jointly explained by audio and semantic features, (b) variance uniquely explained by semantic features, and (c) variance uniquely explained by audio features. Values shown are normalized correlations ($CC_{norm}$) for voxels where the joint model achieved significant prediction ($q(\text{FDR})<0.01$).} 
\label{Fig S3 VP MLP}
\end{figure}

\FloatBarrier

\subsection{Largest variance partitioning for each voxel}\label{App : Largest variance partitioning for each voxel}
\begin{figure}[h]
\begin{center}
\includegraphics[width=\linewidth]{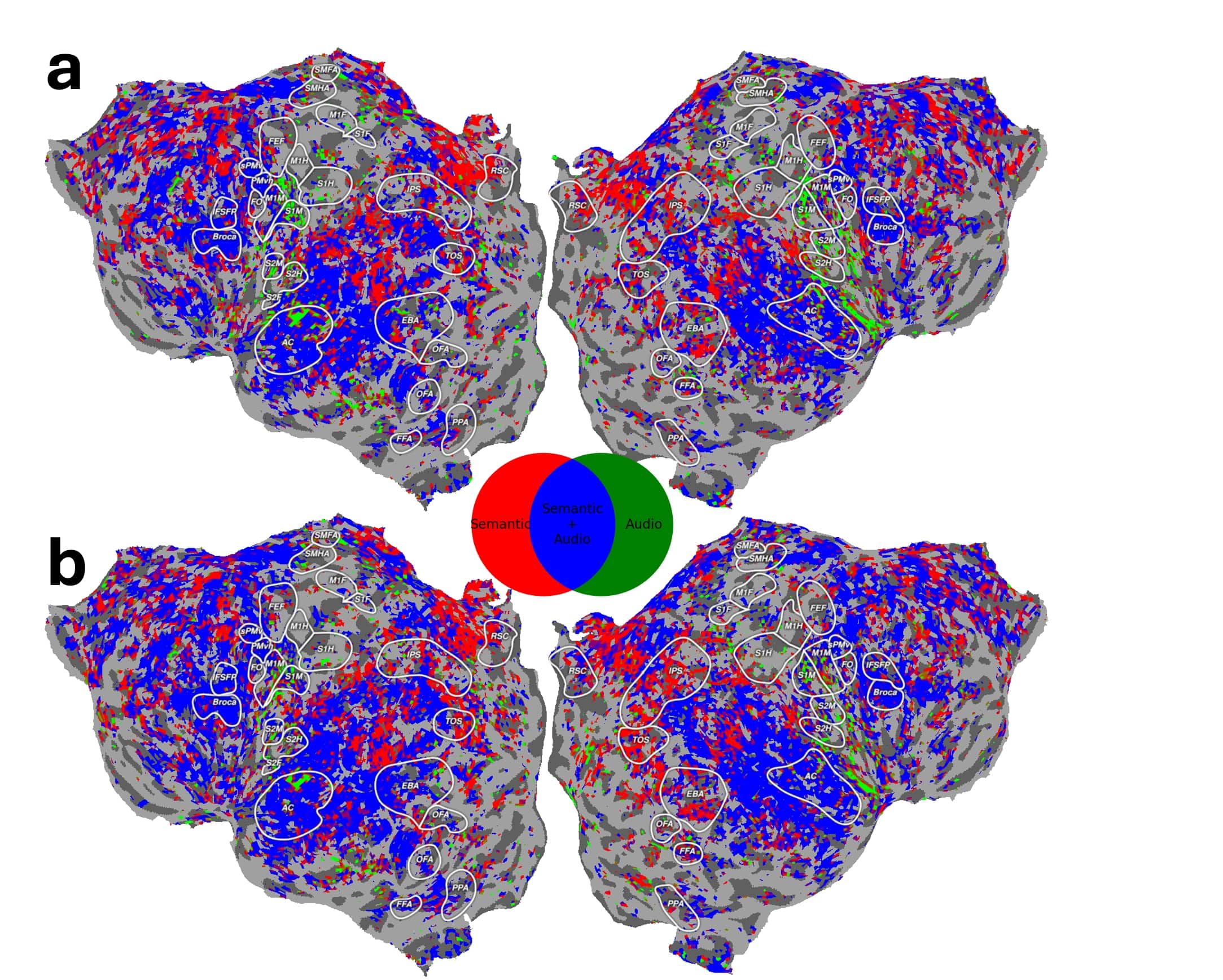}
\end{center}
\caption{Voxelwise analysis showing the largest variance explained by each feature type for all significantly predicted voxels ($q(\text{FDR})<0.01$) for subject S1. The flatmaps display which feature partition (semantic in red, audio in green, or their combination in blue) best explains the variance in each cortical voxel using (a) linear and (b) MLP encoders, with outlined regions indicating key functional areas. } 
\label{Fig VP Fig6 S1}
\end{figure}

\begin{figure}[h]
\begin{center}
\includegraphics[width=0.8\linewidth]{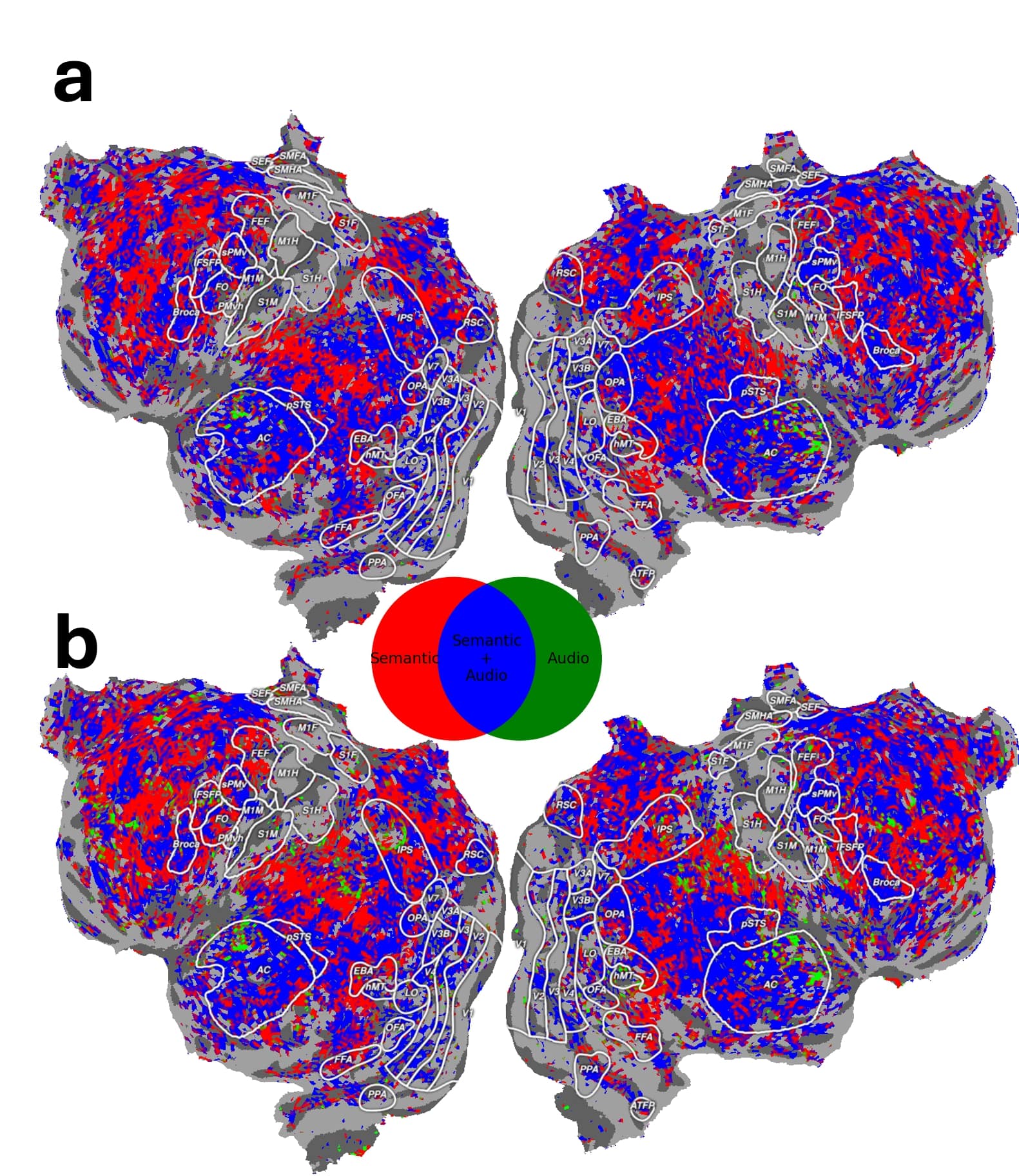}
\end{center}
\caption{Same as Figure \ref{Fig VP Fig6 S1}, but for subject S2} 
\label{Fig VP Fig6 S2}
\end{figure}

\begin{figure}[h]
\begin{center}
\includegraphics[width=0.9\linewidth]{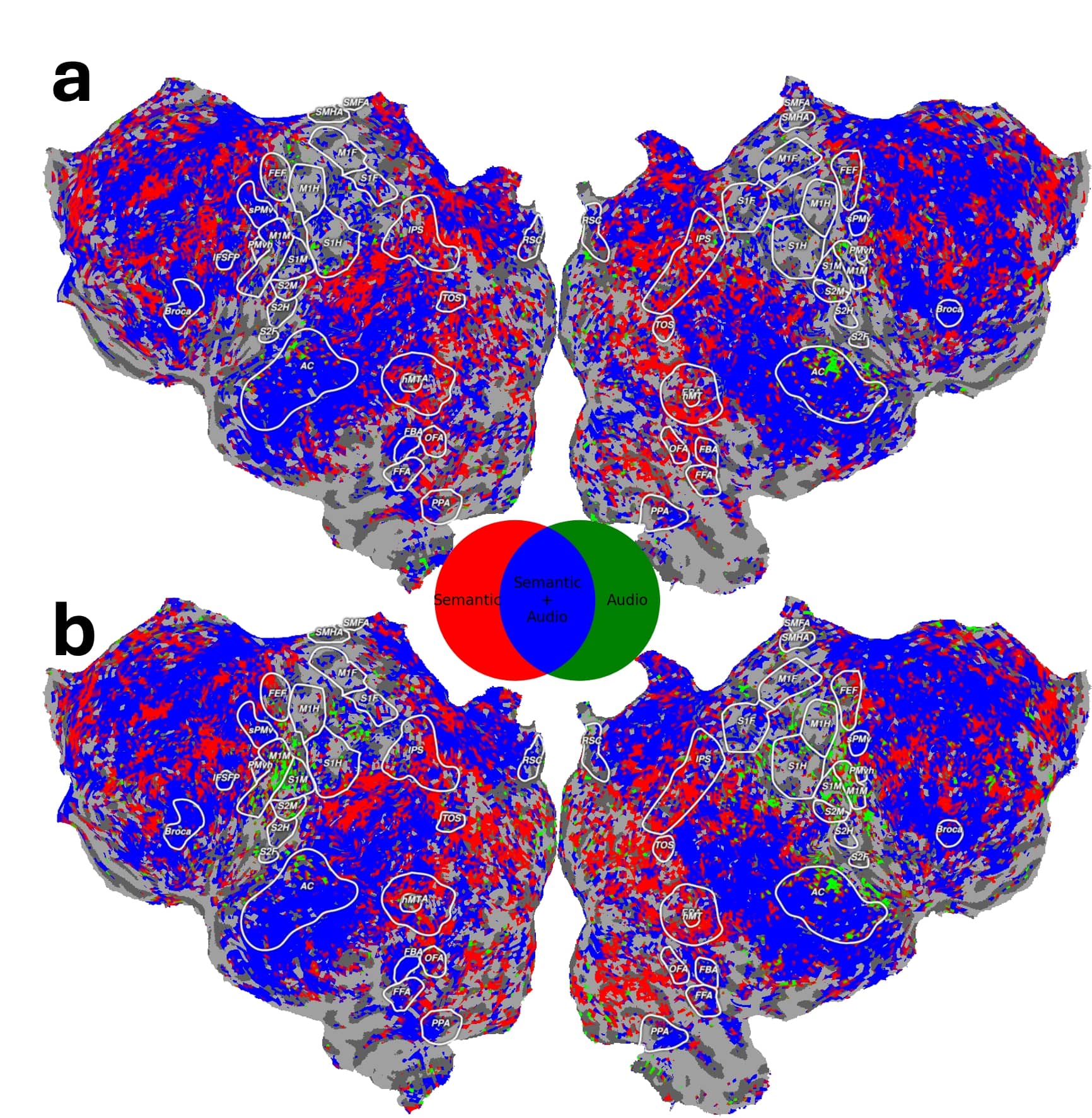}
\end{center}
\caption{Same as Figure \ref{Fig VP Fig6 S1}, but for subject S3} 
\label{Fig VP Fig6 S3}
\end{figure}

\FloatBarrier

\subsection{Variance partitioning Venn diagram}\label{App Variance partitioning Venn diagram}

\begin{figure}[h]
\begin{center}
\includegraphics[width=\linewidth]{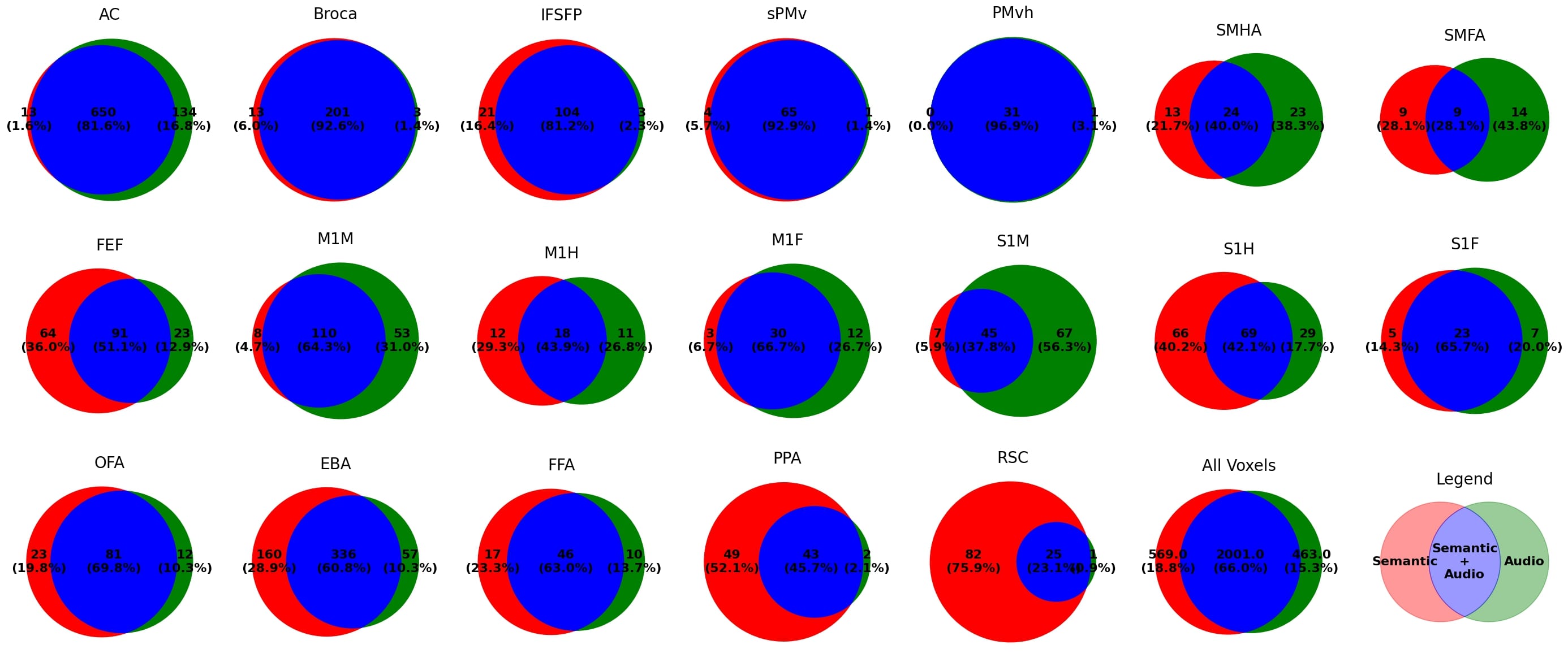}
\end{center}
\caption{Venn diagrams showing the distribution of explained variance across different brain regions of interest (ROIs) for subject S1, using linear encoder. Each diagram displays the unique and shared variance explained by semantic features (red), audio features (green), and their overlap (blue). Values indicate the number of significantly predicted voxels and their percentages. Only the voxels that was predicted statistically significantly ($q(\text{FDR})<0.01$) was used in the analysis} 
\label{Fig VP Venn S1 Linear}
\end{figure}

\begin{figure}[h]
\begin{center}
\includegraphics[width=\linewidth]{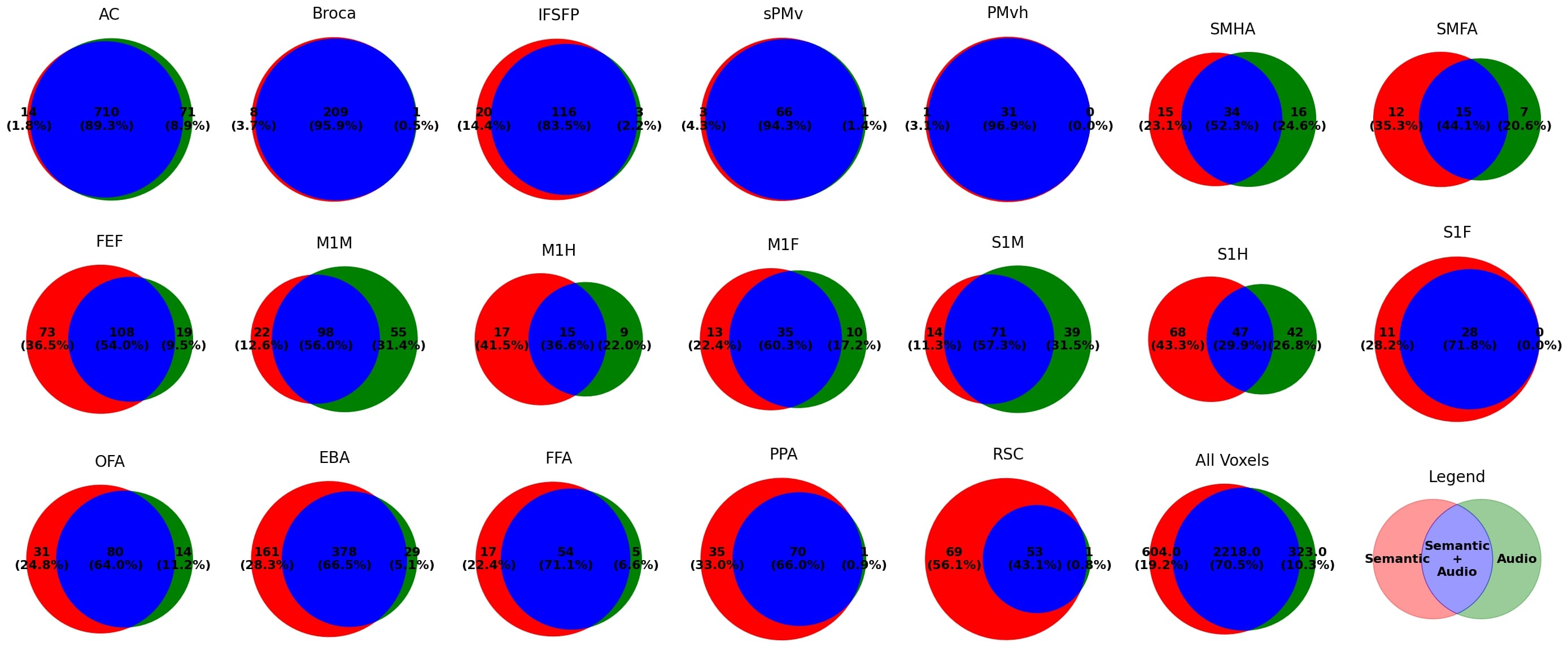}
\end{center}
\caption{Venn diagrams showing the distribution of explained variance across different brain regions of interest (ROIs) for subject S1, using MLP encoder. Refer to Fig \ref{Fig VP Venn S1 Linear} for more detail.} 
\label{Fig VP Venn S1 MLP}
\end{figure}

\begin{figure}[h]
\begin{center}
\includegraphics[width=\linewidth]{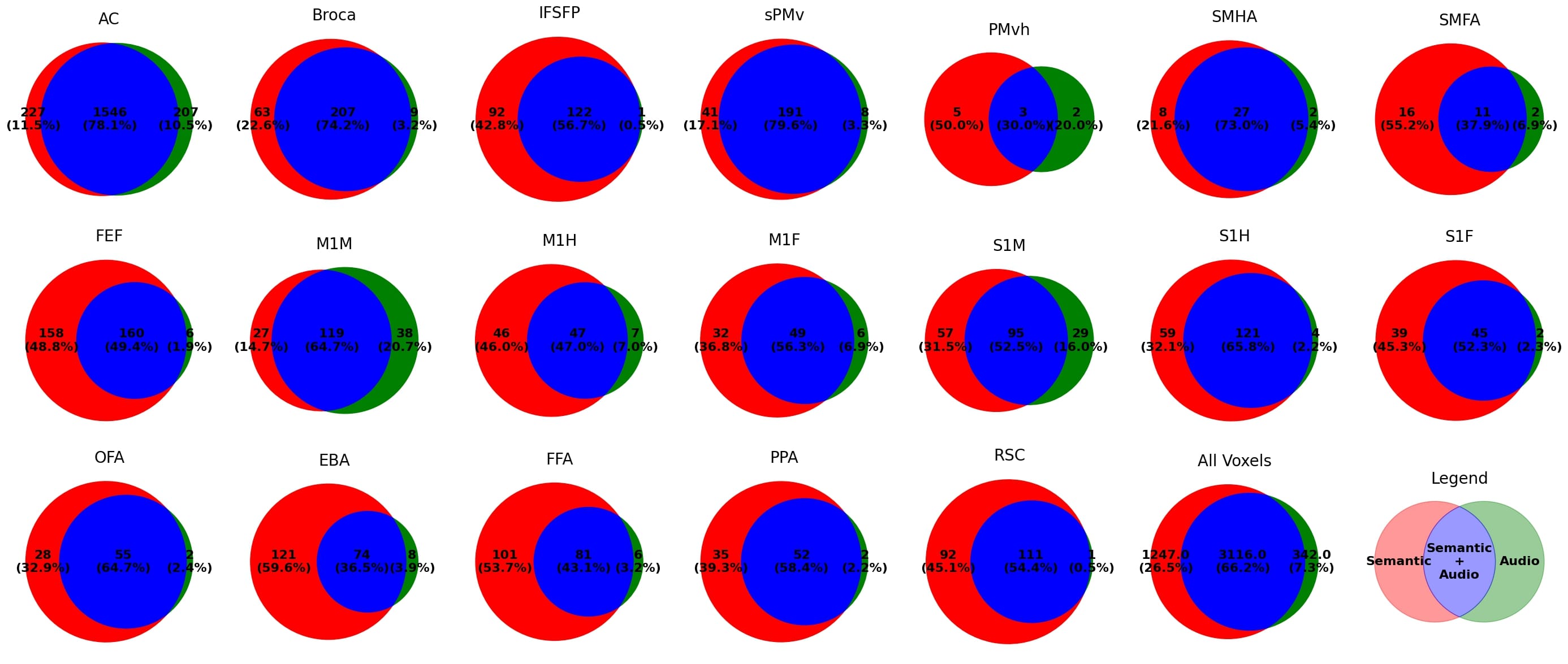}
\end{center}
\caption{Venn diagrams showing the distribution of explained variance across different brain regions of interest (ROIs) for subject S2, using linear encoder. Refer to Fig \ref{Fig VP Venn S1 Linear} for more detail.} 
\label{Fig VP Venn S2 Linear}
\end{figure}

\begin{figure}[h]
\begin{center}
\includegraphics[width=\linewidth]{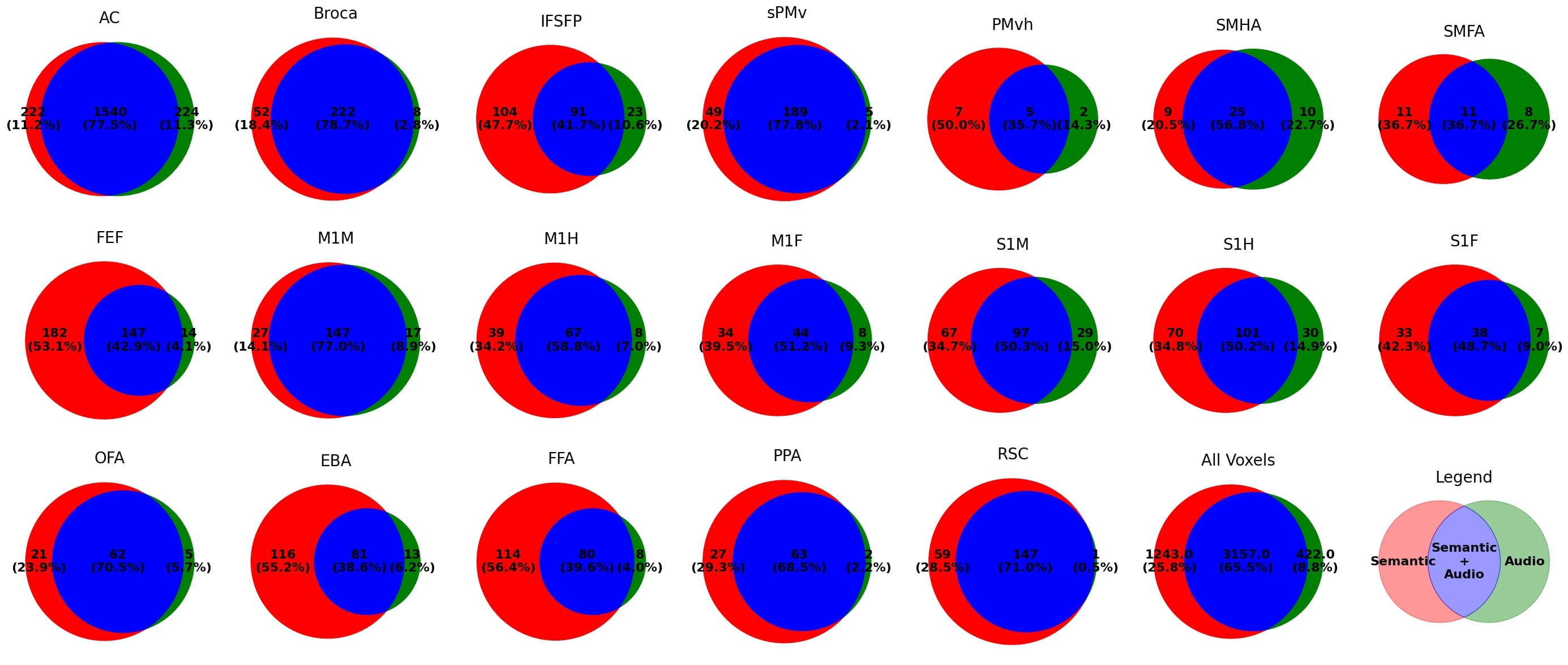}
\end{center}
\caption{Venn diagrams showing the distribution of explained variance across different brain regions of interest (ROIs) for subject S2, using MLP encoder.Refer to Fig \ref{Fig VP Venn S1 Linear} for more detail.} 
\label{Fig VP Venn S2 MLP}
\end{figure}

\begin{figure}[h]
\begin{center}
\includegraphics[width=\linewidth]{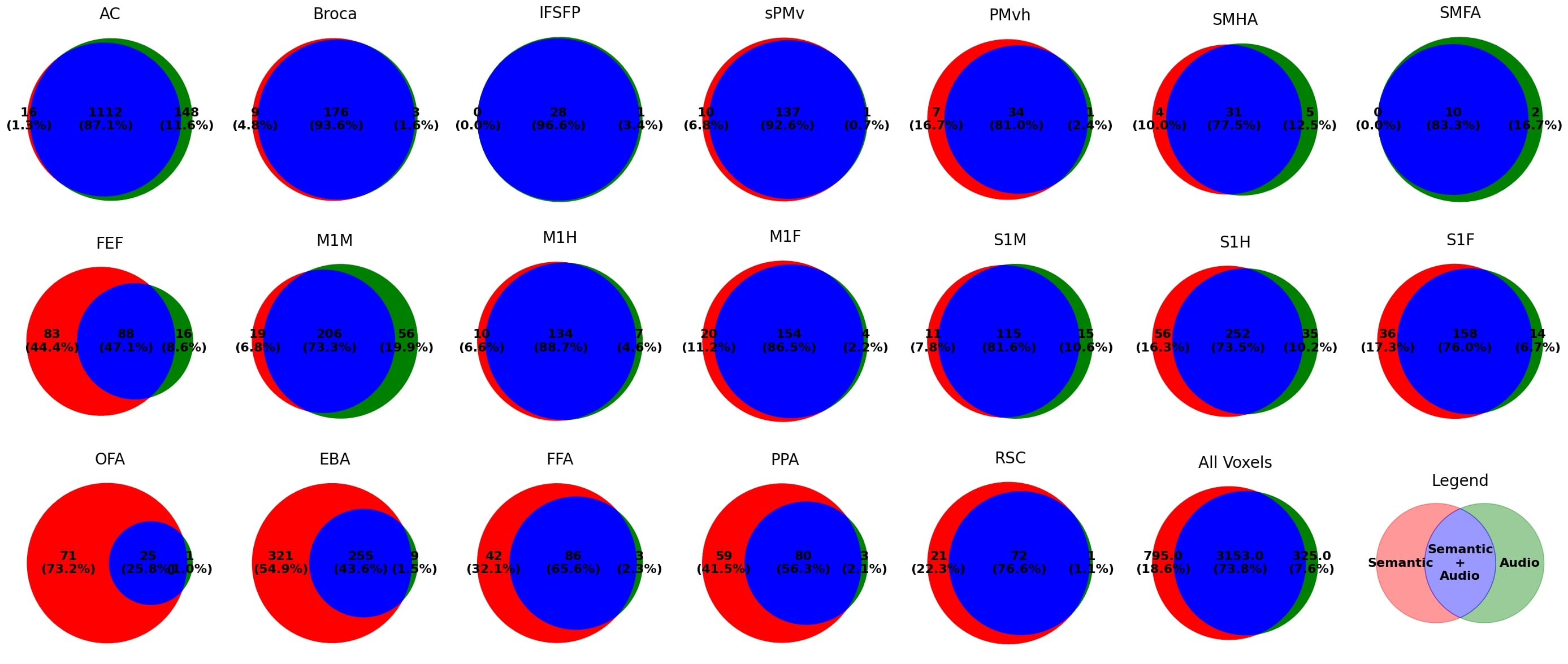}
\end{center}
\caption{Venn diagrams showing the distribution of explained variance across different brain regions of interest (ROIs) for subject S3, using linear encoder. Refer to Fig \ref{Fig VP Venn S1 Linear} for more detail.} 
\label{Fig VP Venn S3 Linear}
\end{figure}

\begin{figure}[h]
\begin{center}
\includegraphics[width=\linewidth]{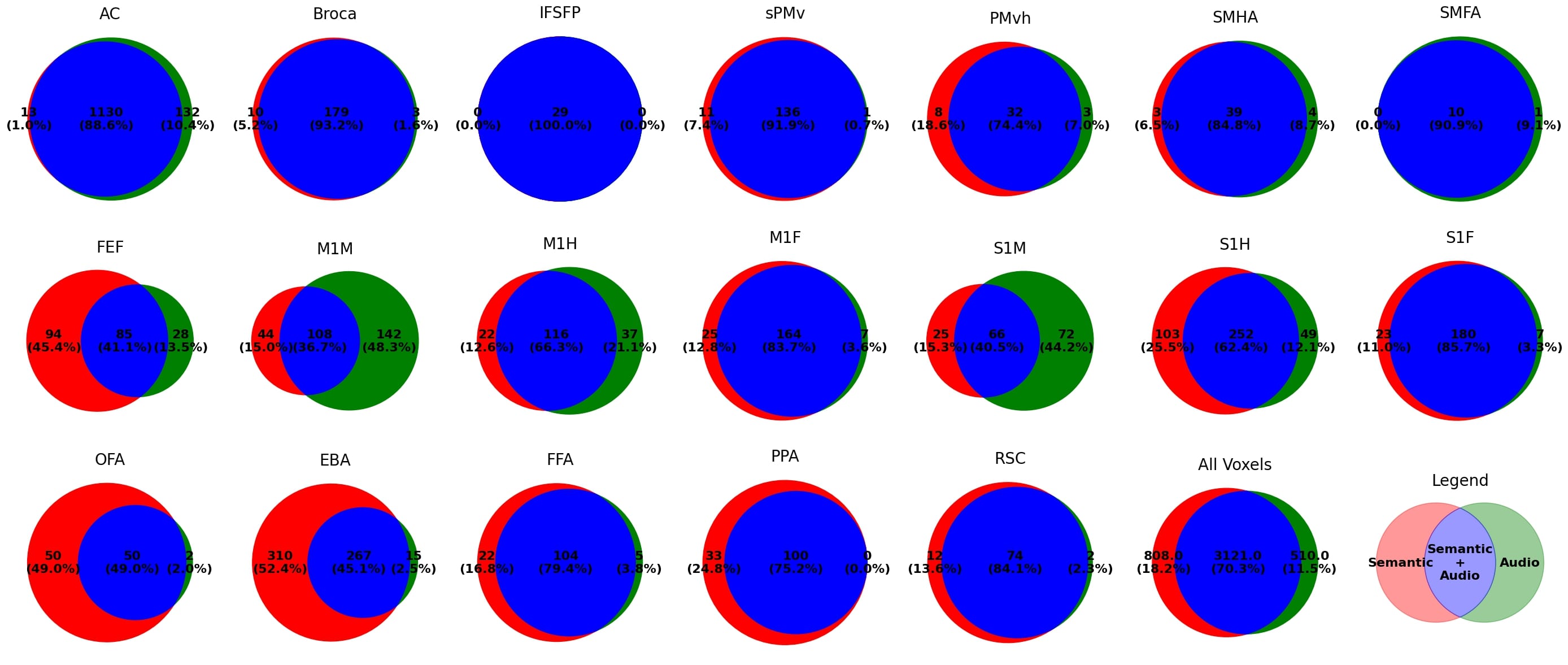}
\end{center}
\caption{Venn diagrams showing the distribution of explained variance across different brain regions of interest (ROIs) for subject S3, using MLP encoder. Refer to Fig \ref{Fig VP Venn S1 Linear} for more detail.} 
\label{Fig VP Venn S3 MLP}
\end{figure}

\FloatBarrier

\end{document}